\documentclass[sigconf]{acmart}

\AtBeginDocument{%
  \providecommand\BibTeX{{%
    \normalfont B\kern-0.5em{\scshape i\kern-0.25em b}\kern-0.8em\TeX}}}

\setcopyright{acmcopyright}
\copyrightyear{2023}
\acmYear{2023}
\setcopyright{acmlicensed}\acmConference[KDD '23]{Proceedings of the 29th
	ACM SIGKDD Conference on Knowledge Discovery and Data Mining}{August
	6--10, 2023}{Long Beach, CA, USA}
\acmBooktitle{Proceedings of the 29th ACM SIGKDD Conference on Knowledge
	Discovery and Data Mining (KDD '23), August 6--10, 2023, Long Beach, CA,
	USA}
\acmPrice{15.00}
\acmDOI{10.1145/3580305.3599404}
\acmISBN{979-8-4007-0103-0/23/08}

\usepackage{array}
\usepackage{courier}  % DO NOT CHANGE THIS
\usepackage{algorithm}
\usepackage{algorithmic}

\usepackage{amsmath}
\usepackage{amsthm}
\usepackage{enumerate}
\usepackage{subfigure}
\usepackage{bm}
\usepackage{tabularx}
\usepackage{enumitem}
\usepackage{multirow}
\usepackage{xcolor}
\usepackage{nicefrac}
\usepackage{booktabs}
\usepackage{etoc}
\etocdepthtag.toc{mtchapter}
\etocsettagdepth{mtchapter}{subsection}
\etocsettagdepth{mtappendix}{none}
\usepackage{balance}

\newtheorem{prop}{Proposition}


\newcolumntype{L}[1]{>{\raggedright\let\newline\\\arraybackslash\hspace{0pt}}m{#1}}
\newcolumntype{C}[1]{>{\centering\let\newline  \\\arraybackslash\hspace{0pt}}m{#1}}%
\newcolumntype{R}[1]{>{\raggedleft\let\newline \\\arraybackslash\hspace{0pt}}m{#1}}

\begin{document}

%\title{Neural Knowledge Graph Reasoning goes Deeper}
%\title{Learning Deeper Propagation for Knowledge Graph Reasoning}
\title{AdaProp: Learning Adaptive Propagation for Graph Neural Network based
	 Knowledge Graph Reasoning}


%\author{%
%	Yongqi Zhang$^{1}$
%	\quad
%	Zhanke Zhou$^{2}$
%	\quad
%	Quanming Yao$^{3}$
%	\quad
%	Xiaowen Chu$^4$
%	\quad
%	Bo Han$^2$\\
%	$^1$4Paradigm Inc., Beijing, China \\
%	$^2$Department of Computer Science, Hong Kong Baptist University, Hong Kong, China \\
%	$^3$Department of Electronic Engineering, Tsinghua University, Beijing, China \\
%	$^4$Data Science and Analytics Thrust, HKUST (Guangzhou), Guangzhou, China\\
%	zhangyongqi@4paradigm.com, 
%	\{cszkzhou,bhanml\}@comp.hkbu.edu.hk\\
%	qyaoaa@tsinghua.edu.cn, 
%	xwchu@ust.hk
%}

\author{Yongqi Zhang}
\authornote{This work was partially performed 
	when Z. Zhou was an intern in 4Paradigm.\\
	Y. Zhang and Z. Zhou have equal contribution,
	and correspondence is to Q. Yao.}
\email{zhangyongqi@4paradigm.com}

\affiliation{
	\institution{4Paradigm Inc.}
	\city{Beijing}
	\country{China}
}

\author{Zhanke Zhou}
\authornotemark[1]
\email{cszkzhou@comp.hkbu.edu.hk}
\affiliation{%
	\institution{Department of Computer Science, Hong Kong Baptist University}
%	\streetaddress{1 Th{\o}rv{\"a}ld Circle}
	\city{Hong Kong}
	\country{China}}

\author{Quanming Yao}
\authornotemark[2]
\email{qyaoaa@tsinghua.edu.cn}
\affiliation{%
	\institution{Department of Electronic Engineering, Tsinghua University}
	\city{Beijing}
	\country{China}
}

\author{Xiaowen Chu}
\email{xwchu@ust.hk}
\affiliation{
	\institution{Data Science and Analytics Thrust, HKUST (Guangzhou)}
	\city{Guangzhou}
	\country{China}
}

\author{Bo Han}
\email{bhanml@comp.hkbu.edu.hk}
\affiliation{%
	\institution{Department of Computer Science, Hong Kong Baptist University}
	\city{Hong Kong}
	\country{China}
}

\begin{abstract}
Due to the popularity of Graph Neural Networks (GNNs), 
various GNN-based methods have been designed to reason on knowledge graphs (KGs).
An important design component of GNN-based KG reasoning methods
is called the propagation path,
which contains a set of involved entities in each propagation step.
Existing methods
use hand-designed propagation paths,
ignoring the correlation between the entities and the query relation.
In addition, the number of involved entities will explosively grow
at larger propagation steps.
%Such a propagation path should
%preserve the promising targets
%while avoiding the exponentially growing number of involved entities.
%However,
%removing entities during propagation is challenging
%since the targets may also be removed.
In this work,
we are motivated to learn an adaptive propagation path
in order to filter out irrelevant entities while preserving promising targets.
%This motivates us to learn an adaptive propagation path
%that filters out irrelevant entities while preserving promising targets
%during the propagation.
First,
we design an incremental sampling mechanism
where the nearby targets 
and layer-wise connections can be preserved
with linear complexity.
Second,
we design a learning-based sampling distribution
to identify the semantically related entities.
Extensive experiments show that
our method is powerful, efficient
and semantic-aware.
%It achieves state-of-the-art performances
%in both transductive and inductive reasoning settings.
The code is available at
\url{https://github.com/LARS-research/AdaProp}.
%Due to the success of Graph Neural Networks (GNNs) in learning from graph-structured data, various GNN-based methods have been introduced to learn from knowledge graphs (KGs). 
%In this paper, 
%to reveal the key factors underneath existing GNN-based methods, 
%we revisit exemplar works from the lens of the propagation path. 
%\textcolor{blue}{
%	We find that the
%	answer entity can be \textit{close} to queried one,
%	but a properly designed propagation path
%	that can go \textit{deeper}
%	has the potential to have better reasoning performance.
%	However,
%	learning a deeper propagation path
%	is challenging
%	since the number of involved entities grows exponentially.}

\end{abstract}


%\ccsdesc[300]{Information systems~Web searching and information discovery}
%\ccsdesc[500]{Computing methodologies~Knowledge representation and reasoning}
%\ccsdesc[500]{Computing methodologies~Machine learning algorithms}

\maketitle

\begin{figure}[t]
	\centering
	\includegraphics[width=0.9\columnwidth]{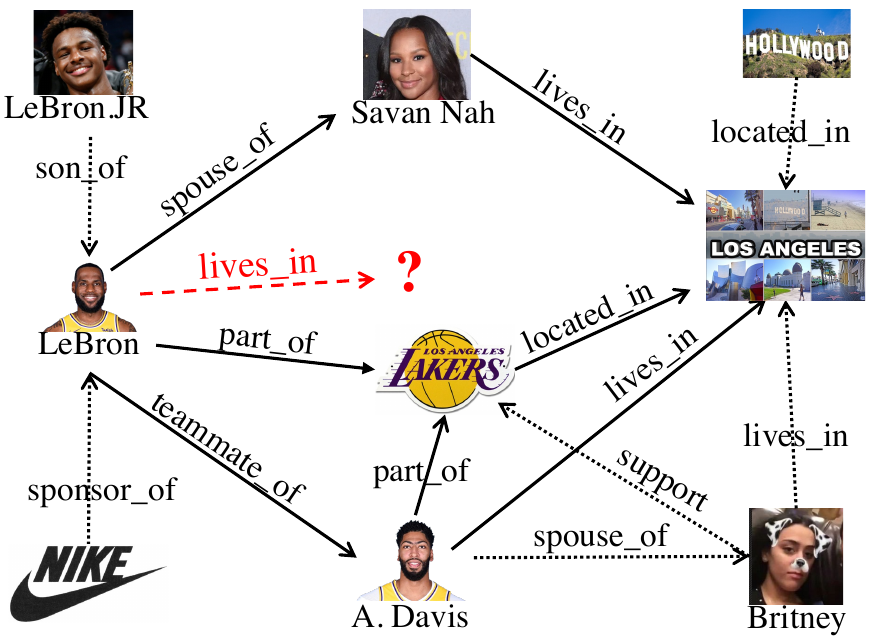}
	\vspace{-10px}
	\caption{An example of KG reasoning.
		The query \textit{(Lebron, lives\_in,?)} asks which city \textit{Lebron} lives in.
		Dashed lines are irrelevant facts to this query
		while solid lines are relevant.
	}
	\label{fig:kg}
	\vspace{-10px}
\end{figure}

\section{Introduction}
\label{sec:intro}

Knowledge graph (KG) is a semantic graph 
representing the relationships between real-world entities \cite{wang2017knowledge,hogan2021knowledge}.
Reasoning on KGs
aims to deduce missing answers for given queries based on existing facts~\cite{battaglia2018relational,ji2020survey,chen2020review}.
It has been widely applied in
drug interaction prediction~\cite{yu2021sumgnn},
personalized recommendation~\cite{cao2019unifying,wang2019kgat},
and question answering~\cite{dhingra2019differentiable,huang2019knowledge}.
%\footnote{+zk+ somehow not smooth. It would be better to add some transition sentences before you elaborate the problem.}
%
An example of KG reasoning with query
\textit{(LeBron, lives\_in, ?)}
and answer
\textit{Los Angeles} is illustrated in Figure~\ref{fig:kg}.
Symbolically,
the reasoning problem can be denoted as a \textit{query} $(e_q,r_q,?)$, 
with the query entity $e_q$ and query relation $r_q$.
The objective of KG reasoning 
is to find the target answer entity $e_a$ for $(e_q,r_q,?)$
by the local evidence 
learned from the given KG.

%An example of KG reasoning with query
%\textit{(LeBron, lives\_in, ?)}
%and answer
%\textit{Los Angeles} is shown in Figure~\ref{fig:kg}.

Representative methods for KG reasoning can be classified into three categories:
(i) triplet-based models \cite{bordes2013translating,dettmers2017convolutional,wang2017knowledge,zhang2019quaternion,zhang2020autosf}
directly score each answer entity 
through the learned
entity and relation embeddings;
(ii) path-based methods \cite{das2017go,sadeghian2019drum,qu2021rnnlogic,cheng2022rlogic,zhang2020interstellar}
learn logic rules to generate
the relational paths
starting from $e_q$ and explore which entity is more likely to be the target answer $e_a$;
(iii) GNN-based methods \cite{schlichtkrull2018modeling,vashishth2019composition,zhang2021knowledge,zhu2021neural}
%including R-GCN \cite{schlichtkrull2018modeling}, CompGCN \cite{vashishth2019composition}, 
%RED-GNN \cite{zhang2021knowledge} and NBFNet \cite{zhu2021neural},
propagate entity representations among the local neigborhoods
by graph neural networks \cite{kipf2016semi}.
Among the three categories,
the GNN-based methods achieve the state-of-the-art performance
\cite{zhang2021knowledge,zhu2021neural,ye2022comprehensive}.
%The main idea is to propagate messages
%around the local neighborhoods
%in the message propagation framework \cite{gilmer2017neural}.
%\footnote{+zk+ of similar meaning with the previous sentence
%	"propagate representations among local neigborhoods of entities".}
%The full propagation methods, 
Specifically,
R-GCN \cite{schlichtkrull2018modeling} and CompGCN
\cite{vashishth2019composition},
propagate in the full neighborhoods over all the entities.
Same as the general GCNs, 
they will easily suffer from over-smoothing problem \cite{oono2019graph}.
NBFNet \cite{zhu2021neural} and RED-GNN \cite{zhang2021knowledge}
propagates in the local neighborhoods around the query entity $e_q$.
By selecting the range of propagation,
they possess the stronger reasoning ability
than R-GCN or CompGCN.
However,
their local neighborhoods 
mainly depend on the query entity,
and the majority of entities will be involved %in the propagation
at larger propagation steps.

\begin{figure*}
	\centering
	\vspace{-4px}
	\includegraphics[height=2.75cm]{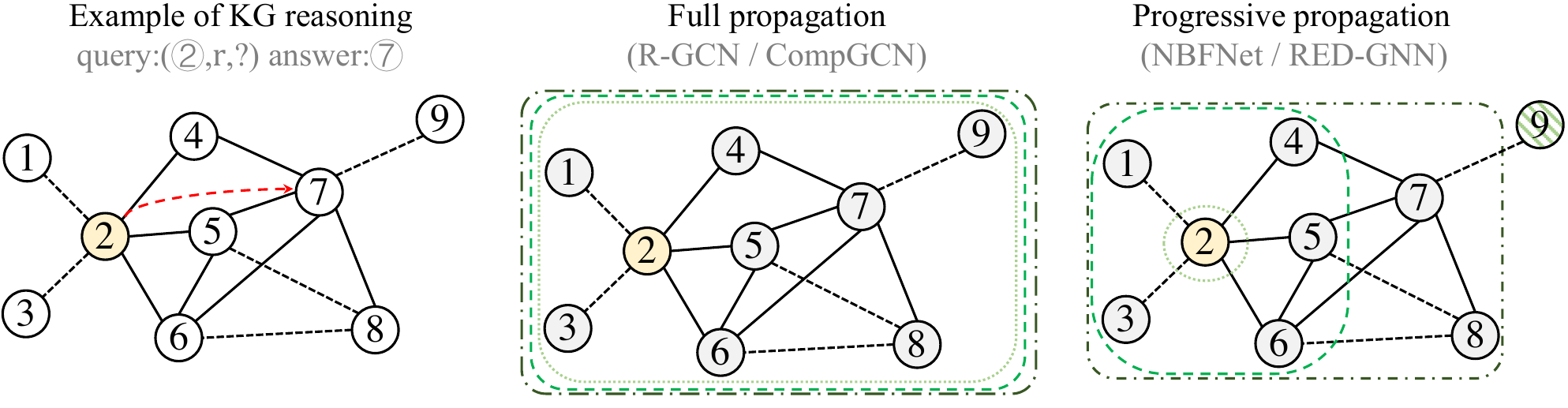}\hfill
	\includegraphics[height=2.75cm]{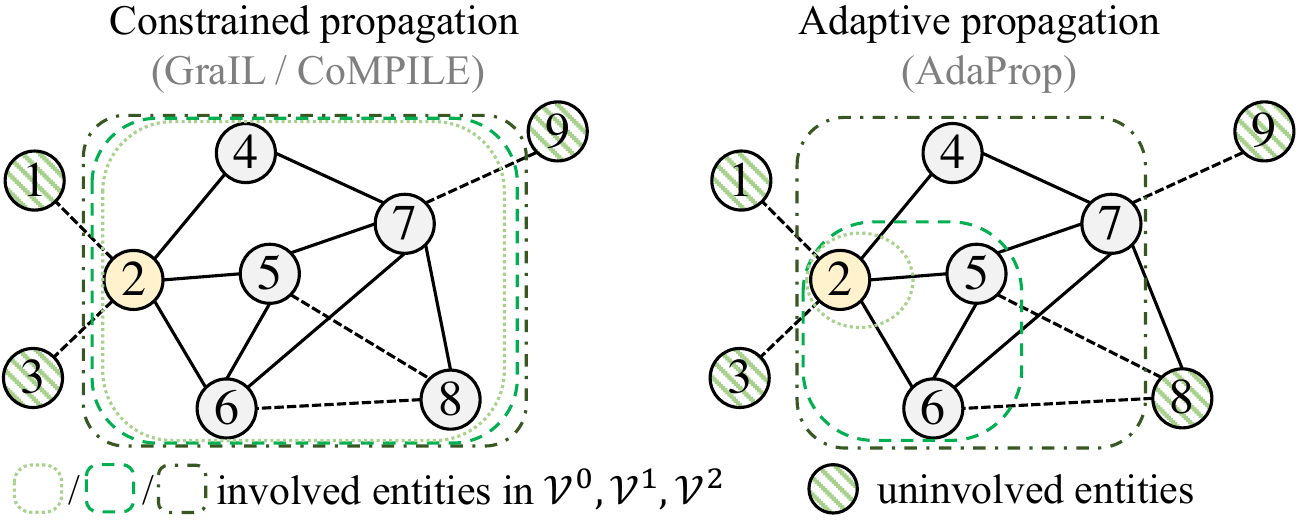}
	\vspace{-10px}
	\caption{Symbolic illustrations of the KG in Figure~\ref{fig:kg} 
		and the different designs of propagation path.
	For simplicity, we use circles to represent entities and ignore the types of relation.
	Specifically,
	(1) full propagation propagates over the full set of entities $\mathcal V$;
	(2) progressive propagation starts from the query entity $e_q$ and gradually propagates to its $\ell$-hop neighbors in the $\ell$-th step;
	(3) constrained propagation propagates within a constrained range $\mathcal V_{e_q,e_a}^L$ with entities
	less than 2 distance away from both $e_q$ and $e_a$;
	(4) adaptive propagation of AdaProp starts from $e_q$ and adaptively selects the semantic-relevant entities.
}
\vspace{-6px}
\label{fig:propagation-path}
\end{figure*}

%However,
%NBFNet and RED-GNN do not consider the relevance between
%local neighborhoods and the query relation,
%and
%they suffer from explosively growing number of entities in larger propagation steps.

The existing GNN-based methods,
which propagate among all the neighborhood over all the entities
or around the query entity,
ignore the semantic relevance between local neighborhoods 
and the query relation.
Intuitively,
the local information we need depends on both the query entity and query relation.
For example,
the most relevant local evidence to the query in Figure~\ref{fig:kg} 
includes the combination of 
\textit{(LeBron, part\_of, Lakers)} \& \textit{(Lakers, located\_in, L.A.)}
and \textit{(LeBron, spouse\_of, Savan Nah)} \& \textit{(Savan Nah, lives\_in, L.A.)}.
If the query changes to \textit{(LeBron, father\_of, ?)},
the most relevant information will be changed to
\textit{(LeBron JR, son\_of, LeBron)}.
In addition,
the existing GNN-based methods will inevitably involve
too many irrelevant entities and facts,
increasing the learning difficulty
and
computing cost,
especially on large-scale KGs.

%Since there is no direct strategy that indicates how to select the 
%semantic similar entities,
%the hand-designed propagation path by existing GNN-based methods
%do not differentiate the different neigborhoods,
%leading to sub-optimal reasoning performance.

Motivated by above examples,
we propose to adaptively sample semantically relevant entities 
during propagation.
There are three challenges when designing the sampling algorithm:
(i) the structure of KG is complex and the size is large;
%the number of involved entities in large KGs will explosively increase as propagation steps 
%get large;
(ii) the KG is multi-relational where the edges and queries have different kinds of relations,
representing different semantic meanings;
(iii) there is no direct supervision and simple heuristic measurement 
to indicate the semantic relevance of entities to the given query.
To address these challenges,
we propose AdaProp,
a GNN-based method with an adaptive propagation path.
The key idea is to reduce the number of involved entities
while preserving the relevant ones in the propagation path.
%we firstly define important factors based on the query dependent propagation path.
%The key factor is to reduce the number of involved entities 
%while preserving the target answer entities in the propagation path.
%Considering that the target entity is unknown given a query,
%we are motivated to learn an adaptive propagation path,
%named as AdaProp,
%which can filter out irrelevant entities
%while preserving the promising targets.
This is achieved by
an incremental sampling mechanism
and a learning-based sampling distribution
that adapts to different queries.
In this way,
our method 
learn an adaptive propagation path
that preserves layer-wise connection,
selects semantic related entities,
with fewer reduction in the promising targets.
%deepens the GNN 
%such that it can capture the long-range information.
The main contributions
are summarized as follows:
\begin{itemize}[leftmargin=12pt]
	\item We propose a connection-preserving sampling scheme,
	called incremental sampling,
	which only has linear complexity with regard to the propagation steps 
	and can preserve the layer-wise connections between sampled entities.
	
	\item We design a semantic-aware Gumble top-$k$ distribution
	which adaptively selects local neighborhoods relevant to the query relation
	and is learned by a straight through estimator.
	
	\item 
	Experiments show that the proposed method achieves the state-of-the-art performance in both transductive and inductive KG reasoning settings.
	It can preserve more target entities than the other samplers.
	And the case study shows that the learned sampler is query-dependent and semantic-aware.
	
\end{itemize}

Let $\mathcal G=(\mathcal V, \mathcal R, \mathcal E, \mathcal Q)$ be an instance of KG,
where {\small$\mathcal{V}$, $\mathcal R$} are the sets of entities and relations,
{\small$\mathcal E\!=\!\{(e_s,r,e_o)|e_s,e_o\!\in\!\mathcal{V}, r\!\in\!\mathcal R\}$}
is the set of fact triplets,
and {\small$\mathcal Q\!=\!\{(e_q,r_q,e_a)|e_q,e_a\!\in\!\mathcal{V}, r\!\in\!\mathcal R\}$}
is the set of query triplets with 
\footnote{Note that predicting missing head in KG can also be formulated in this way by adding inverse relations (details in Appendix~\ref{app:inverse}).}
query $(e_q,r_q,?)$ and target answer entity $e_a$.
The reasoning task is to  learn a function 
that predicts the answer $e_a\in\mathcal V$ 
for each query $(e_q,r_q,?)$
based on the fact triplets $\mathcal E$.

\section{Related works}
\label{sec:related}

\subsection{GNN for KG reasoning}
\label{ssec:gnn}

Due to the success of
graph neural networks (GNNs)~\cite{gilmer2017neural,kipf2016semi}
in modeling the graph-structured data, 
recent works
\cite{vashishth2019composition,teru2019inductive,zhang2021knowledge,zhu2021neural}
attempt to 
utilize the power of 
GNNs for KG reasoning.
They generally follow the message propagation framework~\cite{gilmer2017neural,kipf2016semi}
to propagate messages among entities.
In the $l$-th propagation step, specifically,
the messages are firstly computed on edges 
$\mathcal E^\ell = \{(e_s, r, e_o)\in\mathcal E| e_s \in \mathcal V^{\ell-1}), r\in\mathcal R, e_o \in \mathcal V^\ell\}$
with message function
%\begin{equation*}
{\small $\bm m_{(e_s, r, e_o)}^\ell 
	:= \text{MESS}\left(\bm h_{e_s}^{\ell-1}, \bm h_{r}^{\ell}\right)$}
%	\label{eq:mess}
%\end{equation*}
based on entity representation {\small$\bm h^{\ell-1}_{e_s}$}
and relation representation {\small$\bm h_r^\ell$}.
And then the messages are
propagated from entities of previous step
({\small$e_s\in\mathcal{V}^{\ell-1}$})
to entities of current step
({\small$e_o\in\mathcal{V}^{\ell}$}) via
%\begin{equation*}
{\small $\bm h_{e_o}^{\ell} := \delta\left(\text{AGG}\big(\bm m_{(e_s, r, e_o)}^\ell, (e_s, r, e_o)\!\in\! \mathcal E^{\ell} \big)\right)$},
%\label{eq:agg}
%\end{equation*}
with an activation function $\delta(\cdot)$.
The specified functions $\text{MESS}(\cdot)$ and $\text{AGG}(\cdot)$ 
for different methods
are provided
in Appendix~\ref{app:mess-agg}.
After $L$ steps' propagation,
the entity representations {\small$\bm h_{e_o}^L$}
are obtained
to measure the plausibility of 
each entity {\small$e_o \in \mathcal{V}^{L}$}.

%The propagation path is an important design component for the GNN-based methods.
%It determines 
%the subgraph structure and
%the field of entities and edges 
%where messages are propagated.
%In addition,
%it selects entities as potential candidates
%whose representations are used for final prediction.
%Hence, we classify the existing GNN-based methods into three classes
%based on their design of propagation paths:

For the GNN-based KG reasoning methods,
the sets of entities participated in propagation are different.
The existing GNN-based methods can be generally classified into 
three classes based on their design of propagation range:
%\footnote{[enhanced] +qm+ the switch is too fast,
%	the importance of prorogation is not sufficiently emphasize.
%	since you have extra space now,
%	you can expand this part.
%	\label{ft:2}}
\begin{itemize}[leftmargin=10px, itemsep=1pt,topsep=2pt,parsep=0pt,partopsep=2pt]
	\item \textit{Full} propagation methods,
	e.g., R-GCN~\cite{schlichtkrull2018modeling} and CompGCN~\cite{vashishth2019composition},
	propagate among all the entities,
	i.e., {\small$\mathcal{V}^{\ell} \! = \! \mathcal{V}$}.
	They are limited to
	small propagation steps
	due to the large memory cost
	and the over-smoothing problem of GNNs \cite{oono2019graph}
	over full neighbors.
	
	\item  \textit{Progressive} propagation methods (RED-GNN~\cite{zhang2021knowledge} and NBFNet~\cite{zhu2021neural}),
	propagate from 
	the query entity $e_q$
	and progressively to 
	the $L$-hop neighborhood of $e_q$,
	i.e., 
	{\small$\mathcal V^0\!=\!\{e_q\}$ and $\mathcal V^\ell \!=\! \bigcup_{e\in\mathcal V^{\ell-1}}\!\mathcal N(e)$},
	where {\small$\mathcal N(e)$} contains the 1-hop neighbors of entity $e$.
	
	\item \textit{Constrained} propagation methods, e.g., 
	GraIL~\cite{teru2019inductive} and CoMPILE~\cite{mai2021communicative},
	propagate within a constrained range,
	i.e., {\small$\mathcal{V}^{\ell} \! = \! \mathcal{V}_{e_q,e_a}^L$},
	where {\small$\mathcal{V}_{e_q,e_a}^L \! \subset \! \mathcal V$} is the enclosing subgraph 
	for $e_q$ and $e_a$.
	Since {\small$\mathcal{V}_{e_q,e_a}^L$} are different for different pairs of $(e_q, e_a)$,
	these methods are extremely expensive, 
	especially on large-scale KGs (see discussion in Appendix~\ref{app:grail}).
	
\end{itemize}

We provide graphical illustration of propagation scheme for the three categories in the middle of Figure~\ref{fig:propagation-path}.
Among them,
the progressive propagation methods achieved state-of-the-art reasoning performance.
They are much more efficient than the constrained propagation methods,
and the entities used for propagation depend on the query entity,
filtering out many irrelevant entities.
However,
$\mathcal V^L$ of progressive methods
will also involve in a majority or even all of the entities
when  $L$ gets large.

%\begin{definition}[Query-dependent propagation path]
%	\label{def:propagation_path}
%	The {propagation path}
%	for a {query} $(e_q,r_q,?)$	
%	is composed of entities
%	%	\footnote{+zk+ shall we explain what \textit{effective} means here? }
%	involved in each propagation step.
%	%		as in Algorithm~\ref{alg:propagation}.
%	The set of entities 
%	in the $\ell$-th step
%	is termed as {\small$\mathcal{V}^\ell$}
%	({\small$\ell \! = \! 0,1,...,L, \mathcal{V}^\ell \! \subseteq \! \mathcal{V}$}).
%	Namely, 
%	a $L$-layer propagation path is denoted as 
%	{\small$\widehat{\mathcal G}^{L} = \{ \mathcal{V}^0, \mathcal{V}^1, ..., \mathcal{V}^L \}$}.
%\end{definition}

\subsection{Sampling methods for GNN}
\label{ssec:sampling}

The sampling methods can control and select entities during propagation.
There are several methods introduced to sample nodes 
on homogeneous graphs
with the objective to
improve the scalability of GNNs.
They can be classified into three categories:
(i) node-wise sampling methods,
such as GraphSAGE~\cite{hamilton2017inductive} and PASS~\cite{yoon2021performance},
sample $K$ entities from the neighbor set $\mathcal N(e)$
for each entity $e$ in each propagation step;
(ii) layer-wise sampling methods, 
like FastGCN~\cite{chen2018fastgcn}, 
Adaptive-GCN~\cite{huang2018adaptive}
and
LADIES~\cite{zou2019layer},
sample at most $K$ entities 
in each propagation step;
(iii) subgraph sampling methods,
such as
Cluster-GCN~\cite{chiang2019cluster},
GraphSAINT~\cite{zeng2019graphsaint}
and ShadowGNN~\cite{zeng2021decoupling},
directly extract the local subgraph around the queried entity.
%These methods are mainly designed to reduce the scalability problem
%with simple heuristics on graphs.
All these methods are designed for homogeneous graphs
with the purpose of solving scalability issues.

There are also a few sampling methods proposed  for KG reasoning.
RS-GCN \cite{feeney2021relation} proposes a node-wise sampling method to select neighbors for
R-GCN/CompGCN.
DPMPN \cite{xu2019dynamically} contains a full propagation GNN to learn the global feature, 
and prunes another GNN for reasoning in a layer-wise sampling manner.
These sampling approaches reduce the memory cost for propagation on larger KGs,
but their empirical performance does not 
gain much compared with the base models.

\section{The proposed method}
\label{sec:method}

\subsection{Problem formulation}
\label{ssec:problem}

We denote a GNN model for KG reasoning as 
$F(\bm w, \widehat{\mathcal G}^L)$ 
%that predict the target answer entity of different queries $(e_q, r_q,?)$
with model parameters $\bm w$.
$\widehat{\mathcal G}^L$ here is called the \textit{propagation path},
containing the sets of involved entities in each propagation step,
i.e., $\widehat{\mathcal G}^L= \{ \mathcal{V}^0, \mathcal{V}^1, ..., \mathcal{V}^L \}$
%\footnote{$\surd$+qm+ better first show below two equations 
%	then show \eqref{eq:proppath}.
%	for most of GNN papers,
%	the equation for message passing usually comes first.
%	In this way,
%	\ref{ft:2} can be better addressed.}
where $\mathcal{V}^\ell, \ell\!=\!1\!\dots\! L$ is the set of involved entities in the $\ell$-th propagation step.
The representations of entities are propagated from $\mathcal V^0$ to $\mathcal V^{L}$
in $\widehat{\mathcal G}^L$
for multiple steps.

The objective is to design better propagation path $\widehat{\mathcal G}^L$
such that
the model with optimized parameters can correctly predict target answer entity $e_a$
for each query $(e_q, r_q, ?)$.
The progressive propagation methods have different propagation path according to the query entities,
and achieved state-of-the-art performance.
However,
they ignore the dependency of query relations
and will involve in too many entities at larger propagation steps.
Motivated by the success of progressive propagation methods
and sampling methods on homogeneous graph,
we aim to use sampling technique to
dynamically adapt the propagation paths.
Formally, we define the \textit{query-dependent propagation path} for a query $(e_q,r_q,?)$ as
\begin{align}
\begin{split}
	\widehat{\mathcal G}^L_{e_q,r_q} &= \big\{ \mathcal{V}^0_{e_q,r_q}, \mathcal{V}^1_{e_q,r_q}, ..., \mathcal{V}^L_{e_q,r_q} \big \}, \\
	\text{s.t. }~~ \mathcal V^\ell_{e_q,r_q} &=   
	\begin{cases}
		\{e_q\} & \ell=0  \\
		S(\mathcal{V}^{\ell-1}_{e_q,r_q}) & \ell=1\dots L
	\end{cases}.
\end{split}
\label{eq:proppath}
\end{align}
The propagation here starts from the query entity $e_q$,
and the entities in {\small $\mathcal{V}^\ell_{e_q,r_q}$} is adaptively sampled 
as a subset of {\small $\mathcal{V}^{\ell-1}_{e_q,r_q}$} and its neighbors
with a sampling strategy $S(\cdot)$.

The key problem here is how to design the sampling strategy $S(\cdot)$.
For the KG reasoning task,
we face two main challenges:
\begin{itemize}[leftmargin=10px, itemsep=3pt,topsep=2pt,parsep=0pt,partopsep=2pt]
	\item the target answer entity $e_a$ is unknown given the query $(e_q, r_q,?)$,
		thus directly sampling neighboring entities may loss the connection between query entity 
		and target answer entity;
	\item the semantic dependency between entities in the propagation path
		and the query relation $r_q$ is too complex
		to be captured by simple heuristics.
\end{itemize}

The existing sampling methods are not very applicable since
(i) they do not consider the preserving of unknown target entities;
(ii) they are designed for homogeneous graph without modeling relation types; and
(iii) there is no direct supervision on the dependency of entities 
in the propagation on the query.
Therefore, we would argue that a
specially designed sampling approach is needed for KG reasoning,
which is expected to remove entities
and also preserving the promising targets for corresponding query.

In the following parts,
we introduce the adaptive propagation (AdaProp) algorithm,
which can adaptively select relevant entities into the propagation path
based on the given query.
To solve the first challenge,
we design a connection-preserving sampling scheme,
called incremental sampling, 
which can
reduce the number of involved entities
and preserve the layer-wise connections
in Section~\ref{ssec:sample:increase}.
For the second challenge,
we propose a relation-dependent sampling distribution,
which is jointly optimized with the model parameters,
to select entities that are semantically relevant to the query relation
in Section~\ref{sssec:sample:learn}.
%Section~\ref{ssec:sample:full_algorithm} summarizes the full algorithm of AdaProp.
The full algorithm is summarized in Section~\ref{ssec:sample:full_algorithm}.

%\footnote{[enhanced] +qm+ motivation for the methodology design is not clear.
%	key words ``structure-preserving''
%	and
%	``semantic-aware'' are not mentioned above.
%	\label{ft:1}}

\subsection{The connection-preserving incremental sampling strategy}
\label{ssec:sample:increase}

%Following the state-of-the-art progressive propagation scheme~\cite{zhang2021knowledge,zhu2021neural},
%we start propagating from the query entity {\small$\mathcal{V}^0 \! = \! \{e_q\}$},
%but conduct sampling to reduce the number of involved entities
%in each propagation step.
When designing the sampling strategy,
we should consider what to be sampled, and how to sample effectively and efficiently.
Since the target entity $e_a$ is unknown given the query $(e_q, r_q, ?)$,
freely sampling from all the possible neighbors will
loss the structural connection between query entity and promising targets.
The existing sampling methods cannot well address this problem.
For the node-wise sampling methods,
the complexity grows exponentially w.r.t. propagation steps $L$.
For the layer-wise sampling methods,
the entities in previous step may have little connection with the sampled entities in the current step,
failing to preserve the local structures.
The subgraph sampling methods can better preserve local structure,
but is harder to control which entities are more relevant to the query and deserve to be sampled.

Observing that the majority of the target answer entities
lie close to the query entity 
(see the distributions in Appendix~\ref{app:distance}),
we are motivated to preserve the entities already been selected in each propagation step.
In other words, we use the constraint
{\small $\mathcal{V}^0_{e_q,r_q} \! \subseteq \! \mathcal{V}^1_{e_q,r_q} \! \cdots \! \subseteq \! \mathcal{V}^L_{e_q,r_q}$}
to 
preserve the previous step entities and
sample from the newly-visited ones 
in an incremental manner
to avoid dropping the promising targets.
On one hand,
the nearby targets will be more likely to be preserved when the propagation steps get deeper.
On the other hand,
the layer-wise connection between entities in two consecutive layers
can be preserved.
The incremental sampler $S(\cdot)=\texttt{SAMP}(\texttt{CAND}(\cdot))$ contains two parts, i.e.,
candidate generation $\texttt{CAND}(\cdot)$
and candidate sampling $\texttt{SAMP}(\cdot)$.

%Figure~\ref{fig:propagation_with_sampling} provides an example
%of 
%the sampling procedure
%obtaining $\mathcal{V}^\ell$ based on $\mathcal{V}^{\ell-1}$.

%\begin{figure}[!t]
%	\centering
%	\includegraphics[width=0.97\columnwidth]{figures/propagate-with-sampling}
%	\vspace{-4px}
%	\caption{
%		A graphical illustration of
%		propagation with incremental sampling.
%		For candidate generation,
%		{\small $\overline{\mathcal{V}}^\ell = \texttt{CAND}(\mathcal{V}^{\ell-1})$} 
%		are formed by the newly-visited entities in green.
%		After candidate sampling,
%		the sampled entities 
%		{\small $\texttt{SAMP}(\overline{\mathcal{V}}^\ell)$},
%		i.e., the entity \textcircled{6} and \textcircled{7},
%		are added to the propagation path for next step iteration.
%	}
%	\label{fig:propagation_with_sampling}
%\end{figure}

\subsubsection{Candidate generation}
Given the set of entities in the 
$(\ell\!-\!1)$-th step $\mathcal V^{\ell-1}_{e_q,r_q}$,
we denote the candidate entities to sample in the $\ell$-th step
as {\small $\overline{\mathcal V}^\ell_{e_q,r_q} := \texttt{CAND}(\mathcal V^{\ell-1}_{e_q,r_q})$}.
Basically,
all the neighboring entities in {\small $\mathcal N(\mathcal{V}^{\ell-1}_{e_q,r_q})\!=\!\bigcup_{e \in \mathcal V^{\ell-1}_{e_q,r_q}}\mathcal N(e)$}
can be regarded as candidates for sampling.
However,
to guarantee that {\small $\mathcal V^{\ell-1}_{e_q,r_q}$} will be preserved in the next sampling step,
we regard the newly-visited entities as the candidates,
i.e.,
\[\overline{\mathcal V}^\ell_{e_q,r_q} \; := \; \texttt{CAND}(\mathcal{V}^{\ell-1}_{e_q,r_q}) = \mathcal N(\mathcal{V}_{e_q,r_q}^{\ell-1}) \setminus \mathcal{V}^{\ell-1}_{e_q,r_q}.\]
For example,
given the entity \textcircled{2} in Figure~\ref{fig:propagation-path},
we generate the 1-hop neighbors
\textcircled{1}, \textcircled{3}, \textcircled{4}, \textcircled{5}, \textcircled{6}
as candidates
and directly preserve \textcircled{2} in the next propagation step.

\subsubsection{Candidate sampling}
%We aim to sample from the candidates $\bar{\mathcal V}^\ell$
%and filter out the entities that are irrelevant to the query $(e_q, r_q, ?)$
%in the propagation path.
To control the number of involved entities,
we sample $K$ ({\small $\ll|\overline{\mathcal{V}}^\ell_{e_q,r_q}|$}) entities without replacement 
from the candidate set {\small $\overline{\mathcal{V}}^\ell_{e_q,r_q}$}.
The sampled entities at $\ell$-th step
together with
entities in the $(\ell \! - \! 1)$-th step
constitute the involved entities at the $\ell$ step,
i.e.,
\[\mathcal{V}^\ell_{e_q,r_q} \; := \; \mathcal{V}^{\ell-1}_{e_q,r_q} \cup \texttt{SAMP}(\overline{\mathcal{V}}^\ell_{e_q,r_q}).\]
Given the candidates of entity \textcircled{2} in Figure~\ref{fig:propagation-path},
the entities \textcircled{5} and \textcircled{6}
are sampled and activated from the five candidates
in {\small $\overline{\mathcal V}^{\ell}_{e_q,r_q}$}.
Along with the previous layer entity \textcircled{2},
the three entities are collected as {\small $\mathcal V^\ell_{e_q,r_q}$} for next step propagation.

\subsubsection{Discussion}
There are three advantages of the incremental sampling strategy.
First,
it is more entity-efficient than the node-wise sampling methods,
which have exponentially growing number of entities over propagation steps.
As in Prop.\ref{pr:complexity},
the number of involved entities grows linearly w.r.t. the propagation steps $L$.
Second,
the layer-wise connection can be preserved based on  Prop.\ref{pr:preserve},
which shows advantage over the layer-wise sampling methods.
Third,
as will be shown in Section~\ref{sssec:exp:sampling},
it has larger probability to preserve the target answer entities
than the other sampling mechanisms
with the same amount of involved entities.

\begin{prop}
	\label{pr:complexity}
	The number of involved entities in the propagation path ({\small $\big|\!\bigcup_{\ell=0\dots L} \! \mathcal V^\ell_{e_q,r_q}\big|$}) of incremental sampling
	is bounded by $O(LK)$.
\end{prop}

\begin{prop}
	\label{pr:preserve}
	For all the entities {\small $e\in\mathcal V^{\ell-1}_{e_q,r_q}$} with incremental sampling,
	there exists at least one entity {\small $e'\in\mathcal V^{\ell}_{e_q,r_q}$} and relation $r\in\mathcal R$
	such that $(e, r, e')\in\mathcal E$.
\end{prop}

\subsection{Learning semantic-aware distribution}
\label{sssec:sample:learn}

The incremental sampling mechanism
alleviates the problem of 
reducing target candidates.
However,
randomly sampling $K$ entities from {\small $\overline{\mathcal V}^\ell_{e_q,r_q}$}
does not consider the semantic relevance of different entities for different queries.
The greedy-based  search algorithms like beam search \cite{wilt2010comparison}
are not applicable here as
there is no direct heuristic to measure the semantic relevance.
To solve this problem,
we parameterize the sampler $S(\cdot)$ with parameter $\bm \theta^\ell$
for each propagation step $\ell$,
i.e.,
{\small $\mathcal V^\ell_{e_q,r_q} = S(\mathcal V^{\ell-1}_{e_q,r_q}; \bm \theta^\ell)$}.
In the following two parts,
we introduce the 
sampling distribution and how to optimize the sampling parameters.

%we aim to parameterize the propagation path as 
%$\widehat{\mathcal G}^L(\bm \theta)$,
%and optimize $\bm \theta$ to capture the semantic dependency.
%Two parts should be designed to achieve this goal,
%i.e.,
%sampling distribution
%and learning strategy.
%\footnote{+yq+: the caption of two subsubsections may need to change}

\subsubsection{Parameterized sampling distribution}
\label{sssec:sampling-distribution}
For the GNN-based KG reasoning methods,
the entity representations {\small$\bm h_{e_o}^L$} in the final propagation step
are used to measure the plausibility of entity $e_o$ being the target answer entity.
In other words,
the entity representations are learned to indicate
their relevance to the given query $(e_q, r_q, ?)$.
To share this knowledge,
we introduce a linear mapping function
{$g\big(\bm h_{e_o}^\ell\!; {\bm \theta^\ell}\big) \!=\! {\bm \theta^\ell}^\top \bm h_{e_o}^\ell$}
%(a two-layer MLP)
with parameters $\bm \theta^\ell \!\in\! \mathbb R^d$ in each layer $\ell\!=\!1\dots L$,
and sample according to the probability distribution
\begin{equation}
p^\ell(e) := {\exp\big(g(\bm h_{e}^\ell; {\bm \theta^\ell})/\tau\big)}\Big/{\sum\nolimits_{e'\in\overline{\mathcal{V}}^\ell_{e_q,r_q}}\exp\big(g(\bm h_{e'}^\ell; {\bm \theta^\ell})/\tau\big)},
\label{eq:sampler}
\end{equation}
with temperature value {\small$\tau>0$}.

Sampling $K$ entities without replacement from $p^\ell(e)$
requires to sequentially sample and update the distribution for each sample,
which can be expensive.
We follow the Gumbel top-k trick~\cite{kool2019stochastic,xie2019reparameterizable} to solve this issue.
The Gumbel-trick firstly samples {\small $\big|\overline{\mathcal V}^\ell_{e_q,r_q}\big|$} independent noises from the uniform distribution,
i.e., $U_e\sim \text{Uniform}(0,1)$
to form the Gumbel logits $G_e=g(\bm h_{e}^\ell; {\bm \theta^\ell}) - \log(-\log U_e)$
for each {\small $e\in \overline{\mathcal V}^\ell_{e_q,r_q}$}.
%Rather than directly sample from distribution $p(e)$,
The top-$K$ entities in {\small $\overline{\mathcal{V}}^\ell_{e_q,r_q}$} are then collected
based on their values of Gumbel logits $G_e$.
As proved in  \cite{kool2019stochastic,xie2019reparameterizable},
this procedure is equivalent to sample without replacement from $p^\ell(e)$.

\subsubsection{Learning strategy}
\label{sssec:learning}
We denote $\bm \theta=\{\bm \theta^1, \dots, \bm \theta^L\}$ as the sampling parameters
and the parameterized propagation path as 
{\small $\widehat{\mathcal G}^L_{e_q,r_q}\!(\bm \theta)$}.
Since our sampling distribution in \eqref{eq:sampler}
is strongly correlated with the GNN representations,
jointly optimizing both the model parameters and sampler parameters
can better share knowledge between the two parts.
Specifically, we design the following objective
\begin{equation}
\bm w^*, \bm \theta^* = \arg\min_{\bm w,\bm \theta} 
\sum\nolimits_{(e_q,r_q, e_a)\in\mathcal Q_\text{tra}} \!\!\!\!
\mathcal L\big(F(\bm w, \widehat{\mathcal G}^L_{e_q,r_q}\!(\bm \theta)), e_a \big).
%	\mathcal L\big(F(\bm w, \widehat{\mathcal G}^L\!(\bm \theta)), \mathcal Q_{\text{tra}}\big)
\label{eq:opt_objective_samp}
\end{equation}
The loss function $\mathcal L$ on each instance is a binary cross entropy loss on all the entities {\small $e_o\in\mathcal V^L_{e_q,r_a}$}, i.e.,  
{\small \[
\!\!\mathcal L\big(F(\bm w, \widehat{\mathcal G}^L\!(\bm \theta)), e_a \big) \! =   -  \!\!\!\sum\nolimits_{e_o\in\mathcal{V}^L_{e_q,r_a}}\! y_{e_o}\text{log}(\phi_{e_o}) \!  + \!  (1 \! - \! y_{e_o})\text{log}(1 \! - \! \phi_{e_o}),
\]}
where the likelihood score {\small$\phi_{e_o}\!=\!f\big(\bm h^L_{e_o}; \bm w_\phi^\top\big)\in[0,1]$}
for entity $e_o$
indicating the plausibility of 
$e_o$ being the target answer entity,
and the label {$y_{e_o}\!=\!1$} if {$e_o\!=\!e_a$} otherwise 0.

The reparameterization trick is often used for Gumbel distributions \cite{jang2017categorical}.
However,
it does not do explicit sampling,
thus still requires high computing cost.
Instead,
we choose the straight-through (ST) estimator~\cite{bengio2013estimating,jang2017categorical},
which can approximately estimate gradient for discrete variables.
The key idea of ST-estimator is to back-propagate 
through the sampling signal
as if it was the identity function.
Specifically,
rather than directly use the entity representations $\bm h_{e}^\ell$
to compute messages,
we use
$\bm h_{e}^\ell  \!:=  \! \big( 1 \!  - \!  \texttt{no\_grad}(p^\ell\!(e))   +  p^\ell\!(e) \big) \cdot  \bm h_{e}^\ell$,
where $\texttt{no\_grad}(p^\ell\!(e))$ means that the back-propagation signals will not go through this term.
In this approach,
the forward computation will not be influenced, while
the backward signals can go through the probability multiplier $p^\ell\!(e)$.
An alternative choice to estimate the gradient on discrete sampling distribution is
the REINFORCE technique~\cite{williams1992simple,mnih2014neural}.
However,
REINFORCE gradient is known to have high variance~\cite{sutton2018reinforcement}
and the variance will be accumulated across the propagation steps,
thus not used here.
The model parameters $\bm w$ and sampler parameters $\bm \theta$
are simultaneously updated by Adam optimizer \cite{kingma2014adam}.

Overall,
the objective of learning semantic-aware distribution is to adaptively preserve the promising targets
according to the query relation.
As will be shown in Section~\ref{ssec:casestudy}.
the selected information in the sampled propagation path are semantically
related to the query relation
by sharing knowledge with the entity representations.
On the other hand,
the learned distribution guarantees a higher coverage of target entities
compared with the not learned version.

\subsection{The full algorithm}
\label{ssec:sample:full_algorithm}

\begin{algorithm}[!t]
	\caption{AdaProp: learning adaptive propagation path.}
	\label{alg:incremental}%{
		\small
		\begin{algorithmic}[1]
			\REQUIRE query $(e_q, r_q, ?)$,  $\mathcal{V}^{0}_{e_q,r_q}\!=\!\{e_q\}$, steps $L$, number of sampled entities $K$, 
			and functions $\text{MESS}(\cdot)$ and $\text{AGG}(\cdot)$.
			%		\STATE initialize $\bm{h}_{e_q}^{0}\!=\text{IND}(e_q, r_q, e_a)$ and the entity set ${\mathcal{V}}^{0}\!=\!\{e_q\}$;
			\FOR{$\ell=1\dots L$}
			\STATE get the neighboring entities $\mathcal N(\mathcal{V}^{\ell-1}_{e_q,r_q})=\cup_{e \in \mathcal V^{\ell-1}_{e_q,r_q}}\mathcal N(e)$,\\
			the newly-visited entities 
			$\overline{\mathcal{V}}^\ell_{e_q,r_q}  = \mathcal N(\mathcal{V}^{\ell-1}_{e_q,r_q}) \setminus \mathcal{V}^{\ell-1}_{e_q,r_q}$, and\\
			edges $\mathcal E^{\ell} = \{ (e_s, r, e_o) | e_s \in \mathcal{V}^{\ell-1}_{e_q,r_q}, e_o \in \mathcal N(\mathcal{V}^{\ell-1}_{e_q,r_q})\}$
			\label{step:inc:diff};
			
			\STATE obtain $\bm m_{(e_s, r, e_o)}^\ell \!:= \text{MESS}(\bm h_{e_s}^{\ell-1}, \bm h_{e_o}^{\ell-1}, \bm h_{r}^{\ell}, \bm h_{r_q}^{\ell})$ 
			for edges $(e_s, r, e_o)\in \mathcal E^{\ell}$;
			\label{step:inc:mess}
			
			\STATE obtain $\bm h_{e_o}^{\ell} \!:=\! \delta\big(\text{AGG}\big(\bm m_{(e_s, r, e_o)}^\ell, (e_s, r, e_o)\!\in\! \mathcal E^{\ell} \big)\big)$ 
			for entities $e_o \in \mathcal N(\mathcal{V}^{\ell-1}_{e_q,r_q})$;
			\label{step:inc:agg}
			
			\STATE
			\textbf{logits computation:} obtain the Gumbel logits $G_{e_o} \!=\! g(\bm h_{e_o}^\ell; {\bm \theta^\ell})\! -\!\log(\!-\!\log U_{e_o}\!)$ with $U_{e_o} \!\sim\! \text{Uniform}(0, 1)$
			for entities $e_o \!\in\! \overline{\mathcal{V}}^\ell_{e_q,r_q}$; \label{step:inc:gumb}

			\STATE 	
			\textbf{candidate sampling:} obtain sampled entities $\widetilde{\mathcal V}^\ell_{e_q,r_q} = \{\arg\text{top}_{K} G_{e_o}, e_o\in \overline{\mathcal{V}}^\ell_{e_q,r_q}\}$;
			\label{step:inc:sample}
			
			\STATE
			\textbf{straight-through:}
			$\bm h_{e}^\ell  \!:=  \! \big( 1 \!  - \!  \texttt{no\_grad}(p^\ell\!(e))   +  p^\ell\!(e) \big) \cdot  \bm h_{e}^\ell$ for entities $e\in\widetilde{\mathcal V}^\ell_{e_q,r_q}$;
			\label{step:inc:st}

			\STATE 
			\textbf{update propagation path:}
			update $\mathcal{V}^\ell_{e_q,r_q} = \mathcal{V}^{\ell-1}_{e_q,r_q} \cup \widetilde{\mathcal V}^\ell_{e_q,r_q}$;
			\label{step:inc:reform}

			\ENDFOR
			\RETURN $f\big(\bm h_{e_o}^L; \bm w^\top\big)$ for each $e_o\in\mathcal{V}^L_{e_q,r_q}$.
		\end{algorithmic}%}
	\vspace{-3px}
\end{algorithm}

The full procedure of AdaProp is shown in Algorithm~\ref{alg:incremental}.
Given the entities {\small$\mathcal V^{\ell-1}_{e_q,r_q}$} in the $(\ell  -  1)$-th step,
we first obtain the neighboring entities {\small$\mathcal N(\mathcal V^{\ell-1}_{e_q,r_q})$} of {\small$\mathcal V^{\ell-1}_{e_q,r_q}$}, and generate the newly-visited ones 
{\small$\overline{\mathcal{V}}^\ell_{e_q,r_q}=\mathcal N(\mathcal V^{\ell-1})\setminus \mathcal V^{\ell-1}_{e_q,r_q}$} 
as candidates in line~\ref{step:inc:diff}.
Then,
query-dependent
messages are computed on edges $\mathcal E^\ell$
in line~\ref{step:inc:mess},
and propagated to entities in {\small$\mathcal N(\mathcal V^{\ell-1}_{e_q,r_q})$} 
in line~\ref{step:inc:agg}.
The top-$K$ entities in {\small$\overline{\mathcal{V}}^\ell_{e_q,r_q}$}
are sampled in line~\ref{step:inc:sample}
according to the Gumbel logits computed in line~\ref{step:inc:gumb}.
As for the ST-estimator,
we modify the hidden representations for entities in 
{\small $\widetilde{\mathcal V}^\ell_{e_q,r_q}$}
in line~\ref{step:inc:st}.
Finally,
the set of sampled entities {\small$\widetilde{\mathcal{V}}^\ell_{e_q,r_q}$}
are concatenated with {\small${\mathcal{V}}^{\ell-1}_{e_q,r_q}$}
to form the set of $\ell$-th step entities {\small${\mathcal{V}}^{\ell}_{e_q,r_q}$}
in line~\ref{step:inc:reform}.
After propagating for $L$ steps,
the final step representations 
{\small$\bm h_{e_o}^L$} for 
all the involved entities are returned.
The entity 
with the highest score evaluated by {\small$f(\bm h_{e_o}^L;\bm w_\phi)$} in {\small$\mathcal V^L_{e_q,r_q}$}
is regarded as the predicted target.

The key difference compared with existing GNN-based KG reasoning methods is that 
the propagation path in Algorithm~\ref{alg:incremental}
is no longer given and fixed before the message propagation.
Instead,
it is adaptively adjusted based on the entity representations
and 
is dynamically updated in each propagation step.
In addition,
Algorithm~\ref{alg:incremental} also enjoys higher efficiency than the
full propagation and progressive propagation methods,
since only a small portion of entities are involved,
while the extra computation cost for sampling is relatively low.
Compared with constrained propagation methods,
which needs to propagate in the answer-specific propagation paths for multiple times,
the potential target answer entities are directly scored in the final propagation step
in a single forward pass.

\section{Experiments}
\label{sec:exp}

In this section, we empirically verify the effectiveness 
and analyze the components of the proposed AdaProp
on both transductive and inductive settings.
All the experiments are implemented in Python 
with PyTorch~\cite{paszke2017automatic}
and run on a single NVIDIA RTX 3090 GPU with 24GB memory.

\subsection{Comparison with KG reasoning methods}

\noindent
We compare AdaProp with general KG reasoning methods
in both transductive and inductive reasoning.
In the transductive setting,
the entities sets are the same for training and testing.
While in the testing of inductive reasoning,
the model are required to generalize to unseen entities w.r.t. the training set.
%Denote training set as $\mathcal K_{tra} \! = \! \{\mathcal V_{tra}, \mathcal R_{tra}, \mathcal E_{tra}, \mathcal Q_{tra} \}$
%and test set as $\mathcal K_{tst} \! = \! \{\mathcal V_{tst}, \mathcal R_{tst}, \mathcal E_{tst}, \mathcal Q_{tst} \}$.
%
%In the transductive setting,
%the training set and test set  share the same set of entities and relations,
%i.e., 
%all the entities and relations in the test set are seen in the training set.
%%
%In the inductive setting,
%the training and test set share the same set of relations but different entities,
%namely, 
%the entities in the test set are not seen during training.
%
Besides,
we follow~\cite{bordes2013translating,teru2019inductive,wang2017knowledge,zhu2021neural,zhang2021knowledge}
to use the filtered ranking-based metrics for evaluation,
i.e., mean reciprocal ranking (MRR) and Hit@$k$
(e.g., Hit@$1$ and Hit@$10$).
The higher value indicates the better performance for both metrics.
As for the hyper-parameters,
we tune the number of
propagation steps $L$ from $5$ to $8$,
the number of sampled entities $K$  in $\{100, 200, 500, 1000, 2000\}$,
and the temperature value $\tau$ in $\{0.5, 1.0, 2.0\}$.
%The details of other hyper-parameters are listed in Appendix \ref{app:HP}.
The ranges of other hyper-parameters are kept the same as RED-GNN~\cite{zhang2021knowledge}.

\subsubsection{Transductive setting}
\label{ssec: transductive}

%\footnote{+qm+ better write as two separate parts,
	%	since their setup / methods / datasets are all different.}

\noindent \\
Figure~\ref{fig:kg} shows an example of transductive KG reasoning,
where a KG related to \textit{Lakers} is given
and the task is to predict where \textit{LeBron} lives in based on the given facts.
Formally speaking,
we have one KG $\mathcal G = (\mathcal V, \mathcal R, \mathcal E, \mathcal Q)$
where the query set $\mathcal Q$ is split into three disjoint sets
$\mathcal Q_\text{tra}/\mathcal Q_\text{val}/\mathcal Q_\text{tst}$
for training, validation and testing respectively.

%the training set and test set share the same set of entities and relations,
%i.e., 
%$\mathcal V_{tra} = \mathcal V_{tst}$ and $\mathcal R_{tra} = \mathcal R_{tst}$.

\vspace{2px}
\noindent
\textbf{Datasets.} 
We use six widely used KG completion datasets,
including 
{\sf Family}~\cite{kok2007statistical},
{\sf UMLS}~\cite{kok2007statistical},
{\sf WN18RR}~\cite{dettmers2017convolutional}, 
{\sf FB15k237}~\cite{toutanova2015observed},
{\sf NELL-995}~\cite{xiong2017deeppath}
and
{\sf YAGO3-10}~\cite{suchanek2007yago}.
The statistics of these datasets are provided in Table~\ref{tab:dataset}.

\begin{table}[ht]
	\centering
%	\vspace{-6px}
	\caption{Statistics of the transductive KG datasets. 
		Fact triplets in $\mathcal E$ are used to build the graph,
		and
		$\mathcal Q_\text{tra}$,
		$\mathcal Q_\text{val}$,
		$\mathcal Q_\text{tst}$
		are the query triplets used for reasoning.}
		\vspace{-12px}
	\label{tab:dataset}
	\setlength\tabcolsep{3.2pt}
	\begin{tabular}{c|cc|cccc}
		\toprule
		dataset                 & \#entity & \#relation &  $|\mathcal E|$  & $|\mathcal Q_\text{tra}|$ & $|\mathcal Q_\text{val}|$ & $|\mathcal Q_\text{tst}|$  \\ \midrule
		{\sf Family} & 3.0k &  12 &  23.4k &5.9k & 2.0k  & 2,8k \\
		{\sf UMLS} & 135 &  46 &  5.3k & 1.3k & 569 &  633 \\
		{\sf WN18RR}  &  40.9k  &     11     &  65.1k   &  21.7k	&  3.0k  & 3.1k    \\
		{\sf FB15k237}  &  14.5k  &    237     &  204.1k  &	68.0k	& 17.5k  & 20.4k   \\
		{\sf NELL-995}  &  74.5k  &    200     &  112.2k &	37.4k & 543  & 2.8k   \\
		{\sf YAGO3-10}   &   123.1k  &  37  & 809.2k & 269.7k  &  5.0k  & 5.0k \\
		\bottomrule
	\end{tabular}
\vspace{-10px}
\end{table}

\begin{table*}[ht]
	\centering
%	\vspace{-5px}
	\caption{Transductive setting. 
		Best performance is indicated by the bold face numbers,
		and the underline means the second best. 
		``--'' means unavailable results.
		``H@1'' and ``H@10'' are short for Hit@1 and Hit@10 (in percentage), respectively.}
	\fontsize{8}{10}\selectfont
	\setlength\tabcolsep{2.5pt}
	\label{tab:transd}
	\vspace{-12px}
	\begin{tabular}{cc|ccc|ccc|ccc|ccc|ccc|ccc}
		\toprule
		\multirow{2}{*}{type}  &   \multirow{2}{*}{models}  &  \multicolumn{3}{c|}{{\sf Family}}   &  \multicolumn{3}{c|}{{\sf UMLS}}  &  \multicolumn{3}{c|}{{\sf WN18RR}}   &  \multicolumn{3}{c|}{{\sf FB15k237}}   &  \multicolumn{3}{c|}{{\sf NELL-995}}    &  \multicolumn{3}{c}{{\sf YAGO3-10}}      \\
		&  &  MRR     & H@1     & H@10   &	MRR     & H@1     & H@10   &MRR     & H@1     & H@10   &MRR     & H@1     & H@10   &MRR     & H@1     & H@10   &MRR     & H@1     & H@10   
		\\   \midrule
		\multirow{5}{*}{non-GNN} 
		& ConvE   & 0.912 & 83.7 & 98.2   & 0.937  & 92.2 &  96.7  &  0.427 &  39.2 &  49.8 & 0.325 & 23.7 & 50.1 & 0.511  & 44.6 & 61.9  & 0.520 & 45.0 &  66.0	\\
		&  QuatE    	& 0.941 &89.6 &99.1 & 0.944 &  90.5 &  99.3 & 0.480&44.0&55.1&	0.350&25.6&53.8	&	0.533 & 46.6 & 64.3    & 0.379 & 30.1 & 53.4 \\ 
		& RotatE &  0.921 & 86.6 & 98.8 & 0.925 & 86.3 & 99.3 &  0.477 & 42.8 & 57.1 & 0.337 & 24.1 & 53.3 & 0.508 & 44.8 & 60.8 & 0.495 & 40.2 & 67.0 \\
		&  MINERVA   & 0.885 & 82.5 & 96.1 & 0.825 & 72.8 & 96.8  &  0.448&41.3&51.3	&	0.293&21.7&45.6& 0.513&41.3&63.7   &  -- & -- & --   \\ 
		&  DRUM   & 0.934 & 88.1 & \underline{99.6} & 0.813 & 67.4 & 97.6  & 0.486 & 42.5 & 58.6	&	0.343&25.5&51.6	& 0.532 &  46.0 & 66.2    & 0.531 &  45.3 & 67.6    \\ 
		&  RNNLogic &	0.881  & 85.7   & 90.7  & 0.842 & 77.2 & 96.5  &	0.483& 44.6&55.8	&	0.344&25.2&53.0	&	0.416  &  36.3  & 47.8     & 0.554 &  50.9 & 62.2    \\ 
		& RLogic & -- & -- & -- & -- & -- & -- & 0.47  & 44.3 & 53.7 & 0.31 & 20.3 & 50.1 & -- & -- & -- & 0.36  & 25.2  & 50.4 \\
		\midrule
		\multirow{4}{*}{GNNs}  
		&  CompGCN  & 0.933 &  88.3 &  99.1 & 0.927 & 86.7 & \underline{99.4}  &  0.479&44.3&54.6	&	0.355&26.4&53.5	&	0.463  &  38.3	 & 	59.6   &  0.421 	&  39.2	 &	57.7    \\
		& NBFNet   &  0.989  &  \textbf{98.8}  &   98.9  &  0.948  &  92.0  &  \textbf{99.5} & \underline{0.551}   &  \underline{49.7}  &  \underline{66.6}  & \underline{0.415}  &  \underline{32.1}  &  \textbf{59.9}  &  0.525   &    45.1   &   63.9     &  0.550	& 47.9	& 68.6   \\
		&  RED-GNN   & \textbf{0.992} & \textbf{98.8}   &  \textbf{99.7}   &  \underline{0.964}   &  \underline{94.6}  &  {99.0}  &  0.533 & 48.5 & 62.4& 0.374  &28.3& 55.8  &  \underline{0.543} & \underline{47.6} &  \underline{65.1}  &   0.559		&  48.3  & 68.9    \\
		\cmidrule{2-20}
		& \textbf{AdaProp}  &	\underline{0.988}  & {98.6}  & 99.0  &  \textbf{0.969} & \textbf{95.6} &  \textbf{99.5}   &  \textbf{0.562}  & \textbf{49.9}  &  \textbf{67.1}  &  \textbf{0.417}  &  \textbf{33.1} &  \underline{58.5}   &  \textbf{0.554}  & \textbf{49.3}  &  \textbf{65.5}  &  \textbf{0.573}  & \textbf{51.0}  &  \textbf{68.5}  \\ 
		\bottomrule
	\end{tabular}
\vspace{-4px}
\end{table*}

\vspace{2px}
\noindent
\textbf{Baselines.}
We compare the proposed AdaProp with 
(i) non-GNN methods
ConvE~\cite{dettmers2017convolutional}, QuatE~\cite{zhang2019quaternion},
RotatE~\cite{sun2019rotate},
MINERVA~\cite{das2017go},
DRUM~\cite{sadeghian2019drum},
RNNLogic~\cite{qu2021rnnlogic} and
RLogic~\cite{cheng2022rlogic};
and 
(ii)
GNN-based
full propagation method
CompGCN~\cite{vashishth2019composition}
and progressive propagation methods
NBFNet~\cite{zhu2021neural}
and
RED-GNN~\cite{zhang2021knowledge} here.
R-GCN~\cite{schlichtkrull2018modeling}
is not compared here as it is much inferior to CompGCN~\cite{vashishth2019composition}.
As indicated by \cite{zhang2021knowledge,zhu2021neural},
GraIL~\cite{teru2019inductive}
and CoMPILE~\cite{mai2021communicative}
are intractable to work on the transductive KGs with many entities,
and thus not compared in this setting.
Results of these baselines are taken from their papers
or reproduced by their official codes.
Missing results of RLogic are due to the lack of source code.

\vspace{2px}
\noindent
\textbf{Results.}
The comparison of AdaProp with the transductive reasoning methods
is in Table~\ref{tab:transd}.
First,
The GNN-based methods generally perform 
better than the non-GNN based ones by capturing 
both the structural and semantic information in KGs.
Second,
the progressive propagating methods,
i.e.,  NBFNet and RED-GNN,
perform better than the full propagation method CompGCN,
especially on larger KGs,
like {\sf WN18RR}, {\sf FB15k237}, {\sf NELL-995}
and {\sf YAGO3-10}.
In comparison,
AdaProp achieves leading performance
on {\sf WN18RR}, {\sf NELL-995} and {\sf YAGO3-10}
over all the baselines
and
slightly outperforms NBFNet on {\sf FB15k237}.
Besides,
AdaProp also works well 
with competitive performances on smaller KGs {\sf Family} and {\sf UMLS}.
To further investigate the superiority of propagation path learned by AdaProp,
we analyze the properties in the following content.

\vspace{2px}
\noindent
\textbf{Property of propagation path.}
To quantitatively analyze the property of propagation path,
we introduce the following metric, i.e., the ratio of target over entities (\texttt{ToE}),
to indicates the property of {\small $\widehat{\mathcal G}^L_{e_q,r_q}$} 
with propagation steps $L$:
\begin{equation}
	\texttt{ToE}(L) = \texttt{TC}(L) \big/\texttt{EI}(L),
	\label{eq:factor}
\end{equation}
where the nominator 
$\texttt{TC}(L) = \frac{1}{|\mathcal Q_\text{tst}|}\sum\nolimits_{(e_q,r_q,e_a)\in \mathcal Q_\text{tst}}\mathbb I\big\{e_a \in \mathcal{V}^{L}_{e_q,r_q}\big\}$
measures the ratio of \textit{target entity coverage}
by the propagation path,
and the denominator
$\texttt{EI}(L) = \frac{1}{|\mathcal Q_\text{tst}|}\sum\nolimits_{(e_q,r_q,e_a)\in \mathcal Q_\text{tst}}$ $\big|\bigcup_{\ell=1\dots L}\mathcal{V}^\ell_{e_q,r_q}\big|$
measures the number of \textit{entities involved} in the propagation path. 
{$\mathcal Q_\text{tst}$} contains queries in the testing set,
{$\mathcal{V}^\ell_{e_q,r_q}$} is the set of $\ell$-th step entities 
for query $(e_q,r_q,?)$,
and
{$\mathbb I\{e_a \in \mathcal{V}^{L}_{e_q,r_q}\}=1$} 
if $e_a$  is in $\mathcal V^L_{e_q,r_q}$
otherwise 0.

\begin{figure}[ht]
	\centering
	\vspace{-6px}
	\subfigure[$\texttt{ToE}(L)$]
	{\includegraphics[height=3.9cm]{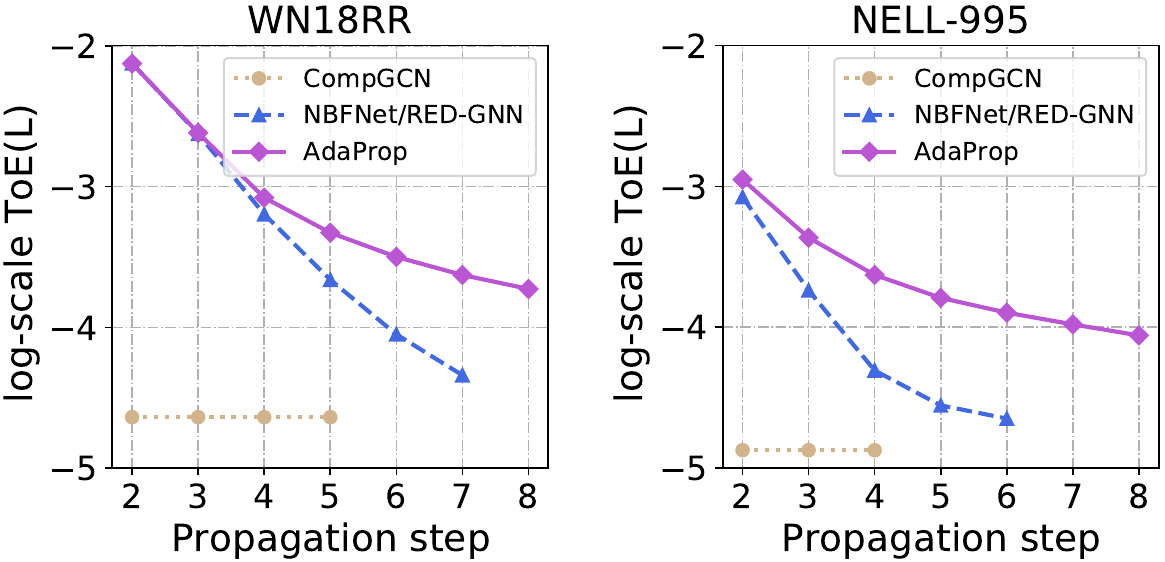}}
	\vspace{-5px}
	
	\subfigure[Testing performance]
	{\includegraphics[height=3.9cm]{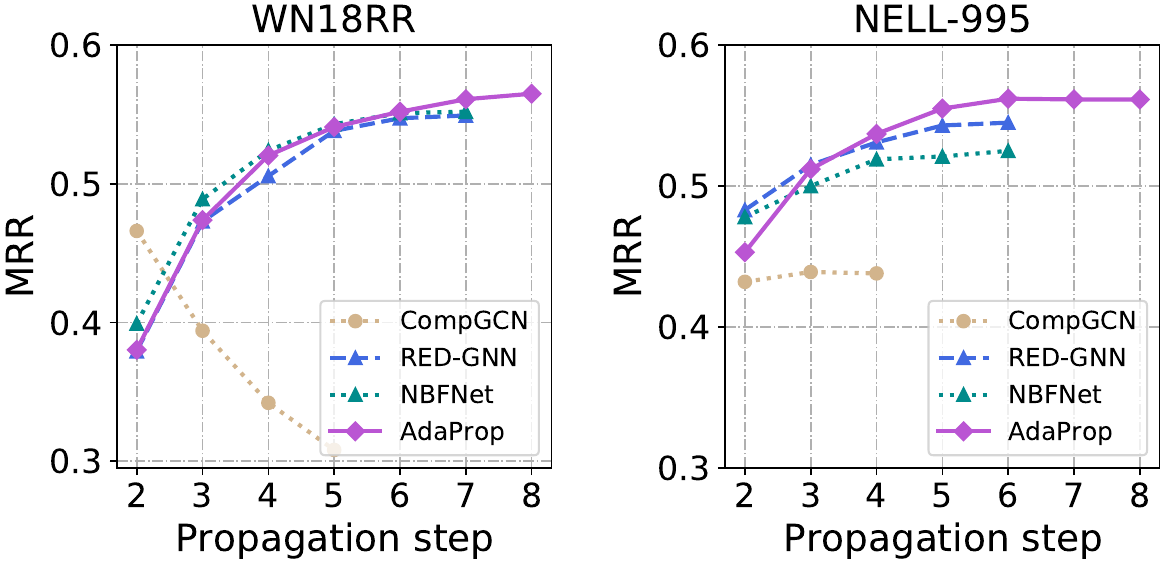}}
	\vspace{-12px}
	\caption{Comparison of GNN-based methods w.r.t. $L$.}
	\label{fig:ablation_steps}
	\vspace{-10px}
\end{figure}

We compare 
the value $\texttt{ToE}(L)$ and
the MRR performance of different GNN-based reasoning methods
with different number of propagation steps $L$ in 
Figure~\ref{fig:ablation_steps}.
{\sf WN18RR} and {\sf NELL-995} are used here.
As shown,
CompGCN quickly decreases on {\sf WN18RR} and does not gain on {\sf NELL-995} when $L \!>\! 2$
since its $\texttt{ToE}(L)$ is fixed and very small.
It will easily suffer from the problem of over-smoothing \cite{oono2019graph}
and quickly run out-of-memory when $L$ increases.
By contrast,
RED-GNN and NBFNet have increasing performance
from $L\!=\!2$ to $L\!=\!5$ by capturing the more long-range information.
But they get stuck in $L\!>\!5$ when too many entities are involved.
Besides,
RED-GNN and NBFNet, with the same design of propagation path,
have similar performance curves here.
%GraIL and CoMPILE also improve when $L=2\dots 4$,
%but will decrease when $L>4$ as the growing size of the enclosing subgraph.
%Due to scalability issue,
%results of CompGCN when $L>5$
%and GraIL and CoMPILE when $L>6$ are not shown.
As for AdaProp,
the performance consistently gets improved with 
larger $\texttt{ToE}(L)$ in
deeper propagation steps.
The benefits of a deeper propagation path
can be attributed to both a larger propagation step
and a deeper GNN model (we provide more details in Appendix~\ref{sssec:long-range}).
AdaProp, with larger $\texttt{ToE}(L)$ at larger $L$,
alleviates the problem of over-smoothing compared with other methods.

%We have the following observations:
%\begin{itemize}[leftmargin=20px, itemsep=0pt,topsep=0pt,parsep=0pt,partopsep=0pt]
%	\item[(i).]
%	The deeper propagation tends to achieve better performance
%	since the long-range information can be captured with a deeper model
%	(see Progressive propagation when $L=1\dots 5$).
%	\item[(ii).]
%	The key factor preventing existing methods to going deeper is the 
%	small $\texttt{ToE}(L)$
%	%		the large $\texttt{EI}(L)$ and small $\texttt{ToC}(L)$
%	(see Full propagation when $L\geq 2$ and Progressive propagation when $L\geq 5$).
%	\item[(iii).]
%	The potential to going deeper with better performance
%	is to control the decreasing trend of $\texttt{ToE}(L)$
%	% reduce $\texttt{EI}(L)$ while guaranteeing large $\texttt{ToC}(L)$
%	(see Adaptive propagation v.s. Progressive propagation when $L\geq 5$).
%\end{itemize}

%In addition, 
%the hyper-parameter configurations are provided in Appendix~\ref{app:HP},
%showing that
%AdaProp indeed learns with deeper propagation paths
%than the other GNN-based reasoning methods.

\begin{table*}[ht]
	\centering
	\caption{Inductive setting (evaluated with Hit@10).
		Full evaluation results of MRR and Hit@1 are in Appendix~\ref{app:induc_full}.}
	\label{tab:induc}
	\small
	\setlength\tabcolsep{8pt}
%	\fontsize{8}{10}\selectfont
	\vspace{-12px}
	\begin{tabular}{cc|cccc|cccc|cccc}
		\toprule
		\multirow{2}{*}{metric} &  \multirow{2}{*}{methods}   & \multicolumn{4}{c|}{{\sf WN18RR}} & \multicolumn{4}{c|}{{\sf FB15k237}}  & \multicolumn{4}{c}{{\sf NELL-995}} \\
		&          & V1    & V2   & V3   & V4   & V1    & V2    & V3    & V4    & V1  & V2  & V3  & V4  \\   
		%		\midrule 
		\midrule
%		\multirow{7}{*}{H@1}
%		& RuleN   &  63.5   &   61.1  &  34.7  &  59.2  &  \textbf{30.9}  & 34.7 & 34.5 & 33.8  & \textbf{54.5}&30.4& 30.3  & \underline{24.8}\\
%		&  Neural LP &	59.2	&	57.5	&	30.4	&	58.3	&	24.3	&	28.6	&	30.9	&	28.9	&	50.0	&		24.9   &	26.7 	& 13.7	\\
%		& DRUM      &  61.3  &   59.5   &    33.0  &  58.6    &  24.7     &   28.4    &   30.8    &   30.9    & 50.0  &  27.1  &  26.2  &   16.3  \\  \cline{2-14}
%		& GraIL     &   55.4    &    54.2  &   27.8   &    44.3  &   20.5    &   20.2    &   16.5    &    14.3   &  42.5  & 19.9  &   22.4  &  15.3  \\ 
%		& CoMPILE  &    47.3   &  48.5   &   25.8  & 47.3     &    20.8   &  17.8     &   16.6    &    13.4   &   10.5 &   15.6  &  22.6   & 15.9   \\ 
%		& {RED-GNN}      &  \underline{65.3}     &   \underline{63.3}   &  \underline{36.8}    &   \underline{60.6}   &  \underline{30.2}     &   \underline{38.1}    &   \underline{35.1}    &  \underline{34.0}   & \textbf{52.5}   &  \underline{31.9}  &  \textbf{34.5}  &  \textbf{25.9}  \\    
%		& NBFNet     &    59.2   &   57.5   &   30.4   &  57.4    &  19.0     &   22.9    &   20.6  &    18.5  &  50.0 & 27.1 &  26.2  & 23.3  \\
%		\cline{2-14}
%		&    \textbf{AdaProp} &   \textbf{66.8}   &  \textbf{64.2}  & \textbf{39.6}   &    \textbf{61.1}	 &  {19.1} & \textbf{37.2}  &  \textbf{37.7} &   \textbf{35.3}	&  52.2 &  \textbf{34.4}  & \underline{33.7}  &  24.7   \\
%		\midrule
		\multirow{8}{*}{Hit@10 (\%)}
		& RuleN   &  73.0   &  69.4  &  40.7   &  68.1  & 44.6  & 59.9 & 60.0  &  60.5  & 76.0&51.4&53.1&48.4  \\
		&  Neural LP &	77.2	&	74.9 &	47.6	&	70.6	&	46.8	&	58.6	&	57.1	&	59.3	&	{87.1}	&		56.4   &	{57.6}	&53.9 \\
		& DRUM      &  77.7  &  74.7    &   47.7   &   70.2   &   {47.4}    &  59.5     &   {57.1}    &  59.3 &  \underline{87.3}  &  {54.0}  & 57.7  &  {53.1}  \\ 
				\cmidrule{2-14}
	 & GraIL     &   76.0    &  77.6    &   40.9   &   68.7   &  42.9     &    42.4   &   42.4    &   38.9     &  56.5 & 49.6 & 51.8  &  50.6  \\
		& CoMPILE  &    74.7   &   74.3  &  40.6   &   67.0   &   43.9    &    45.7   &     44.9  &   35.8    &  57.5  &  44.6   &  51.5   &  42.1  \\ 
		& NBFNet     &   \underline{82.7}   &   \underline{79.9}   &   \underline{56.3}   &  70.2    &   \underline{51.7}     &   \underline{63.9}     &    58.8    &    55.9     &     79.5 	 &  	\underline{63.5}	 &   \underline{60.6}   &  	\underline{59.1}  \\
		& {RED-GNN}      &   {79.9}    &  {78.0}    &   {52.4}   &  \underline{72.1}    &   {48.3}    &   {62.9}    &  {60.3}     &  \underline{62.1}   &  {86.6} &  {60.1}  &  {59.4}  &  {55.6} \\ 	
		\cmidrule{2-14}
		&    \textbf{AdaProp} &  \textbf{86.6}    &  \textbf{83.6}  &   \textbf{62.6} &     \textbf{75.5}	&  \textbf{55.1}  &  \textbf{65.9} &  \textbf{63.7}   &   	\textbf{63.8}	&  \textbf{88.6}  &  \textbf{65.2} & \textbf{61.8} &   \textbf{60.7}  \\
%		& std &   &   &  &   &   &   &   &    &   &   &   &   \\  
		\bottomrule
	\end{tabular}
\vspace{-4px}
\end{table*}

%and is the runner up in FB15k237.
%Generally,
%the GNN models display
%prominent superiority
%over the other two types,
%and the query-dependent propagation methods, 
%e.g., DPMPN, RED-GNN, NBFNet and AdaProp
%are obviously better than the other methods.
%Even though the entity embeddings are not used in 
%RED-GNN, NBFNet, and AdaProp,
%they outperform all the triple-based,
%path-based approaches by a large margin.
%This demonstrates the superiority in learning
%complexity topology patterns in KGs with the query-dependent propagation.
%It can be found that 
%in large and sparse KGs, 
%e.g., the WN18RR,
%the improvement is more significant.
%In FB15k237, the denser KG,
%AdaProp is slightly worse than NBFNet
%but has significant improvement over RED-GNN.
%Note that NBFNet is very resource-demanding,
%which runs on four NVIDIA V100 GPUs with 128 GB memory in total.
%In comparison,
%AdaProp only works on a single GPU with 24 GB memory.
%If sufficient resources like those in NBFNet are provided,
%AdaProp can expect further improvement.

\subsubsection{Inductive setting}

\noindent \\
The inductive setting cares about reasoning on unseen entities.
If the KG related to \textit{Lakers} in Figure~\ref{fig:kg} is used for training,
then the testing can be reason on another KG related to \textit{Warriors}
and a query \textit{(Curry, lives\_in,?)} with unseen entity \textit{Curry}.
Formally speaking,
we have two KGs
$\mathcal G_1 = (\mathcal V_1, \mathcal R, \mathcal E_1, \mathcal Q_1)$
and 
$\mathcal G_2 = (\mathcal V_2, \mathcal R, \mathcal E_2, \mathcal Q_2)$.
$\mathcal G_1$ is used for training and validation
by splitting $\mathcal Q_1$ into two disjoint sets $\mathcal Q_\text{tra}$ and $\mathcal Q_\text{val}$,
and 
the unseen KG $\mathcal G_2$ is used for testing with $\mathcal Q_\text{tst}:=\mathcal Q_2$.
The relation sets of the two KGs are the same,
while the entity sets are disjoint, 
i.e.,  $\mathcal V_{1}  \cap \mathcal V_{2} \! = \! \emptyset$.

\vspace{2px}
\noindent
\textbf{Datasets.}
Following~\cite{teru2019inductive, zhang2021knowledge},
we use the same subsets (4 versions each and 12 subsets in total) of {\sf WN18RR}, {\sf FB15k237} and {\sf NELL-995},
where each subset has a different split of training set and test set.
We refer the readers to~\cite{teru2019inductive, zhang2021knowledge}
for the detailed splits and statistics.
%The scales of the inductive reasoning datasets are much smaller
%than those in the transductive settings.

\vspace{2px}
\noindent
\textbf{Baselines.} 
All the reasoning methods which learn entity embeddings in training 
(ConvE~\cite{dettmers2018convolutional}, QuatE~\cite{zhang2019quaternion},
RotatE~\cite{sun2019rotate}, 
MINERVA~\cite{das2017go}, CompGCN~\cite{vashishth2019composition})
cannot work in this setting.
Hence,
we compare AdaProp with non-GNN methods that learn rules
without entity embeddings,
i.e., RuleN~\cite{meilicke2018fine},  NeuralLP~\cite{yang2017differentiable} and DRUM~\cite{sadeghian2019drum}.
For GNN-based methods,
we take GraIL~\cite{teru2019inductive},
CoMPILE~\cite{mai2021communicative},
RED-GNN~\cite{zhang2021knowledge}
and NBFNet~\cite{zhu2021neural}
as baselines.
RNNLogic \cite{qu2021rnnlogic} and RLogic \cite{cheng2022rlogic}
are not compared here
since there lack source codes and reported results in this setting.
We follow \cite{zhang2021knowledge} to
evaluate the ranking of target answer entity over all the negative entities,
rather than 50 randomly sampled negative ones adopted in~\cite{teru2019inductive}.

%We compare AdaProp
%with several state-of-the-arts models
%in the category of triple-based, path-based, and GNN-based.
%In the transductive setting,
%we select three triple-based embedding models,
%ConvE~\cite{dettmers2017convolutional}, RotatE~\cite{sun2019rotate}, and QuatE~\cite{zhang2019quaternion};
%four latest path-based approaches,
%MINERVA~\cite{das2017go}, NeuralLP~\cite{yang2017differentiable}, DURUM~\cite{sadeghian2019drum}, and RNNLogic~\cite{qu2021rnnlogic};
%and four up-to-date GNN-based methods, 
%i.e., CompGCN~\cite{vashishth2019composition}, DPMPN~\cite{xu2019dynamically}, RED-GNN~\cite{zhang2021knowledge}, and NBFNet~\cite{zhu2021neural}.
%As for the inductive setting,
%we also add RuleN~\cite{meilicke2018fine} and GraIL~\cite{teru2019inductive} as baselines for comparison.
%Note that the GraIL is not compared in the transductive setting 
%for its high computation cost 
%brought by the expensive offline sampling. 
%In addition,
%those methods that learn entity-dependent representations,
%e.g., 
%the triple-based,
%MINERVA, DPMPN, and CompGCN,
%can merely work in the transductive setting,
%and thus they are only compared in the inductive setting.

\vspace{2px}
\noindent
\textbf{Results.} 
%
%The results of the Hit@10 metric (MRR and Hit@1 in Appendix~\ref{app:induc_full})
%are summarized in Table~\ref{tab:induc}.
As shown in Table~\ref{tab:induc},
the constrained propagation methods GraIL and CoMPILE,
even though have the inductive reasoning ability,
are much worse than the progressive propagation methods
RED-GNN and NBFNet.
The patterns learned in the constrained range in GraIL and CoMPILE 
are not well generalized to the unseen KG.
AdaProp consistently performs the best or second best
in different datasets with all the split versions.
This demonstrates that the learned adaptive propagation paths
can be generalized to a new KG where entities are not seen during training,
resulting in strong inductive reasoning ability.

\subsubsection{Running time}
\noindent
\\
An advantage of AdaProp is that it has less computing cost due to the reduced number of entities.
%In the largest KG {\sf YAGO3-10},
%the existing GNN-based methods will run out-of-memory,
%while AdaProp,
%with less memory cost achieves state-of-the-art performance.
In this part,
we compare the running time of different methods.

\begin{figure}[ht]
	\centering
%	\vspace{-4px}
	\includegraphics[height=3.9cm]{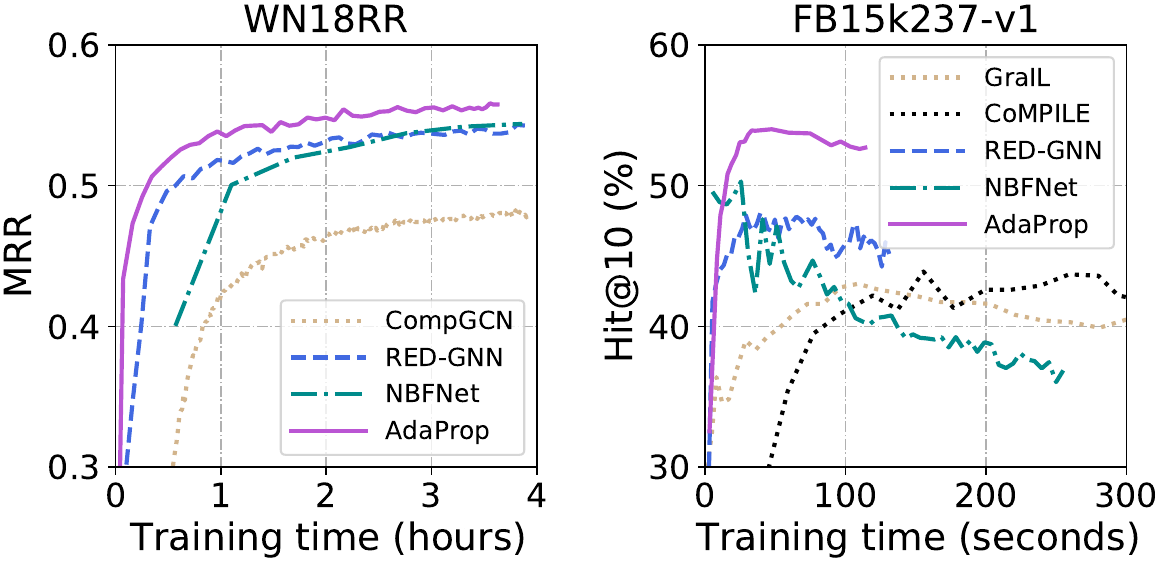} 
	\vspace{-10px}
	\caption{Learning curves of different GNN-based methods.}
	\vspace{-8px}
	\label{fig:ablation_time}
\end{figure}

We show the learning curves of 
different GNN-based methods 
on transductive data {\sf WN18RR}
and inductive data {\sf FB15k237-v1}
in Figure~\ref{fig:ablation_time}.
First,
the full propagation method CompGCN is the slowest method
in transductive setting
since it has the largest number of involved entities.
Second,
the constrained propagation methods,
i.e., GraIL and CoMPILE,
are slow in inductive setting,
since their process of subgraph extraction
is expensive.
CoMPILE with the more complex message passing functions
is even slower than GraIL.
Finally,
comparing AdaProp with the 
progressive propagation methods
NBFNet and RED-GNN,
the learning curve of AdaProp grows faster since
it propagates with much fewer entities
and the sampling cost is not high.
Hence,
the adaptive sampling technique 
is not only effective in both reasoning setting,
but also efficient in propagating within fewer entities.

%\begin{figure*}[t]
%	\centering
%	\vspace{-4px}
%	\subfigure[Property of propagation paths with target coverage over entities involved ($\texttt{ToE}(L)$).]
%	{\includegraphics[width=15.5cm]{figures/propagation-depth-analysis-v6-X}}
%	\subfigure[Performance with MRR metric.]
%	{\includegraphics[width=15.5cm]{figures/propagation-depth-analysis-v6-MRR}}
%	\vspace{-14px}
%	\caption{Properties and performance of different propagation depth $L$.}
%	%	\label{fig:propagation_stats}
%	\label{fig:propagation_path_properties}
%	\vspace{-6px}
%\end{figure*}

\subsection{Understanding the sampling mechanism}
\label{exp:understanding}

In this section,
we conduct a thorough ablation study
via evaluating the sampling mechanism from 
four perspectives, i.e.,
%the effectiveness of deeper propagation steps,
the comparison on alternative sampling strategies, 
the importance of learnable sampling distribution,
the comparison on different learning strategies,
and the influence of sampled entities $K$.
The evaluation is based on
a transductive dataset {\sf WN18RR}
with MRR metric
and an inductive dataset {\sf FB15k237-v1}
 with Hit@10 metric.

%\subsubsection{The influence of long-range information and a deeper model}

%\begin{table}[ht]
%	\centering
%	\caption{Hit@10 of AdaProp with different GNN layers and propagation depth on FB15k237-v1.}
%	\label{tab:decouple2}
%	\vspace{-10px}
%	\setlength\tabcolsep{1.8pt}
%	\begin{tabular}{cc|cc|cc|cc}
%		\toprule
%	\end{tabular}
%\end{table}

%\begin{table}[ht]
%	\centering
%	\caption{The performance (MRR) of AdaProp with different GNN layers and propagation depth on FB15k237.}
%	\label{tab:decouple2}
%	\vspace{-10px}
%	%	\setlength\tabcolsep{1.8pt}
%	\begin{tabular}{cc|cc|cc}
%		\toprule
%	 depth & layers & 2-hop & $\uparrow$ per row & 3-hop & $\uparrow$ per row
%	 	\\  \midrule
%		3 	&	3 		& 	0.351	& -	& 	0.202	&  	-	\\
%		3   &	4 		&  0.364	& 0.013	& 	0.228	& 	0.026	 \\
%		4  	&	4		&	0.406	& 0.042	& 0.279	&  0.051  \\
%		\bottomrule
%	\end{tabular}
%\end{table}

\subsubsection{Sampling strategy}
\label{sssec:exp:sampling}
\noindent \\
In this part,
we compare the sampling strategies (with design details in Appendix~\ref{app:sampling-for-kg}) discussed in Section~\ref{ssec:sampling},
i.e., node-wise sampling,
layer-wise sampling
and subgraph sampling,
and the proposed incremental sampling strategy
with metrics $\texttt{EI}(L)$, $\texttt{ToE}(L)$ in Eq.\eqref{eq:factor},
and testing performance.
For a fair comparison,
we guarantee that the different variants 
have the same number of propagation steps $L$
and the similar amount of entities involved,
i.e., $\texttt{EI}(L)$.

As shown in Table~\ref{tab:sampling-methods-comparison},
the incremental sampling is better than the other sampling strategies
in both the learned and not learned settings.
The success is due to the larger $\texttt{ToE}(L)$
by preserving the previously sampled entities.
The node-wise and layer-wise sampling perform
much worse since the sampling there is uncontrolled.
As the target entities lie close to the query entities,
the subgraph sampling is relatively better than
the node-wise and layer-wise sampling
by propagating within the relevant local area around the query entity.

\begin{table}[ht]
	\centering
	\vspace{-5px}
	\caption{Comparison of different sampling methods.}
	\label{tab:sampling-methods-comparison}
	\setlength\tabcolsep{2.3pt}
	\small
	\vspace{-10px}
	\begin{tabular}{cc|ccc|ccc}
		\toprule
		\multirow{2}{*}{learn}                                                 & \multirow{2}{*}{methods} & \multicolumn{3}{c|}{WN18RR} & \multicolumn{3}{c}{FB15k237-v1} \\
		&                                   & $\texttt{EI}(L)$    &  $\texttt{ToE}(L)$     & MRR    & $\texttt{EI}(L)$       & $\texttt{ToE}(L)$       & Hit@10   \\
		\midrule
		\multirow{4}{*}{\begin{tabular}[c]{@{}c@{}}not\\ learned\end{tabular}} 
		& Node-wise                         &    $4831$     &   $1.38 \text{E}\!-\!4$      &   $.416$     &   $585$    &  $1.35 \text{E}\!-\!3$      &   $38.9$         \\
		& Layer-wise                        &   $5035$       &    $1.46\text{E}\!-\!4$    &   $.428$     &    $554$        &  $1.45 \text{E}\!-\!3$     &    $37.2$        \\
		& Subgraph                        &  $5098$        &    $1.57\text{E}\!-\!4$   &   $.461$     &    $578$        &   $1.50 \text{E}\!-\!3$    &   $40.5$         \\
		& Incremental                       &   $4954$       &   $1.61 \text{E}\!-\!4$    &   $.472$     &   $559$        &  $1.52 \text{E}\!-\!3$      &    $40.1$        \\
		\midrule
		\multirow{3}{*}{learned}                                               
		& Node-wise                         &    $4913$     &   $1.52\text{E}\!-\!4$     &   $.529$     &    $561$        &    $1.47 \text{E}\!-\!3$    &    50.4        \\
		& Layer-wise                        &   $4871$      & $1.64 \text{E}\!-\!4$  &    $.533$    &       $556$     &    $1.55 \text{E}\!-\!3$    &    52.4        \\
		& Incremental                       &     $4749$     &  $1.78\text{E}\!-\!4$  &   $.562$     &     $564$       &   $1.57 \text{E}\!-\!3$      &    55.1     \\
		\bottomrule
	\end{tabular}
\vspace{-5px}
\end{table}

\subsubsection{Importance of learning}
\noindent \\
To show the necessity of learning,
we apply the same sampling distribution 
and learning strategy
in Section~\ref{sssec:sample:learn}
to node-wise sampling and layer-wise sampling.
The subgraph sampling is hard to be learned with 
the proposed sampling distribution,
thus not compared in this setting.
By comparing the top part and bottom
part in Table~\ref{tab:sampling-methods-comparison},
we observe 
that learning is important for all the three sampling strategies.
Since the entity representations can be used to measure 
the semantic relevance of different entities,
a parameterized sampling distribution
in proportion to {$g(\bm h_{e_o}^\ell\!; {\bm \theta^\ell})$}
helps to identify the promising targets,
increasing $\texttt{ToE}(L)$
and as a result leading to better performance.
%We further provide a comparison of the ST estimator
%and REINFORCE estimator in Appendix~\ref{app:leanring-comp}.
Overall,
the incremental sampling equipped with the learnable sampling distribution
is the best choice among the sampling methods in Table~\ref{tab:sampling-methods-comparison}.

\subsubsection{Learning strategies}
\noindent
\\
\label{sssec:leanring-comp}
In this part,
we compare the variants of learning strategies.
Apart from straight-through estimator (ST)
introduced in Section~\ref{sssec:learning},
another common gradient estimators
for discrete variables is the REINFORCE \cite{williams1992simple,mnih2014neural},
where the gradients are back-propagated by the log-trick (named as REINFORCE).
Alternatively,
the optimization problem in Eq.\eqref{eq:opt_objective_samp}
can also be formed as a bi-level problem (named as Bi-level),
where the sampler parameters are optimized by loss on validation data
and are alternatively updated with the model parameters.
The random sampler without optimizer (named as Random)
is included as a reference baseline.

\begin{figure}[ht]
	\centering
	\vspace{-4px}
	\includegraphics[height=4cm]{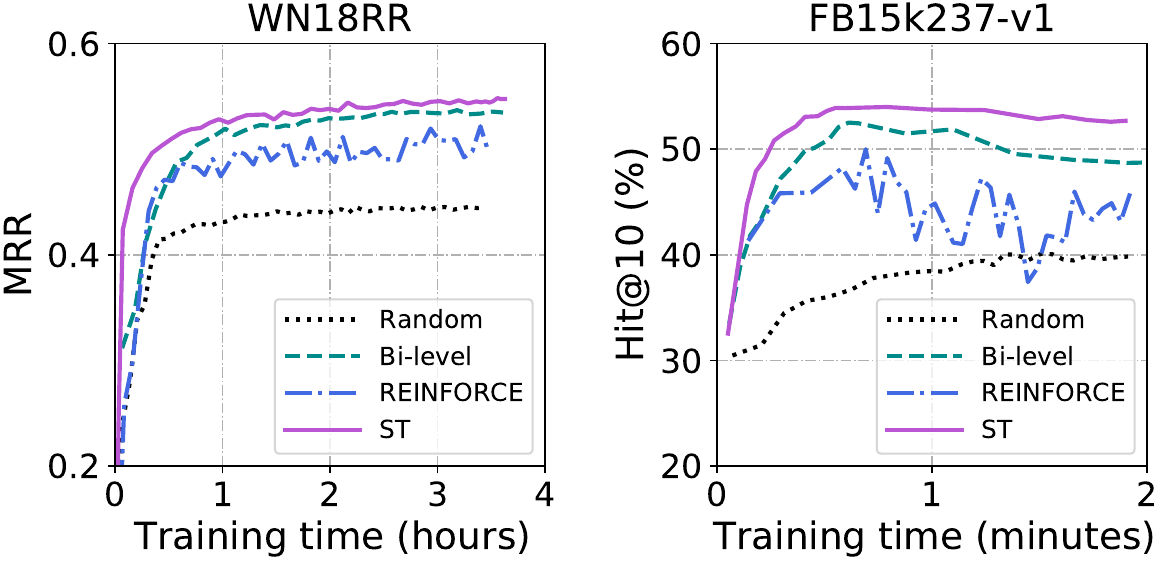}
	\vspace{-10pt}
	\caption{Comparison of learning strategies for the sampler.}
	\label{fig:optimization_sampler}
	\vspace{-5px}
\end{figure}

The evaluation results of the four learning strategies are provided
in Figure~\ref{fig:optimization_sampler}.
First,
compared with the random sampler,
all of the three learned distributions by Bi-level, REINFORCE and ST,
are better,
again verifying the importance of learning
when sampling.
Second,
the learning curves of REINFORCE is quite unstable 
as this estimator has high variance.
Third,
the ST estimator in a single level optimization is better than the other variants
with higher any-time performance.

\subsubsection{Influence of $K$}
\begin{figure}[ht]
	\centering
	\vspace{-4px}
	\includegraphics[height=4cm]{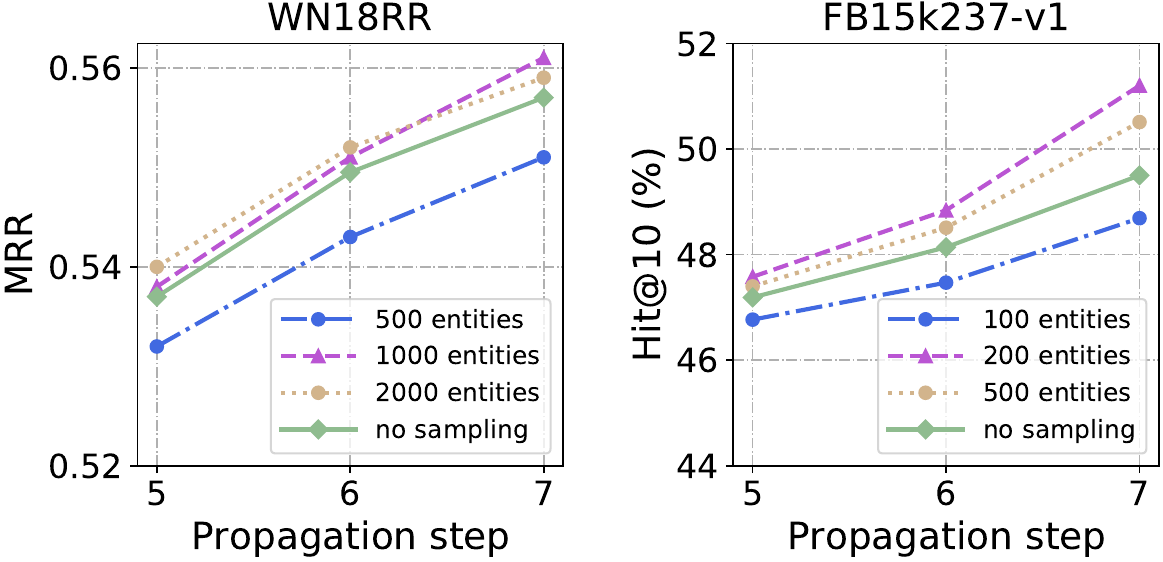}
	\vspace{-10px}
	\caption{Ablation study with $K$ on different $L$. 
		Each line represents the performance of a given $K$ with different $L$.}
	\label{fig:ablation_L_and_K}
	\vspace{-5px}
\end{figure}

\noindent \\
As discussed in Section~\ref{ssec:sample:increase},
the number of sampled entities $K$ bounds the number of involved entities
in the propagation path.
To figure out the influence of $K$,
we show the performance of different $K$ with respect to 
different propagation depth $L$
in Figure~\ref{fig:ablation_L_and_K}.
As shown,
when $K$ is very small,
the results are worse than the no sampling baseline (RED-GNN)
since the ratio of target entity coverage is also small.
When increasing $K$,
the performance tends to be better.
However,
the performance gains 
by sampling more entities
would gradually become marginal or even worse,
especially for larger $L$.
It means that 
the more irrelevant information to the queries will
be included with a larger $K$ and $L$.
Besides,
a larger $K$ will increase the training cost.
Thus,
$K$ should be chosen with a moderate value.

\subsection{Case study: adaptive propagation path}
\label{ssec:casestudy}

%\footnote{+yq+: what about adding a heatmap for selected relations for different query relation?}

\begin{figure}[t]
	\centering
	\vspace{-3px}
	\subfigure[Progressive on {\sf Family}]{\includegraphics[height=3.4cm]{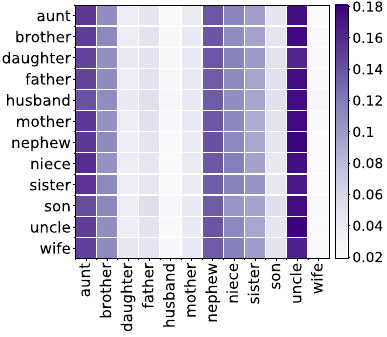}
		\label{fig:heatmap_family_redgnn}}
	\hfill
	\subfigure[AdaProp on {\sf Family}]{\includegraphics[height=3.4cm]{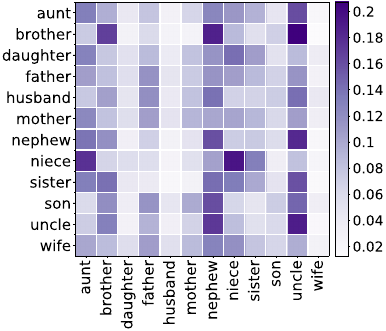}
		\label{fig:heatmap_family_adaprop}}
	\vspace{-5px}
	
	\subfigure[Progressive on {\sf FB15k237}]{\includegraphics[height=3.5cm]{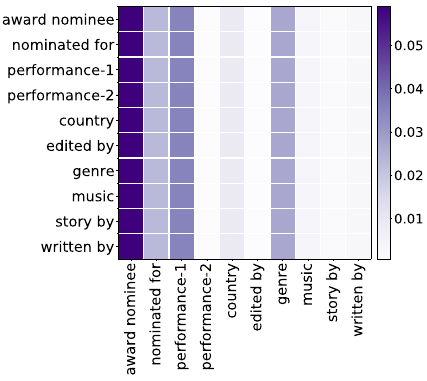}
		\label{fig:heatmap_fb237_redgnn}}
	\hfill
	\subfigure[AdaProp on {\sf FB15k237}]{\includegraphics[height=3.5cm]{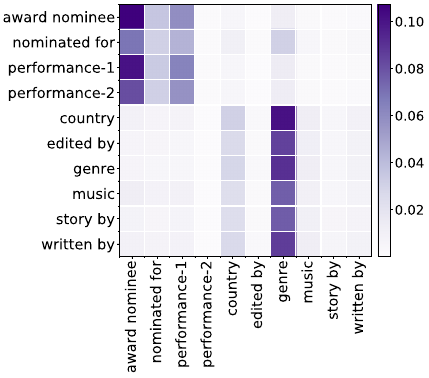}
		\label{fig:heatmap_fb237_adaprop}}
	\vspace{-12px}
	\caption{Heatmaps of relation type ratios in the propagation path.
		Rows are different query relations 
		and columns are relation types in the selected edges.
		For {\sf FB15k237}, we only show 10 selected relations that are related to ``\textit{film}'' here.}
	\label{fig:heatmap}
	\vspace{-14px}
\end{figure}

In Section~\ref{sssec:sample:learn},
we claim that
the sampling distribution is semantic-aware.
To demonstrate this point,
we count the number of relation types
in the sampled edges of the propagation path
for different query relations.
We plot the ratio of relation types
as heatmap
on two semantic friendly datasets,
i.e. {\sf Family} and {\sf FB15k237},
in Figure~\ref{fig:heatmap}.
Specially, as there are 237 relations in {\sf FB15k237},
we plot 10 relations related to ``\textit{film}'' in
Figure~\ref{fig:heatmap_fb237_redgnn}
and Figure~\ref{fig:heatmap_fb237_adaprop}.
We compare the propagation path of progressive propagation methods 
(NBFNet and RED-GNN have the same range)
and AdaProp.

First,
as claimed in the introduction,
the progressive propagation methods
ignore the semantic relevance between local neighborhoods and the query relation.
As a result,
the different rows for different query relations
in Figure~\ref{fig:heatmap_family_redgnn}
and \ref{fig:heatmap_fb237_redgnn}
are the same.
In comparison,
the heatmaps in Figure~\ref{fig:heatmap_family_adaprop} and Figure~\ref{fig:heatmap_fb237_adaprop}
vary for different query relations.
Second,
the more frequent relation types are semantic-aware.
For the heatmap in Figure~\ref{fig:heatmap_family_adaprop},
we observe that when the query relation is \textit{brother},
columns with deeper color are all males,
i.e., \textit{brother}, \textit{nephew} and \textit{uncle}.
When the query relation is \textit{niece},
columns with deeper color are all females,
i.e., \textit{aunt} and \textit{niece}.
For the heatmap in Figure~\ref{fig:heatmap_fb237_adaprop},
the first four relations are about film award 
and the last six relations are about film categories.
As a result,
this figure shows two groups of these relations.
Above evidence
demonstrates that 
AdaProp can indeed learn semantic-aware propagation path.

\begin{figure}[ht]
\vspace{-3px}	
\centering
\subfigure[AdaProp, $q_1$]{\includegraphics[height=3.6cm]{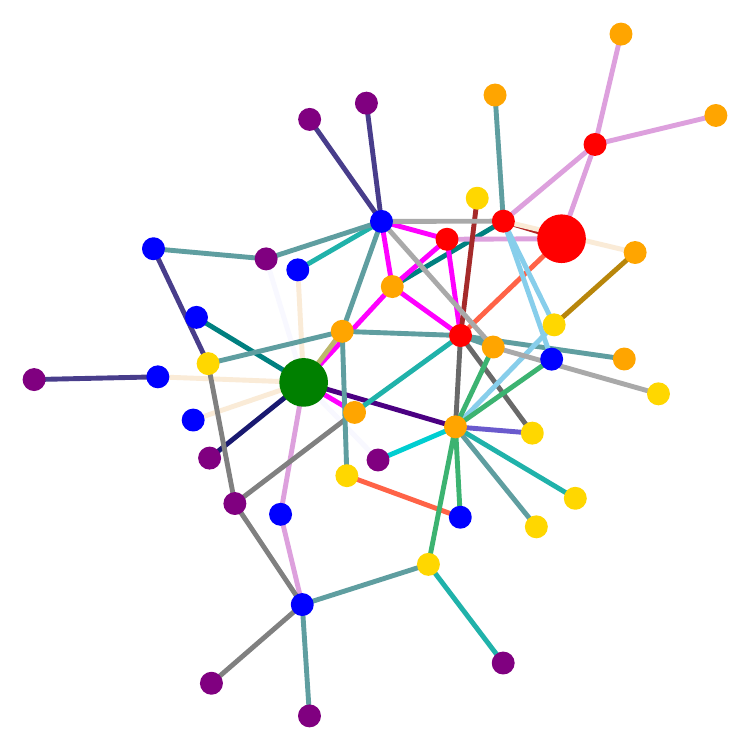}
		\label{fig:ablation_vis_ada_1}}
\quad
\subfigure[Progressive, $q_1$]{\includegraphics[height=3.6cm]{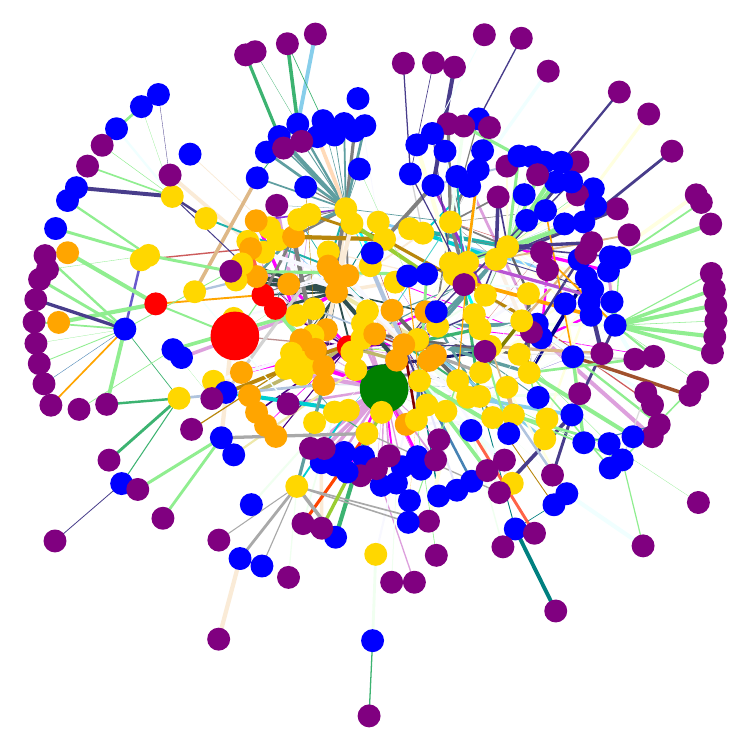}
		\label{fig:ablation_vis_redgnn_1}}

\subfigure[AdaProp, $q_2$]{\includegraphics[height=3.5cm]{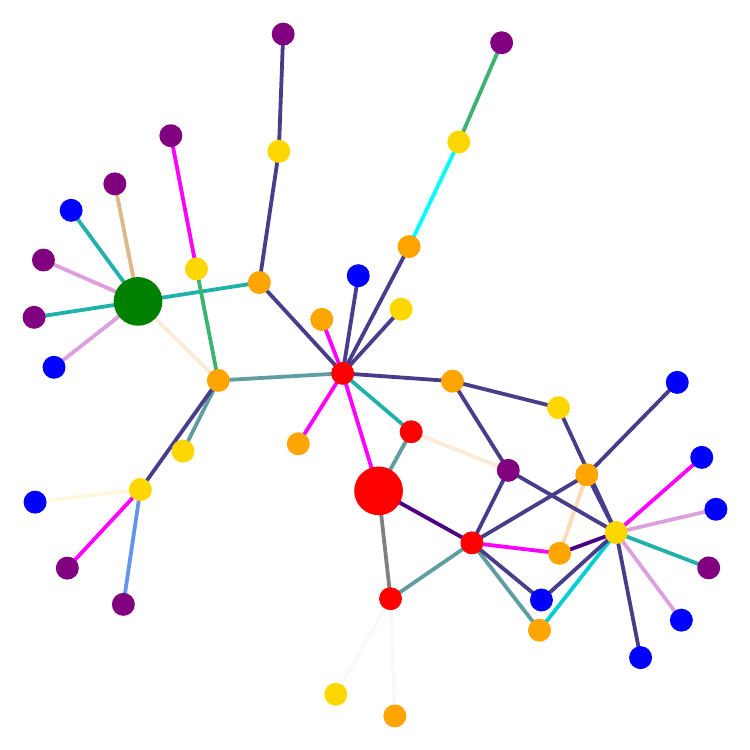}
		\label{fig:ablation_vis_ada_2}}
\quad
\subfigure[Progressive, $q_2$]{\includegraphics[height=3.5cm]{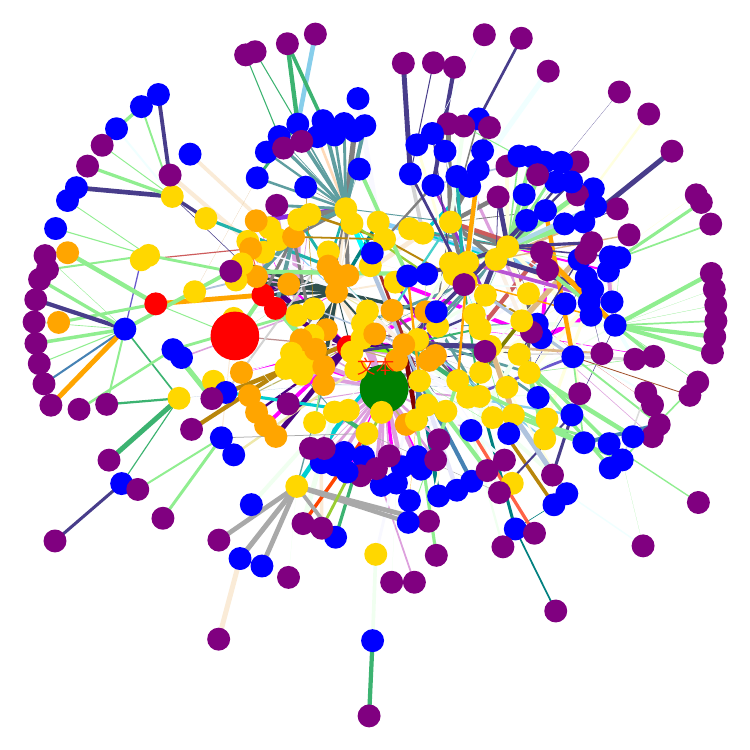}
		\label{fig:ablation_vis_redgnn_2}}
\vspace{-12px}
\caption{Exemplar propagation paths on {\sf FB15k237-v1} dataset. 
		The big red and green nodes indicate the query entity and answer entity, respectively. 
		The small nodes in red, yellow, orange, blue, purple are new entities involved in the $1\sim 5$ propagation steps.}
\label{fig:ablation_vis}
\vspace{-6px}
\end{figure}

In addition,
we visualize the exemplar propagation paths of AdaProp
and progressive propagation methods
for two different queries $q_1=(e_q, r_1,?)$ and $q_2=(e_q,r_2,?)$ 
in Figure~\ref{fig:ablation_vis} (with more examples in in Figure~\ref{fig:more_visualized_examples_2} in Appendix),
which have the same query entity $e_q$ but different query relations 
($r_1\neq r_2$)
on {\sf FB15k237-v1}.
As shown,
there is a clear difference between
the two propagation paths of $q_1$ and $q_2$ learned by AdaProp.
But they are the same for the progressive propagation methods,
where only the edge weights are different.
Besides,
the propagation paths for AdaProp have much fewer entities.
The visualization cases here further demonstrate that
{AdaProp indeed can learn the query dependent propagation paths
	and is entity efficient.}

\section*{Acknowledgment}
Y. Zhang is supported by National Key Research and Development Plan under Grant 2021YFE0205700.
Q. Yao is supported by NSF of China (No. 92270106)
and
Independent scientific research fund (Foshan advanced manufacturing research institute, Tsinghua University).
Z. Zhou and B. Han 
are supported by NSFC Young Scientists Fund No. 62006202, Guangdong Basic and Applied Basic Research Foundation No. 2022A1515011652, and HKBU CSD Departmental Incentive Grant.

\section{Conclusion}
\label{sec:conclude}
In this paper,
we 
propose a new GNN-based KG reasoning method, called AdaProp.
Different from existing GNN-based methods with hand-designed propagation path,
AdaProp learns adaptive propagation path during message propagation.
AdaProp contains two important components,
i.e., an incremental sampling strategy
where nearby targets and layer-wise connections can be preserved with linear complexity
of involved entities,
and a learning based sampling distribution
that can identify semantically related entities during propagation
and sampler is jointly optimized with the GNN models.
Empirical results
on several benchmarking datasets in both transductive and inductive KG reasoning settings
prove the superiority of AdaProp
by learning adaptive propagation path.
The ablation studies show that incremental sampling is better than the other sampling strategies,
and it is important as well as effective to learn the sampling distribution
with straight-through gradient estimator.
Case studies on the learned propagation path
show that the selected entities are semantic-aware and query-dependent.

%By comparing different propagation schemes
%based on the ratio of target entity coverage
%over entities involved,
%and performance with respect to propagation steps,
%we observe that
%a properly designed propagation path
%can lead to better performance when going deeper.
%The key factor is to
%improving the coverage of the true target while reducing increasing number of
%involved entities in deeper steps.
%Hence,
%we propose AdaProp
%to learn an adaptive propagation path
%that filters out irrelevant entities  while
%preserving promising targets during propagation.
%The proposed method
%contains an incremental sampling mechanism 
%and a learning-based sampling distribution 
%to preserve the promising entities while sampling.

\clearpage

%There are also a few limitations for AdaProp.
%First,
%the straight-through estimator is a biased gradient estimator.
%A better design of the learning strategy may further improve the performance.
%Second,
%AdaProp is only evaluated on reasoning missing triplet.
%Reasoning on the multi-hop queries \cite{ren2019query2box}
%is more complex.
%In the future,
%we would like to try the other learning strategies
%and apply AdaProp to more complex reasoning scenarios.

\bibliographystyle{ACM-Reference-Format}
\balance
\bibliography{acmart}


\begin{thebibliography}{57}

%%% the Web address.

\ifx \showCODEN    \undefined \def \showCODEN     #1{\unskip}     \fi
\ifx \showDOI      \undefined \def \showDOI       #1{#1}\fi
\ifx \showISBNx    \undefined \def \showISBNx     #1{\unskip}     \fi
\ifx \showISBNxiii \undefined \def \showISBNxiii  #1{\unskip}     \fi
\ifx \showISSN     \undefined \def \showISSN      #1{\unskip}     \fi
\ifx \showLCCN     \undefined \def \showLCCN      #1{\unskip}     \fi
\ifx \shownote     \undefined \def \shownote      #1{#1}          \fi
\ifx \showarticletitle \undefined \def \showarticletitle #1{#1}   \fi
\ifx \showURL      \undefined \def \showURL       {\relax}        \fi
% invisible to TeX
\providecommand\bibfield[2]{#2}
\providecommand\bibinfo[2]{#2}
\providecommand\natexlab[1]{#1}
\providecommand\showeprint[2][]{arXiv:#2}

\bibitem[\protect\citeauthoryear{Battaglia, Hamrick, Bapst, Sanchez-Gonzalez,
  Zambaldi, et~al\mbox{.}}{Battaglia et~al\mbox{.}}{2018}]%
        {battaglia2018relational}
\bibfield{author}{\bibinfo{person}{P.~W Battaglia}, \bibinfo{person}{J.~B
  Hamrick}, \bibinfo{person}{V. Bapst}, \bibinfo{person}{A. Sanchez-Gonzalez},
  \bibinfo{person}{V. Zambaldi}, {et~al\mbox{.}}}
  \bibinfo{year}{2018}\natexlab{}.
\newblock \bibinfo{booktitle}{\emph{Relational inductive biases, deep learning,
  and graph networks}}.
\newblock \bibinfo{type}{{T}echnical {R}eport}.
  \bibinfo{institution}{arXiv:1806.01261}.
\newblock


\bibitem[\protect\citeauthoryear{Bengio, L{\'e}onard, and Courville}{Bengio
  et~al\mbox{.}}{2013}]%
        {bengio2013estimating}
\bibfield{author}{\bibinfo{person}{Y. Bengio}, \bibinfo{person}{N.
  L{\'e}onard}, {and} \bibinfo{person}{A. Courville}.}
  \bibinfo{year}{2013}\natexlab{}.
\newblock \bibinfo{booktitle}{\emph{Estimating or propagating gradients through
  stochastic neurons for conditional computation}}.
\newblock \bibinfo{type}{{T}echnical {R}eport}.
  \bibinfo{institution}{arXiv:1308.3432}.
\newblock


\bibitem[\protect\citeauthoryear{Bordes, Usunier, Garcia-Duran, Weston, and
  Yakhnenko}{Bordes et~al\mbox{.}}{2013}]%
        {bordes2013translating}
\bibfield{author}{\bibinfo{person}{A. Bordes}, \bibinfo{person}{N. Usunier},
  \bibinfo{person}{A. Garcia-Duran}, \bibinfo{person}{J. Weston}, {and}
  \bibinfo{person}{O. Yakhnenko}.} \bibinfo{year}{2013}\natexlab{}.
\newblock \showarticletitle{Translating embeddings for modeling
  multi-relational data}. In \bibinfo{booktitle}{\emph{NeurIPS}}.
  \bibinfo{pages}{2787--2795}.
\newblock


\bibitem[\protect\citeauthoryear{Cao, Wang, He, Hu, and Chua}{Cao
  et~al\mbox{.}}{2019}]%
        {cao2019unifying}
\bibfield{author}{\bibinfo{person}{Y. Cao}, \bibinfo{person}{X. Wang},
  \bibinfo{person}{X. He}, \bibinfo{person}{Z. Hu}, {and} \bibinfo{person}{T.
  Chua}.} \bibinfo{year}{2019}\natexlab{}.
\newblock \showarticletitle{Unifying knowledge graph learning and
  recommendation: Towards a better understanding of user preferences}. In
  \bibinfo{booktitle}{\emph{TheWebConf}}. \bibinfo{pages}{151--161}.
\newblock


\bibitem[\protect\citeauthoryear{Chen, Ma, and Xiao}{Chen
  et~al\mbox{.}}{2018}]%
        {chen2018fastgcn}
\bibfield{author}{\bibinfo{person}{J. Chen}, \bibinfo{person}{T. Ma}, {and}
  \bibinfo{person}{C. Xiao}.} \bibinfo{year}{2018}\natexlab{}.
\newblock \showarticletitle{FastGCN: Fast Learning with Graph Convolutional
  Networks via Importance Sampling}. In \bibinfo{booktitle}{\emph{ICLR}}.
\newblock


\bibitem[\protect\citeauthoryear{Chen, Jia, and Xiang}{Chen
  et~al\mbox{.}}{2020}]%
        {chen2020review}
\bibfield{author}{\bibinfo{person}{X. Chen}, \bibinfo{person}{S. Jia}, {and}
  \bibinfo{person}{Y. Xiang}.} \bibinfo{year}{2020}\natexlab{}.
\newblock \showarticletitle{A review: Knowledge reasoning over knowledge
  graph}.
\newblock \bibinfo{journal}{\emph{Expert Systems with Applications}}
  \bibinfo{volume}{141} (\bibinfo{year}{2020}), \bibinfo{pages}{112948}.
\newblock


\bibitem[\protect\citeauthoryear{Cheng, Liu, Wang, and Sun}{Cheng
  et~al\mbox{.}}{2022}]%
        {cheng2022rlogic}
\bibfield{author}{\bibinfo{person}{K. Cheng}, \bibinfo{person}{J. Liu},
  \bibinfo{person}{W. Wang}, {and} \bibinfo{person}{Y. Sun}.}
  \bibinfo{year}{2022}\natexlab{}.
\newblock \showarticletitle{RLogic: Recursive Logical Rule Learning from
  Knowledge Graphs}. In \bibinfo{booktitle}{\emph{KDD}}.
  \bibinfo{pages}{179--189}.
\newblock


\bibitem[\protect\citeauthoryear{Chiang, Liu, Si, Li, Bengio, and Hsieh}{Chiang
  et~al\mbox{.}}{2019}]%
        {chiang2019cluster}
\bibfield{author}{\bibinfo{person}{W. Chiang}, \bibinfo{person}{X. Liu},
  \bibinfo{person}{S. Si}, \bibinfo{person}{Y. Li}, \bibinfo{person}{S.
  Bengio}, {and} \bibinfo{person}{C. Hsieh}.} \bibinfo{year}{2019}\natexlab{}.
\newblock \showarticletitle{Cluster-gcn: An efficient algorithm for training
  deep and large graph convolutional networks}. In
  \bibinfo{booktitle}{\emph{SIGKDD}}. \bibinfo{pages}{257--266}.
\newblock


\bibitem[\protect\citeauthoryear{Das, Dhuliawala, Zaheer, Vilnis, Durugkar,
  Krishnamurthy, Smola, and McCallum}{Das et~al\mbox{.}}{2017}]%
        {das2017go}
\bibfield{author}{\bibinfo{person}{R. Das}, \bibinfo{person}{S. Dhuliawala},
  \bibinfo{person}{M. Zaheer}, \bibinfo{person}{L. Vilnis}, \bibinfo{person}{I.
  Durugkar}, \bibinfo{person}{A. Krishnamurthy}, \bibinfo{person}{A. Smola},
  {and} \bibinfo{person}{A. McCallum}.} \bibinfo{year}{2017}\natexlab{}.
\newblock \showarticletitle{Go for a walk and arrive at the answer: Reasoning
  over paths in knowledge bases using reinforcement learning}. In
  \bibinfo{booktitle}{\emph{ICLR}}.
\newblock


\bibitem[\protect\citeauthoryear{Dettmers, Minervini, Stenetorp, and
  Riedel}{Dettmers et~al\mbox{.}}{2017}]%
        {dettmers2017convolutional}
\bibfield{author}{\bibinfo{person}{T. Dettmers}, \bibinfo{person}{P.
  Minervini}, \bibinfo{person}{P. Stenetorp}, {and} \bibinfo{person}{S.
  Riedel}.} \bibinfo{year}{2017}\natexlab{}.
\newblock \showarticletitle{Convolutional {2D} knowledge graph embeddings}. In
  \bibinfo{booktitle}{\emph{AAAI}}.
\newblock


\bibitem[\protect\citeauthoryear{Dettmers, Minervini, Stenetorp, and
  Riedel}{Dettmers et~al\mbox{.}}{2018}]%
        {dettmers2018convolutional}
\bibfield{author}{\bibinfo{person}{T. Dettmers}, \bibinfo{person}{P.
  Minervini}, \bibinfo{person}{P. Stenetorp}, {and} \bibinfo{person}{S.
  Riedel}.} \bibinfo{year}{2018}\natexlab{}.
\newblock \showarticletitle{Convolutional 2d knowledge graph embeddings}. In
  \bibinfo{booktitle}{\emph{AAAI}}.
\newblock


\bibitem[\protect\citeauthoryear{Dhingra, Zaheer, Balachandran, Neubig,
  Salakhutdinov, and Cohen}{Dhingra et~al\mbox{.}}{2019}]%
        {dhingra2019differentiable}
\bibfield{author}{\bibinfo{person}{B. Dhingra}, \bibinfo{person}{M. Zaheer},
  \bibinfo{person}{V. Balachandran}, \bibinfo{person}{G. Neubig},
  \bibinfo{person}{R. Salakhutdinov}, {and} \bibinfo{person}{W.~W Cohen}.}
  \bibinfo{year}{2019}\natexlab{}.
\newblock \showarticletitle{Differentiable Reasoning over a Virtual Knowledge
  Base}. In \bibinfo{booktitle}{\emph{ICLR}}.
\newblock


\bibitem[\protect\citeauthoryear{Feeney, Gupta, Thost, Angell, Chandu,
  Adhikari, and Ma}{Feeney et~al\mbox{.}}{2021}]%
        {feeney2021relation}
\bibfield{author}{\bibinfo{person}{A. Feeney}, \bibinfo{person}{R. Gupta},
  \bibinfo{person}{V. Thost}, \bibinfo{person}{R. Angell}, \bibinfo{person}{G.
  Chandu}, \bibinfo{person}{Y. Adhikari}, {and} \bibinfo{person}{T. Ma}.}
  \bibinfo{year}{2021}\natexlab{}.
\newblock \bibinfo{booktitle}{\emph{Relation matters in sampling: a scalable
  multi-relational graph neural network for drug-drug interaction prediction}}.
\newblock \bibinfo{type}{{T}echnical {R}eport}.
  \bibinfo{institution}{arXiv:2105.13975}.
\newblock


\bibitem[\protect\citeauthoryear{Gilmer, Schoenholz, Riley, Vinyals, and
  Dahl}{Gilmer et~al\mbox{.}}{2017}]%
        {gilmer2017neural}
\bibfield{author}{\bibinfo{person}{J. Gilmer}, \bibinfo{person}{S.~S
  Schoenholz}, \bibinfo{person}{P.~F Riley}, \bibinfo{person}{O. Vinyals},
  {and} \bibinfo{person}{G.~E Dahl}.} \bibinfo{year}{2017}\natexlab{}.
\newblock \showarticletitle{Neural Message Passing for Quantum Chemistry}. In
  \bibinfo{booktitle}{\emph{ICML}}. \bibinfo{pages}{1263--1272}.
\newblock


\bibitem[\protect\citeauthoryear{Hamilton, Ying, and Leskovec}{Hamilton
  et~al\mbox{.}}{2017}]%
        {hamilton2017inductive}
\bibfield{author}{\bibinfo{person}{W. Hamilton}, \bibinfo{person}{Z. Ying},
  {and} \bibinfo{person}{J. Leskovec}.} \bibinfo{year}{2017}\natexlab{}.
\newblock \showarticletitle{Inductive representation learning on large graphs}.
  In \bibinfo{booktitle}{\emph{NeurIPS}}. \bibinfo{pages}{1024--1034}.
\newblock


\bibitem[\protect\citeauthoryear{Hogan, Blomqvist, Cochez, d’Amato, Melo,
  Gutierrez, Kirrane, Gayo, Navigli, Neumaier, et~al\mbox{.}}{Hogan
  et~al\mbox{.}}{2021}]%
        {hogan2021knowledge}
\bibfield{author}{\bibinfo{person}{A. Hogan}, \bibinfo{person}{E. Blomqvist},
  \bibinfo{person}{M. Cochez}, \bibinfo{person}{C. d’Amato},
  \bibinfo{person}{G.~de Melo}, \bibinfo{person}{C. Gutierrez},
  \bibinfo{person}{S. Kirrane}, \bibinfo{person}{J.~Emilio~L. Gayo},
  \bibinfo{person}{R. Navigli}, \bibinfo{person}{S. Neumaier}, {et~al\mbox{.}}}
  \bibinfo{year}{2021}\natexlab{}.
\newblock \showarticletitle{Knowledge graphs}.
\newblock \bibinfo{journal}{\emph{ACM Computing Surveys (CSUR)}}
  \bibinfo{volume}{54}, \bibinfo{number}{4} (\bibinfo{year}{2021}),
  \bibinfo{pages}{1--37}.
\newblock


\bibitem[\protect\citeauthoryear{Huang, Zhang, Rong, and Huang}{Huang
  et~al\mbox{.}}{2018}]%
        {huang2018adaptive}
\bibfield{author}{\bibinfo{person}{W. Huang}, \bibinfo{person}{T. Zhang},
  \bibinfo{person}{Y. Rong}, {and} \bibinfo{person}{J. Huang}.}
  \bibinfo{year}{2018}\natexlab{}.
\newblock \showarticletitle{Adaptive Sampling Towards Fast Graph Representation
  Learning}. In \bibinfo{booktitle}{\emph{NeurIPS}}, Vol.~\bibinfo{volume}{31}.
  \bibinfo{pages}{4558--4567}.
\newblock


\bibitem[\protect\citeauthoryear{Huang, Zhang, Li, and Li}{Huang
  et~al\mbox{.}}{2019}]%
        {huang2019knowledge}
\bibfield{author}{\bibinfo{person}{X. Huang}, \bibinfo{person}{J. Zhang},
  \bibinfo{person}{D. Li}, {and} \bibinfo{person}{P. Li}.}
  \bibinfo{year}{2019}\natexlab{}.
\newblock \showarticletitle{Knowledge graph embedding based question
  answering}. In \bibinfo{booktitle}{\emph{WSDM}}. \bibinfo{pages}{105--113}.
\newblock


\bibitem[\protect\citeauthoryear{Jang, Gu, and Poole}{Jang
  et~al\mbox{.}}{2017}]%
        {jang2017categorical}
\bibfield{author}{\bibinfo{person}{E. Jang}, \bibinfo{person}{S. Gu}, {and}
  \bibinfo{person}{B. Poole}.} \bibinfo{year}{2017}\natexlab{}.
\newblock \showarticletitle{Categorical Reparameterization with
  Gumbel-Softmax}. In \bibinfo{booktitle}{\emph{ICLR}}.
\newblock


\bibitem[\protect\citeauthoryear{Ji, Pan, Cambria, Marttinen, and Yu}{Ji
  et~al\mbox{.}}{2020}]%
        {ji2020survey}
\bibfield{author}{\bibinfo{person}{S. Ji}, \bibinfo{person}{S. Pan},
  \bibinfo{person}{E. Cambria}, \bibinfo{person}{P. Marttinen}, {and}
  \bibinfo{person}{P.~S Yu}.} \bibinfo{year}{2020}\natexlab{}.
\newblock \bibinfo{booktitle}{\emph{A survey on knowledge graphs:
  Representation, acquisition and applications}}.
\newblock \bibinfo{type}{{T}echnical {R}eport}.
  \bibinfo{institution}{arXiv:2002.00388}.
\newblock


\bibitem[\protect\citeauthoryear{Kingma and Ba}{Kingma and Ba}{2014}]%
        {kingma2014adam}
\bibfield{author}{\bibinfo{person}{D.~P Kingma} {and} \bibinfo{person}{J. Ba}.}
  \bibinfo{year}{2014}\natexlab{}.
\newblock \bibinfo{booktitle}{\emph{Adam: A method for stochastic
  optimization}}.
\newblock \bibinfo{type}{{T}echnical {R}eport}.
  \bibinfo{institution}{arXiv:1412.6980}.
\newblock


\bibitem[\protect\citeauthoryear{Kipf and Welling}{Kipf and Welling}{2016}]%
        {kipf2016semi}
\bibfield{author}{\bibinfo{person}{T.~N Kipf} {and} \bibinfo{person}{M.
  Welling}.} \bibinfo{year}{2016}\natexlab{}.
\newblock \showarticletitle{Semi-supervised classification with graph
  convolutional networks}. In \bibinfo{booktitle}{\emph{ICLR}}.
\newblock


\bibitem[\protect\citeauthoryear{Kok and Domingos}{Kok and Domingos}{2007}]%
        {kok2007statistical}
\bibfield{author}{\bibinfo{person}{S. Kok} {and} \bibinfo{person}{P.
  Domingos}.} \bibinfo{year}{2007}\natexlab{}.
\newblock \showarticletitle{Statistical predicate invention}. In
  \bibinfo{booktitle}{\emph{ICML}}. \bibinfo{pages}{433--440}.
\newblock


\bibitem[\protect\citeauthoryear{Kool, Van~Hoof, and Welling}{Kool
  et~al\mbox{.}}{2019}]%
        {kool2019stochastic}
\bibfield{author}{\bibinfo{person}{W. Kool}, \bibinfo{person}{H. Van~Hoof},
  {and} \bibinfo{person}{M. Welling}.} \bibinfo{year}{2019}\natexlab{}.
\newblock \showarticletitle{Stochastic beams and where to find them: The
  gumbel-top-k trick for sampling sequences without replacement}. In
  \bibinfo{booktitle}{\emph{ICML}}. PMLR, \bibinfo{pages}{3499--3508}.
\newblock


\bibitem[\protect\citeauthoryear{Mai, Zheng, Yang, and Hu}{Mai
  et~al\mbox{.}}{2021}]%
        {mai2021communicative}
\bibfield{author}{\bibinfo{person}{S. Mai}, \bibinfo{person}{S. Zheng},
  \bibinfo{person}{Y. Yang}, {and} \bibinfo{person}{H. Hu}.}
  \bibinfo{year}{2021}\natexlab{}.
\newblock \showarticletitle{Communicative Message Passing for Inductive
  Relation Reasoning.}. In \bibinfo{booktitle}{\emph{AAAI}}.
  \bibinfo{pages}{4294--4302}.
\newblock


\bibitem[\protect\citeauthoryear{Meilicke, Fink, Wang, Ruffinelli, Gemulla, and
  Stuckenschmidt}{Meilicke et~al\mbox{.}}{2018}]%
        {meilicke2018fine}
\bibfield{author}{\bibinfo{person}{C. Meilicke}, \bibinfo{person}{M. Fink},
  \bibinfo{person}{Y. Wang}, \bibinfo{person}{D. Ruffinelli},
  \bibinfo{person}{R. Gemulla}, {and} \bibinfo{person}{H. Stuckenschmidt}.}
  \bibinfo{year}{2018}\natexlab{}.
\newblock \showarticletitle{Fine-grained evaluation of rule-and embedding-based
  systems for knowledge graph completion}. In \bibinfo{booktitle}{\emph{ISWC}}.
  Springer, \bibinfo{pages}{3--20}.
\newblock


\bibitem[\protect\citeauthoryear{Mnih and Gregor}{Mnih and Gregor}{2014}]%
        {mnih2014neural}
\bibfield{author}{\bibinfo{person}{A. Mnih} {and} \bibinfo{person}{K. Gregor}.}
  \bibinfo{year}{2014}\natexlab{}.
\newblock \showarticletitle{Neural variational inference and learning in belief
  networks}. In \bibinfo{booktitle}{\emph{ICML}}. PMLR,
  \bibinfo{pages}{1791--1799}.
\newblock


\bibitem[\protect\citeauthoryear{Oono and Suzuki}{Oono and Suzuki}{2019}]%
        {oono2019graph}
\bibfield{author}{\bibinfo{person}{K. Oono} {and} \bibinfo{person}{T. Suzuki}.}
  \bibinfo{year}{2019}\natexlab{}.
\newblock \showarticletitle{Graph Neural Networks Exponentially Lose Expressive
  Power for Node Classification}. In \bibinfo{booktitle}{\emph{ICLR}}.
\newblock


\bibitem[\protect\citeauthoryear{Paszke, Gross, Chintala, Chanan, Yang, DeVito,
  Lin, Desmaison, Antiga, and Lerer}{Paszke et~al\mbox{.}}{2017}]%
        {paszke2017automatic}
\bibfield{author}{\bibinfo{person}{A. Paszke}, \bibinfo{person}{S. Gross},
  \bibinfo{person}{S. Chintala}, \bibinfo{person}{G. Chanan},
  \bibinfo{person}{E. Yang}, \bibinfo{person}{Z. DeVito}, \bibinfo{person}{Z.
  Lin}, \bibinfo{person}{A. Desmaison}, \bibinfo{person}{L. Antiga}, {and}
  \bibinfo{person}{A. Lerer}.} \bibinfo{year}{2017}\natexlab{}.
\newblock \showarticletitle{Automatic differentiation in {PyTorch}}. In
  \bibinfo{booktitle}{\emph{ICLR}}.
\newblock


\bibitem[\protect\citeauthoryear{Qu, Chen, Xhonneux, Bengio, and Tang}{Qu
  et~al\mbox{.}}{2021}]%
        {qu2021rnnlogic}
\bibfield{author}{\bibinfo{person}{M. Qu}, \bibinfo{person}{J. Chen},
  \bibinfo{person}{L. Xhonneux}, \bibinfo{person}{Y. Bengio}, {and}
  \bibinfo{person}{J. Tang}.} \bibinfo{year}{2021}\natexlab{}.
\newblock \showarticletitle{RNNLogic: Learning Logic Rules for Reasoning on
  Knowledge Graphs}. In \bibinfo{booktitle}{\emph{ICLR}}.
\newblock


\bibitem[\protect\citeauthoryear{Sadeghian, Armandpour, Ding, and
  Wang}{Sadeghian et~al\mbox{.}}{2019}]%
        {sadeghian2019drum}
\bibfield{author}{\bibinfo{person}{A. Sadeghian}, \bibinfo{person}{M.
  Armandpour}, \bibinfo{person}{P. Ding}, {and} \bibinfo{person}{D.~Z. Wang}.}
  \bibinfo{year}{2019}\natexlab{}.
\newblock \showarticletitle{DRUM: End-To-End Differentiable Rule Mining On
  Knowledge Graphs}. In \bibinfo{booktitle}{\emph{NeurIPS}}.
  \bibinfo{pages}{15347--15357}.
\newblock


\bibitem[\protect\citeauthoryear{Schlichtkrull, Kipf, Bloem, Van Den~Berg,
  Titov, and Welling}{Schlichtkrull et~al\mbox{.}}{2018}]%
        {schlichtkrull2018modeling}
\bibfield{author}{\bibinfo{person}{M. Schlichtkrull}, \bibinfo{person}{T.~N
  Kipf}, \bibinfo{person}{P. Bloem}, \bibinfo{person}{R. Van Den~Berg},
  \bibinfo{person}{I. Titov}, {and} \bibinfo{person}{M. Welling}.}
  \bibinfo{year}{2018}\natexlab{}.
\newblock \showarticletitle{Modeling relational data with graph convolutional
  networks}. In \bibinfo{booktitle}{\emph{ESWC}}. Springer,
  \bibinfo{pages}{593--607}.
\newblock


\bibitem[\protect\citeauthoryear{Suchanek, Kasneci, and Weikum}{Suchanek
  et~al\mbox{.}}{2007}]%
        {suchanek2007yago}
\bibfield{author}{\bibinfo{person}{F. Suchanek}, \bibinfo{person}{G. Kasneci},
  {and} \bibinfo{person}{G. Weikum}.} \bibinfo{year}{2007}\natexlab{}.
\newblock \showarticletitle{Yago: A core of semantic knowledge}. In
  \bibinfo{booktitle}{\emph{The WebConf}}. \bibinfo{pages}{697--706}.
\newblock


\bibitem[\protect\citeauthoryear{Sun, Deng, Nie, and Tang}{Sun
  et~al\mbox{.}}{2019}]%
        {sun2019rotate}
\bibfield{author}{\bibinfo{person}{Z. Sun}, \bibinfo{person}{Z. Deng},
  \bibinfo{person}{J. Nie}, {and} \bibinfo{person}{J. Tang}.}
  \bibinfo{year}{2019}\natexlab{}.
\newblock \showarticletitle{Rotate: Knowledge graph embedding by relational
  rotation in complex space}. In \bibinfo{booktitle}{\emph{ICLR}}.
\newblock


\bibitem[\protect\citeauthoryear{Sutton and Barto}{Sutton and Barto}{2018}]%
        {sutton2018reinforcement}
\bibfield{author}{\bibinfo{person}{R.~S Sutton} {and} \bibinfo{person}{A.
  Barto}.} \bibinfo{year}{2018}\natexlab{}.
\newblock \bibinfo{booktitle}{\emph{Reinforcement learning: An introduction}}.
\newblock \bibinfo{publisher}{MIT press}.
\newblock


\bibitem[\protect\citeauthoryear{Teru, Denis, and Hamilton}{Teru
  et~al\mbox{.}}{2020}]%
        {teru2019inductive}
\bibfield{author}{\bibinfo{person}{K.~K Teru}, \bibinfo{person}{E. Denis},
  {and} \bibinfo{person}{W.~L Hamilton}.} \bibinfo{year}{2020}\natexlab{}.
\newblock \bibinfo{booktitle}{\emph{Inductive Relation Prediction by Subgraph
  Reasoning}}.
\newblock \bibinfo{type}{{T}echnical {R}eport}.
  \bibinfo{institution}{arXiv:1911.06962}.
\newblock


\bibitem[\protect\citeauthoryear{Toutanova and Chen}{Toutanova and
  Chen}{2015}]%
        {toutanova2015observed}
\bibfield{author}{\bibinfo{person}{K. Toutanova} {and} \bibinfo{person}{D.
  Chen}.} \bibinfo{year}{2015}\natexlab{}.
\newblock \showarticletitle{Observed versus latent features for knowledge base
  and text inference}. In \bibinfo{booktitle}{\emph{PWCVSMC}}.
  \bibinfo{pages}{57--66}.
\newblock


\bibitem[\protect\citeauthoryear{Vashishth, Sanyal, Nitin, and
  Talukdar}{Vashishth et~al\mbox{.}}{2019}]%
        {vashishth2019composition}
\bibfield{author}{\bibinfo{person}{S. Vashishth}, \bibinfo{person}{S. Sanyal},
  \bibinfo{person}{V. Nitin}, {and} \bibinfo{person}{P. Talukdar}.}
  \bibinfo{year}{2019}\natexlab{}.
\newblock \showarticletitle{Composition-based multi-relational graph
  convolutional networks}.
\newblock


\bibitem[\protect\citeauthoryear{Wang, Mao, Wang, and Guo}{Wang
  et~al\mbox{.}}{2017}]%
        {wang2017knowledge}
\bibfield{author}{\bibinfo{person}{Q. Wang}, \bibinfo{person}{Z. Mao},
  \bibinfo{person}{B. Wang}, {and} \bibinfo{person}{L. Guo}.}
  \bibinfo{year}{2017}\natexlab{}.
\newblock \showarticletitle{Knowledge graph embedding: A survey of approaches
  and applications}.
\newblock \bibinfo{journal}{\emph{TKDE}} \bibinfo{volume}{29},
  \bibinfo{number}{12} (\bibinfo{year}{2017}), \bibinfo{pages}{2724--2743}.
\newblock


\bibitem[\protect\citeauthoryear{Wang, He, Cao, Liu, and Chua}{Wang
  et~al\mbox{.}}{2019}]%
        {wang2019kgat}
\bibfield{author}{\bibinfo{person}{X. Wang}, \bibinfo{person}{X. He},
  \bibinfo{person}{Y. Cao}, \bibinfo{person}{M. Liu}, {and} \bibinfo{person}{T.
  Chua}.} \bibinfo{year}{2019}\natexlab{}.
\newblock \showarticletitle{Kgat: Knowledge graph attention network for
  recommendation}. In \bibinfo{booktitle}{\emph{SIGKDD}}.
  \bibinfo{pages}{950--958}.
\newblock


\bibitem[\protect\citeauthoryear{Williams}{Williams}{1992}]%
        {williams1992simple}
\bibfield{author}{\bibinfo{person}{R.~J. Williams}.}
  \bibinfo{year}{1992}\natexlab{}.
\newblock \showarticletitle{Simple statistical gradient-following algorithms
  for connectionist reinforcement learning}.
\newblock \bibinfo{journal}{\emph{Machine Learning Journal}}
  \bibinfo{volume}{8}, \bibinfo{number}{3-4} (\bibinfo{year}{1992}),
  \bibinfo{pages}{229--256}.
\newblock


\bibitem[\protect\citeauthoryear{Wilt, Thayer, and Ruml}{Wilt
  et~al\mbox{.}}{2010}]%
        {wilt2010comparison}
\bibfield{author}{\bibinfo{person}{C. Wilt}, \bibinfo{person}{J. Thayer}, {and}
  \bibinfo{person}{W. Ruml}.} \bibinfo{year}{2010}\natexlab{}.
\newblock \showarticletitle{A comparison of greedy search algorithms}. In
  \bibinfo{booktitle}{\emph{Proceedings of the International Symposium on
  Combinatorial Search}}, Vol.~\bibinfo{volume}{1}. \bibinfo{pages}{129--136}.
\newblock


\bibitem[\protect\citeauthoryear{Xie and Ermon}{Xie and Ermon}{2019}]%
        {xie2019reparameterizable}
\bibfield{author}{\bibinfo{person}{S. Xie} {and} \bibinfo{person}{S. Ermon}.}
  \bibinfo{year}{2019}\natexlab{}.
\newblock \showarticletitle{Reparameterizable Subset Sampling via Continuous
  Relaxations}. In \bibinfo{booktitle}{\emph{IJCAI}}.
\newblock


\bibitem[\protect\citeauthoryear{Xiong, Hoang, and Wang}{Xiong
  et~al\mbox{.}}{2017}]%
        {xiong2017deeppath}
\bibfield{author}{\bibinfo{person}{W. Xiong}, \bibinfo{person}{T. Hoang}, {and}
  \bibinfo{person}{W. Wang}.} \bibinfo{year}{2017}\natexlab{}.
\newblock \showarticletitle{DeepPath: A Reinforcement Learning Method for
  Knowledge Graph Reasoning}. In \bibinfo{booktitle}{\emph{EMNLP}}.
  \bibinfo{pages}{564--573}.
\newblock


\bibitem[\protect\citeauthoryear{Xu, Feng, Jiang, Xie, Sun, and Deng}{Xu
  et~al\mbox{.}}{2019}]%
        {xu2019dynamically}
\bibfield{author}{\bibinfo{person}{X. Xu}, \bibinfo{person}{W. Feng},
  \bibinfo{person}{Y. Jiang}, \bibinfo{person}{X. Xie}, \bibinfo{person}{Z.
  Sun}, {and} \bibinfo{person}{Z. Deng}.} \bibinfo{year}{2019}\natexlab{}.
\newblock \showarticletitle{Dynamically Pruned Message Passing Networks for
  Large-Scale Knowledge Graph Reasoning}. In \bibinfo{booktitle}{\emph{ICLR}}.
\newblock


\bibitem[\protect\citeauthoryear{Yang, Yang, and Cohen}{Yang
  et~al\mbox{.}}{2017}]%
        {yang2017differentiable}
\bibfield{author}{\bibinfo{person}{F. Yang}, \bibinfo{person}{Z. Yang}, {and}
  \bibinfo{person}{W.~W Cohen}.} \bibinfo{year}{2017}\natexlab{}.
\newblock \showarticletitle{Differentiable learning of logical rules for
  knowledge base reasoning}. In \bibinfo{booktitle}{\emph{NeurIPS}}.
  \bibinfo{pages}{2319--2328}.
\newblock


\bibitem[\protect\citeauthoryear{Ye, Kumar, Sing, Song, and Wang}{Ye
  et~al\mbox{.}}{2022}]%
        {ye2022comprehensive}
\bibfield{author}{\bibinfo{person}{Z. Ye}, \bibinfo{person}{Y.~Jaya Kumar},
  \bibinfo{person}{G. Sing}, \bibinfo{person}{F. Song}, {and}
  \bibinfo{person}{J. Wang}.} \bibinfo{year}{2022}\natexlab{}.
\newblock \showarticletitle{A comprehensive survey of graph neural networks for
  knowledge graphs}.
\newblock \bibinfo{journal}{\emph{IEEE Access}}  \bibinfo{volume}{10}
  (\bibinfo{year}{2022}), \bibinfo{pages}{75729--75741}.
\newblock


\bibitem[\protect\citeauthoryear{Yoon, Gervet, Shi, Niu, He, and Yang}{Yoon
  et~al\mbox{.}}{2021}]%
        {yoon2021performance}
\bibfield{author}{\bibinfo{person}{M. Yoon}, \bibinfo{person}{T. Gervet},
  \bibinfo{person}{B. Shi}, \bibinfo{person}{S. Niu}, \bibinfo{person}{Q. He},
  {and} \bibinfo{person}{J. Yang}.} \bibinfo{year}{2021}\natexlab{}.
\newblock \showarticletitle{Performance-Adaptive Sampling Strategy Towards Fast
  and Accurate Graph Neural Networks}. In \bibinfo{booktitle}{\emph{SIGKDD}}.
  \bibinfo{pages}{2046--2056}.
\newblock


\bibitem[\protect\citeauthoryear{Yu, Huang, Zhang, Glass, Sun, and Xiao}{Yu
  et~al\mbox{.}}{2021}]%
        {yu2021sumgnn}
\bibfield{author}{\bibinfo{person}{Y. Yu}, \bibinfo{person}{K. Huang},
  \bibinfo{person}{C. Zhang}, \bibinfo{person}{L.~M Glass}, \bibinfo{person}{J.
  Sun}, {and} \bibinfo{person}{C. Xiao}.} \bibinfo{year}{2021}\natexlab{}.
\newblock \showarticletitle{SumGNN: multi-typed drug interaction prediction via
  efficient knowledge graph summarization}.
\newblock \bibinfo{journal}{\emph{Bioinformatics}} \bibinfo{volume}{37},
  \bibinfo{number}{18} (\bibinfo{year}{2021}), \bibinfo{pages}{2988--2995}.
\newblock


\bibitem[\protect\citeauthoryear{Zeng, Zhang, Xia, Srivastava, Malevich,
  Kannan, Prasanna, Jin, and Chen}{Zeng et~al\mbox{.}}{2021}]%
        {zeng2021decoupling}
\bibfield{author}{\bibinfo{person}{H. Zeng}, \bibinfo{person}{M. Zhang},
  \bibinfo{person}{Y. Xia}, \bibinfo{person}{A. Srivastava},
  \bibinfo{person}{A. Malevich}, \bibinfo{person}{R. Kannan},
  \bibinfo{person}{V. Prasanna}, \bibinfo{person}{L. Jin}, {and}
  \bibinfo{person}{R. Chen}.} \bibinfo{year}{2021}\natexlab{}.
\newblock \showarticletitle{Decoupling the Depth and Scope of Graph Neural
  Networks}. In \bibinfo{booktitle}{\emph{NeurIPS}}, Vol.~\bibinfo{volume}{34}.
\newblock


\bibitem[\protect\citeauthoryear{Zeng, Zhou, Srivastava, Kannan, and
  Prasanna}{Zeng et~al\mbox{.}}{2019}]%
        {zeng2019graphsaint}
\bibfield{author}{\bibinfo{person}{H. Zeng}, \bibinfo{person}{H. Zhou},
  \bibinfo{person}{A. Srivastava}, \bibinfo{person}{R. Kannan}, {and}
  \bibinfo{person}{V. Prasanna}.} \bibinfo{year}{2019}\natexlab{}.
\newblock \showarticletitle{GraphSAINT: Graph Sampling Based Inductive Learning
  Method}. In \bibinfo{booktitle}{\emph{ICLR}}.
\newblock


\bibitem[\protect\citeauthoryear{Zhang, Tay, Yao, and Liu}{Zhang
  et~al\mbox{.}}{2019}]%
        {zhang2019quaternion}
\bibfield{author}{\bibinfo{person}{S. Zhang}, \bibinfo{person}{Y. Tay},
  \bibinfo{person}{L. Yao}, {and} \bibinfo{person}{Q. Liu}.}
  \bibinfo{year}{2019}\natexlab{}.
\newblock \showarticletitle{Quaternion knowledge graph embeddings}.
\newblock \bibinfo{journal}{\emph{NeurIPS}}  \bibinfo{volume}{32}
  (\bibinfo{year}{2019}).
\newblock


\bibitem[\protect\citeauthoryear{Zhang and Yao}{Zhang and Yao}{2022}]%
        {zhang2021knowledge}
\bibfield{author}{\bibinfo{person}{Y. Zhang} {and} \bibinfo{person}{Q. Yao}.}
  \bibinfo{year}{2022}\natexlab{}.
\newblock \showarticletitle{Knowledge Graph Reasoning with Relational Directed
  Graph}. In \bibinfo{booktitle}{\emph{TheWebConf}}.
\newblock


\bibitem[\protect\citeauthoryear{Zhang, Yao, and Chen}{Zhang
  et~al\mbox{.}}{2020a}]%
        {zhang2020interstellar}
\bibfield{author}{\bibinfo{person}{Y. Zhang}, \bibinfo{person}{Q. Yao}, {and}
  \bibinfo{person}{L. Chen}.} \bibinfo{year}{2020}\natexlab{a}.
\newblock \showarticletitle{Interstellar: Searching Recurrent Architecture for
  Knowledge Graph Embedding}. In \bibinfo{booktitle}{\emph{NeurIPS}},
  Vol.~\bibinfo{volume}{33}.
\newblock


\bibitem[\protect\citeauthoryear{Zhang, Yao, Dai, and Chen}{Zhang
  et~al\mbox{.}}{2020b}]%
        {zhang2020autosf}
\bibfield{author}{\bibinfo{person}{Y. Zhang}, \bibinfo{person}{Q. Yao},
  \bibinfo{person}{W. Dai}, {and} \bibinfo{person}{L. Chen}.}
  \bibinfo{year}{2020}\natexlab{b}.
\newblock \showarticletitle{AutoSF: Searching scoring functions for knowledge
  graph embedding}. In \bibinfo{booktitle}{\emph{ICDE}}. IEEE,
  \bibinfo{pages}{433--444}.
\newblock


\bibitem[\protect\citeauthoryear{Zhu, Zhang, Xhonneux, and Tang}{Zhu
  et~al\mbox{.}}{2021}]%
        {zhu2021neural}
\bibfield{author}{\bibinfo{person}{Z. Zhu}, \bibinfo{person}{Z. Zhang},
  \bibinfo{person}{L. Xhonneux}, {and} \bibinfo{person}{J. Tang}.}
  \bibinfo{year}{2021}\natexlab{}.
\newblock \showarticletitle{Neural Bellman-Ford Networks: A General Graph
  Neural Network Framework for Link Prediction}. In
  \bibinfo{booktitle}{\emph{NeurIPS}}.
\newblock


\bibitem[\protect\citeauthoryear{Zou, Hu, Wang, Jiang, Sun, and Gu}{Zou
  et~al\mbox{.}}{2019}]%
        {zou2019layer}
\bibfield{author}{\bibinfo{person}{D. Zou}, \bibinfo{person}{Z. Hu},
  \bibinfo{person}{Y. Wang}, \bibinfo{person}{S. Jiang}, \bibinfo{person}{Y.
  Sun}, {and} \bibinfo{person}{Q. Gu}.} \bibinfo{year}{2019}\natexlab{}.
\newblock \showarticletitle{Layer-Dependent Importance Sampling for Training
  Deep and Large Graph Convolutional Networks}.
\newblock \bibinfo{journal}{\emph{NeurIPS}}  \bibinfo{volume}{32}
  (\bibinfo{year}{2019}), \bibinfo{pages}{11249--11259}.
\newblock


\end{thebibliography}

\clearpage

\appendix

% toc
%\etocdepthtag.toc{mtappendix}
%\etocsettagdepth{mtchapter}{none}
%\etocsettagdepth{mtappendix}{subsection}
%\renewcommand{\contentsname}{Appendix}
%\tableofcontents

\vspace{+10pt}

\noindent
\textbf{Notations.}
In this paper, vectors are denoted by lowercase boldface, e.g., $\bm m, \bm h, \bm w$.
Matricies are denoted by uppercase boldface, e.g., $\bm W$.
Sets are denoted by script fonts, e.g,. $\mathcal E, \mathcal{V}$.
%We summarize the frequently used notations in Tab~\ref{tab:notations}.
As introduced in Section~\ref{sec:method},
a knowledge graph is in the form of $\mathcal K\!=\!\{\mathcal{V}, \mathcal R, \mathcal E, \mathcal Q \}$.
\section{Details of works in the literature}
\label{app:review}

\subsection{Summary of $\text{MESS}(\cdot)$ and $\text{AGG}(\cdot)$}
%\subsection{$\text{MESS}(\cdot)$ and $\text{AGG}(\cdot)$ for GNN-based methods}
\label{app:mess-agg}

\begin{table*}[t]
\centering
\vspace{-8px}
\caption{Summary of the GNN functions for message propagation.
	The values in $\{\}$ represent different operation choices that are tuned as hyper-parameters 
	for different datasets.}
\label{tab:GNN_functions}
\vspace{-10px}
\setlength\tabcolsep{3pt}
\begin{tabular}{c|C{270px}|C{150px}}
	\toprule
	method  & $\bm m_{(e_s, r, e_o)}^\ell:=\text{MESS}(\cdot)$ &  $\bm h_{e_o}^\ell:=\text{AGG}(\cdot)$ \\
	\midrule 
	R-GCN~\cite{schlichtkrull2018modeling}  & $\bm{W}^\ell_r \bm{h}^{\ell-1}_{e_s}$, where $\bm W_r$ depends on the relation $r$  & $\bm{W}^\ell_o \bm{h}^{\ell-1}_{e_o} \!+\!\sum_{e_o\in\mathcal N(e_s)} \frac{1}{c_{o,r}} \bm m_{(e_s, r, e_o)}^\ell$  \\
	\midrule
	CompGCN~\cite{vashishth2019composition}  &  $\bm{W}^\ell_{\lambda(r)} \{-,*,\star \}(\bm{h}^{\ell-1}_{e_s}, \bm{h}^{\ell}_{r})$, where $\bm{W}^\ell_{\lambda(r)} $ depends on the direction of $r$  &  $\sum_{e_o\in\mathcal N(e_s)} \bm m_{(e_s, r, e_o)}^\ell$ \\
	\midrule
	GraIL~\cite{teru2019inductive}  &  ${\alpha}^\ell_{(e_s, r, e_o)|r_q} (\bm{W}^\ell_1 \bm{h}^{\ell-1}_{e_s}+\bm{W}^\ell_2\bm h_{e_o}^{\ell-1})$, where  ${\alpha}^\ell_{(e_s, r, e_o)|r_q}$ is the attention weight.	&  $\bm{W}^\ell_o \bm{h}^{\ell-1}_{e_o} \!+\!\sum_{e_o\in\mathcal N(e_s)} \frac{1}{c_{o,r}} \bm m_{(e_s, r, e_o)}^\ell$
	\\
	\midrule
	NBFNet~\cite{zhu2021neural}  &  $\bm W^\ell \{+,*,\circ \}(\bm h_{e_s}^{\ell-1}, \bm w_q(r, r_q))$, where 
	$\bm w_q(r, r_q)$ is a query-dependent weight vector  &  
	%$\phi_{\text{AGG}} (\{\bm m_{(e_s, r, e_o)}^\ell\} \cup \{\bm{h}^{\ell}_{e_o})\})$, $\phi_{\text{AGG}} \! \in \! \{ {\texttt{Sum, Mean, Max, PNA}} \}$ 
	$\!\!\{{\texttt{Sum,Mean,Max,PNA}} \}_{e_o\in\mathcal N(e_s)} \bm m_{(e_s, r, e_o)}^\ell $ \\
	\midrule
	RED-GNN~\cite{zhang2021knowledge}  &  $\alpha_{(e_s,r,e_o)|r_q}^\ell (\bm{h}^{\ell-1}_{e_s} + \bm{h}^{\ell}_{r})$, where $\alpha_{(e_s,r,e_o)|r_q}^\ell$ is the attention weight  &   $ \sum_{e_o\in\mathcal N(e_s)} \bm m_{(e_s, r, e_o)}^\ell$ \\
	\midrule
	AdaProp &  $\alpha_{(e_s,r,e_o)|r_q}^\ell \{ +, *, \circ\}(\bm{h}^{\ell-1}_{e_s}, \bm{h}^{\ell}_{r})$, where $\alpha_{(e_s,r,e_o)|r_q}^\ell$ is as in ~\cite{zhang2021knowledge}.
	%		and $\phi_{\text{MESS}} \! \in \! \{ +, \times, \texttt{RotatE}\}$ 
	&   $\!\!\{{\texttt{Sum,Mean,Max}} \}_{e_o\in\mathcal N(e_s)} \bm m_{(e_s, r, e_o)}^\ell $  \\
	\bottomrule
\end{tabular}
%\vspace{-5px}
\end{table*}

The two functions of the GNN-based methods
are summarized in Table~\ref{tab:GNN_functions}.
The main differences of these methods
are the combinators for entity and relation representations on the edges,
the operators on the different representations,
and the attention weights.
Recent works \cite{vashishth2019composition,zhu2021neural} have shown 
that the different combinators and operators only have slight influence on the performance.
Comparing GraIL and RED-GNN,
even though the message functions and aggregation functions are similar,
their empirical performances are quite different with different propagation patterns.
Hence,
the design of propagation path is the \textit{key factor} influencing the performance of different methods.

\subsection{Discussion on constrained propagation}
%\subsection{Discussion on the constrained propagation GraIL \cite{teru2019inductive}}
\label{app:grail}

In KG reasoning,
the key objective is to
find the target answer entity $e_a$ given $(e_q, r_q, ?)$.
To fulfill this goal,
the model is required to evaluate multiple target answer entities.
Since
all the entities in the KG can be potential answers,
we need to evaluate $|\mathcal V|$ entities for a single query $(e_q, r_q, ?)$.
%
%For full propagation methods,
%the high-level embeddings of all the entities can be obtained in a single forward pass.
%Then the evaluation on all the candidate answers can be efficient
%since computing the score based on high-level embedding is cheap.
%For progressive propagation,
%the candidates can also be evaluated in a single forward pass
%as the last step representations are directly used to score the candidates.
%
However,
for the constrained propagation method GraIL~\cite{teru2019inductive} 
and CoMPILE~\cite{mai2021communicative},
the propagation path depends on both the query entity $e_q$
and the answer entity $e_a$,
i.e., $\mathcal V^\ell = \mathcal V^L_{e_q,e_a}$.
Hence,
it requires $|\mathcal V|$ times of forward passes
to evaluate all the entities for a single query,
which is extremely expensive
when the number of entities is large.
For the settings of transductive reasoning,
all datasets have tens of thousands of entities.
Thus, running GraIL or CoMPILE on them is intractable.

\section{Details of the sampling method}

\subsection{Augmentation with inverse relations}
\label{app:inverse}

As aforementioned,
predicting missing head in KG can also be formulated in this way by adding inverse relations.
This is achieved by
augmenting the original KG by the reverse relations and the identity relation,
as adopted in path-based methods~\cite{sadeghian2019drum, qu2021rnnlogic}.
Taking Figure~\ref{fig:kg} as example,
we add the reverse relation "part\_of inv"
to the original relation "part\_of",
which leads to a new triplet \textit{(Lakers, part\_of inv, LeBron)}.
We also add the identity relation as the self-loop,
which leads to new triplets, such as \textit{(LeBron, identity, LeBron)}.

\subsection{Proofs of propositions}

\subsubsection{Proposition~\ref{pr:complexity}}
Since $|\mathcal V^0_{e_q,r_q}| = 1$, $\mathcal{V}^\ell_{e_q,r_q}:=\mathcal{V}^{\ell-1}_{e_q,r_q} \cup \texttt{SAMP}(\overline{\mathcal{V}}^\ell_{e_q,r_q})$ and $|\texttt{SAMP}(\overline{\mathcal{V}}^\ell_{e_q,r_q})| \leq K$,
we have 
\[ \mathcal{V}^\ell_{e_q,r_q} \leq 1 + \ell\cdot K,\]
for $\ell=1\dots L$.
Meanwhile, with the constraint $\mathcal{V}^0_{e_q,r_q} \! \subseteq \! \mathcal{V}^1_{e_q,r_q} \! \cdots \! \subseteq \! \mathcal{V}^L_{e_q,r_q}$,
we have 
\[\big|\bigcup_{\ell=0\dots L} \mathcal V^\ell_{e_q,r_q}\big| = \big| \mathcal V^L_{e_q,r_q}\big| \leq 1 + L\cdot K.\]
Namely,
the number of involved entities is bounded by $O(L\cdot K)$.

\subsubsection{Proposition~\ref{pr:preserve}}
Since $\mathcal{V}^\ell_{e_q,r_q}:=\mathcal{V}^{\ell-1}_{e_q,r_q} \cup \texttt{SAMP}(\overline{\mathcal{V}}^\ell_{e_q,r_q})$,
we have $\mathcal{V}^{\ell-1}_{e_q,r_q} \subseteq \mathcal{V}^\ell_{e_q,r_q}$.
With the introduction of identity relation in Section~\ref{app:inverse},
for each $e\in\mathcal{V}^{\ell-1}_{e_q,r_q}$,
there exist at least one entity $e\in \mathcal{V}^{\ell}_{e_q,r_q}$
such that $(e, \textit{identity}, e) \in \mathcal E^\ell$.
Thus,
the connection can be preserved.

\subsection{Adapting sampling methods for KG}
\label{app:sampling-for-kg}

When modeling the homogeneous graphs,
the sampling methods are designed 
to solve the scalability problem.
Besides,
they are mainly used in the node classification task,
which is quite different from the link-level KG reasoning task here.
Hence,
we introduce the adaptations we made
to use the sampling methods designed for homogeneous graphs
in the KG reasoning scenarios as follows.
\vspace{-2pt}
%\begin{itemize}[leftmargin=18px, itemsep=3pt,topsep=2pt,parsep=3pt,partopsep=0pt]
\begin{itemize}[leftmargin=*]
\item Node-wise sampling. 
The design in this type is based on GraphSAGE \cite{hamilton2017inductive}.
Given $\mathcal V^{\ell-1}_{e_q,r_q}$,
we sample $K$ neighboring entities from $\mathcal N(e)$
for entities $e\in \mathcal V^{\ell-1}_{e_q,r_q}$.
Then the union of these sampled entities
forms $\mathcal V^\ell_{e_q,r_q}$ for next step propagation.
For the not-learned version,
we randomly sample from $\mathcal N(e)$.
For the learned version,
we sample from $\mathcal N(e)$ based on the distribution $p^\ell(e) \sim g(\bm h^\ell_e;\bm \theta^\ell)$.

\item Layer-wise sampling.
The design in this type is based on LADIES \cite{zou2019layer}.
Given $\mathcal V^{\ell-1}_{e_q,r_q}$,
we firstly collect the set of neighboring entities 
$\mathcal N(\mathcal V^{\ell-1}) = \bigcup_{e\in\mathcal V^{\ell-1}_{e_q,r_q}}\mathcal N(e)$.
Then,
we sample $K$ entities without replacement from $\mathcal N(\mathcal V^{\ell-1}_{e_q,r_q})$
as $\mathcal V^\ell$ for next step propagation.
For the not-learned version,
we sample according to the degree distribution for entities in $\mathcal N(\mathcal V^{\ell-1}_{e_q,r_q})$.
For the learned version,
we sample from $\mathcal N(\mathcal V^{\ell-1}_{e_q,r_q})$ based on the distribution $p^\ell(e)\sim g(\bm h^\ell_e;\bm \theta^\ell)$.

\item Subgraph sampling.
Given the query $(e_q, r_q, ?)$,
we generate multiple random walks starting from the query entity $e_q$
and use the sampled entities to induce the subgraph $\mathcal V_{e_q}$.
Then, the propagation starts from $e_q$ like the progressive propagation,
but it is constrained within the range of $\mathcal V_{e_q}$.
Since learning to sample the subgraph is challenging,
we only evaluate the subgraph sampling in the not-learned version.
\end{itemize}

\subsection{Implementation of ST estimator}
\label{app:st}
%Considering the discrete nature of the 
%propagation path $\hat{\mathcal G}^L\!(\bm \theta)$,
%the direct gradient of loss function $\mathcal L$
%w.r.t. the sampling parameters $\bm \theta$
%is intractable.
%The reparameterization technique for Gumbel distributions~\cite{jang2017categorical}
%does not do explicit sampling.
%Thus, it is not appropriate for the case here.
%Instead,
%we choose the straight-through (ST) estimator~\cite{bengio2013estimating,jang2017categorical},
%which can approximately estimate gradient 
%for discrete variables.
The key idea of straight-through (ST) estimator is to back-propagate 
through the sampling signal
as if it was the identity function.
Given the sampled entities 
$e_o \in \widetilde{\mathcal{V}}^\ell_{e_q,r_q}$
as well as their sampling probability $p^\ell\!(e)$,
the derivative of the sampling parameters can be 
obtained by multiplying $p^\ell(e)$ with the hidden states $\bm h_{e}^\ell$.
Specifically,
rather than directly use 
%the entity representations $\bm h_{e}^\ell$
the $\bm h_{e}^\ell$
to compute messages,
we use
\begin{equation}
\bm h_{e}^\ell \; :=  \; \big( 1 \;  - \;  \texttt{no\_grad}(p^\ell\!(e))  \; + \; p^\ell\!(e) \big) \; \cdot \; \bm h_{e}^\ell,
\label{eq:ST}
\end{equation}
where $\texttt{no\_grad}(p^\ell\!(e))$ means that the back-propagation signals will not go through this term.
In this approach,
the forward computation will not be influenced, while
the backward signals can go through the multiplier $p^\ell\!(e)$.
Besides,
the backward signals, even though not exactly equivalent to the true gradient,
is in positive propagation to it.
Hence,
we can use such a backward signal to approximately 
estimate the gradient of $\mathcal L$ with regard to $\bm\theta$.

\section{Further materials of experiments}
%\section{Supplementary materials of experiments}
\label{app:exp}

\subsection{Hop distributions of datasets}
\label{app:distance}
%\footnote{$\surd$+yq+: write something here}

We summarize the distance distribution
of test queries in Table~\ref{tab:distance_distribution}
for the largest four KGs.
Here, the distance for a query triple $(e_q, r_q, e_a)$
is calculated as the length of the shortest path
connecting $e_q$ and $e_a$ from the training triples.
As shown,
\textit{most answer entities are close to query entities},
i.e., within the 3-hop neighbors of query entities.

\begin{table}[ht]
	\vspace{-6px}
	\centering
	\caption{
		Distance distribution (in \%)
		of queries in $\mathcal{Q}_{tst}$.
	}
	\vspace{-10pt}
	\label{tab:distance_distribution}
%	\fontsize{8}{6}\selectfont
%	\small 
	\setlength\tabcolsep{8pt}
	\begin{tabular}{c|cccccc}
		\toprule
		distance & 1 & 2 &  3 &  4 &  5 &  >5 \\
		\hline
		WN18RR   & 34.9 & 9.3  &  21.5  &  7.5 &  8.9  &  17.9  \\
		FB15k237 & 0.0 & 73.4  &  25.8  &  0.2 &  0.1  &  0.5 \\
		NELL-995 & 40.9 & 17.2  &  36.5  &  2.5 &  1.3  &  1.6 \\
		YAGO3-10 & 56.0 & 12.9 & 30.1 & 0.5 & 0.1  & 0.4  \\
		\bottomrule
	\end{tabular}
%\vspace{-6px}
\end{table}

\subsection{Hyper-parameters}
\label{app:HP}

%We tune the number of
%propagation steps $L$ among $\{5,6,7,8\}$,
%the number of incrementally sampled entities $K$  among $\{100, 500, 1000, 2000\}$,
%temperature $\tau$ in $\{0.5, 1.0, 2.0\}$,
%batch size in $\{20,50,100\}$,
%$\texttt{MESS}(\cdot)$ in $\{+,*,\circ\}$
%and 
%$\texttt{AGG}(\cdot)$ in $\{\texttt{Sum, Mean, Max} \}$
%The ranges of the other hyper-parameters are kept the same as RED-GNN.
We select the optimal hyper-parameters for evaluation
based on the MRR metric or the Hit@10 metric on validation set
for transductive setting or indcutive setting, respectively.
We provide the hyper-parameters of 
$L$, $K$, $\tau$ and batch size in Table~\ref{tab:HP_transductive}.
%and Tab~\ref{tab:HP_inductive}
%for transductive and inductive setting, respectively.
We observe that
(1) In most cases, we have a larger number of propagation steps with $L\geq6$.
(2) In the transductive setting, the values of $K$ are larger since the KGs are larger.
But in the inductive setting, $K$'s are generally small.
This means that $K$ is not as large as possible.
More irrelevant entities will influence the performance.
(3) There is no much regularity in the choices of $\tau$,
which depends on specific KGs.
%(4) The sum aggregator is always the best.
%
%\begin{itemize}[leftmargin=10px]
%\item In most of the datasets, we have a larger number of propagation steps with $L\geq6$.
%\item In the transductive setting, the values of $K$ are larger since the KGs are larger.
%But in the inductive setting, $K$'s are generally small.
%This means that $K$ is not as large as possible.
%More irrelevant entities will influence the performance.
%\item There is no much regularity in the choices of $\tau$.
%\item The sum aggregator is always the best.
%\end{itemize}

\begin{table}[ht]
%\vspace{-6px}
\centering
\caption{Optimal hyper-parameter (HP) for each dataset. ``BS'' is short for batch size.}
\vspace{-10px}
\label{tab:HP_transductive}
\fontsize{10}{10}\selectfont
\setlength\tabcolsep{1pt}
\begin{tabular}{c|c|c|c|c|c|c}
	\toprule
	HP  & Family &  UMLS &  WN18RR  &  FB15k237  &  NELL-995  & YAGO3-10   \\ \midrule
	$L$		 &	8  & 5  & 8	&	7	&	6  & 8	\\
	$K$		 &	100 & 100  & 1000	&	2000	&	2000  &	1000 \\
	$\tau$   &1.0	& 1.0  & 0.5	&	2.0	&	0.5	 &	1.0 \\
	BS  & 20	& 10  & 	100	&	50	&	20	&  5 \\
%	$\text{MESS}(\cdot)$ &	$*$	&	$+$	&	$*$	 \\
%	$\text{AGG}(\cdot)$	&	\texttt{SUM}	&	\texttt{SUM}	&	\texttt{SUM}	 \\	
	\midrule
\end{tabular}
\setlength\tabcolsep{2.3pt}
\begin{tabular}{c|cccc|cccc|cccc}
	\midrule
	& \multicolumn{4}{c}{WN18RR}  &  \multicolumn{4}{|c|}{FB15k237} & \multicolumn{4}{c}{NELL-995} \\ 
	HP  & V1 & V2 & V3 & V4 & V1 & V2 & V3 & V4 & V1 & V2 & V3 & V4 \\ \midrule
	$L$		 &	7	&	7	&	7	&	6	&	8	&	6	&	7	&  	7	 &	6	&	5	&	8	& 	5	 \\
	$K$		 &	200	&	100	&	200	&	200	&	200	&	200	&	200	&  	500	 &	200	&	200	&	500	& 	100	 \\
	$\tau$		&	0.5	&	0.5	&	1.0	&	1.0	&	0.5	&	2.0	&	1.0	&  1.0		 &	0.5	&	1.0	&	1.0	& 	2.0	 \\
	BS  &	20	&	20	&	50	&	50	&	50	&	20	&	20	&  	50	 &	20	&	50	&	20	& 	50	 \\
	%	$\text{MESS}(\cdot)$ &	$+$ &	$+$		&	$+$		&	$+$		&	$+$		&	$+$		&	$+$		&	$+$		&	$+$		&	$+$		&	$+$		&	$+$		\\
	%	$\text{AGG}(\cdot)$	&	\texttt{SUM}	&	\texttt{SUM}	&	\texttt{SUM}	&	\texttt{SUM}	&	\texttt{SUM}	&	\texttt{SUM}	&	\texttt{SUM}	&	\texttt{SUM}	&	\texttt{SUM}	&	\texttt{SUM}	&	\texttt{SUM}	&	\texttt{SUM}  \\
	\bottomrule
\end{tabular}
%\vspace{-10px}
\end{table}

\subsection{Hit@1 results of inductive reasoning}
\label{app:induc_full}

The full results of inductive reasoning,
for MRR and Hit@1 metrics
are shown in Table~\ref{tab:induc-full}.
%In the other metrics,
%i.e., MRR and Hit@1,
AdaProp is also the best in most of the cases.

\begin{table*}[ht]
	\vspace{-5px}
\centering
\caption{Inductive setting on MRR and Hit@1.}
\label{tab:induc-full}
%\tiny
%\fontsize{6}{9}\selectfont
\setlength\tabcolsep{5pt}
\vspace{-10px}
\begin{tabular}{cc|cccc|cccc|cccc}
	\hline
	\multirow{2}{*}{metrics}&    \multirow{2}{*}{methods}   & \multicolumn{4}{c|}{WN18RR} & \multicolumn{4}{c|}{FB15k237}  & \multicolumn{4}{c}{NELL-995} \\
	&          & V1    & V2   & V3   & V4   & V1    & V2    & V3    & V4    & V1  & V2  & V3  & V4  \\   
			\midrule 
				\multirow{7}{*}{MRR}   
			& RuleN   &  0.668   & 0.645  & 0.368  & 0.624  &  \underline{0.363}  & 0.433  &  0.439  & 0.429  & 0.615 &  0.385& 0.381 &  0.333\\
			&  Neural LP &	0.649	&	0.635	&	0.361	&	0.628	&	0.325	&	0.389	&	0.400	&	0.396	&	0.610	&   0.361 & 0.367  &   0.261	\\
			& DRUM      &  0.666  &  0.646    &   0.380   &   0.627   &   0.333    &  0.395     &   0.402    &   0.410    &  0.628  &  0.365  & 0.375  &  0.273  \\
			\cmidrule{2-14}
			& GraIL     &    0.627    &   0.625   &   0.323   &  0.553    &  0.279     &   0.276    &    0.251   &    0.227  &  0.481 & 0.297 &  0.322  & 0.262  \\
			& CoMPILE  &  0.577     &   0.578  &  0.308   &  0.548    &   0.287    &   0.276    &    0.262   &  0.213     &  0.330  &     0.248 &  0.319   &  0.229  \\ 
			& {RED-GNN}      &   \underline{0.701}  &  \underline{0.690}   &  \underline{0.427}    &   \underline{0.651}   &     \textbf{0.369}     &  \underline{0.469}     &  \underline{0.445}     &  \underline{0.442}  &  \underline{0.637}   &  \underline{0.419} &   \textbf{0.436}  &  \underline{0.363}   \\    
			& NBFNet     &    0.684    &   0.652   &   0.425   &  0.604    &  0.307     &   0.369   &    0.331   &    0.305  &  0.584 & 0.410 &  0.425  & 0.287  \\
			\cmidrule{2-14}
			&    \textbf{AdaProp} &  \textbf{0.733}    &  \textbf{0.715}  &  \textbf{0.474}   &  	\textbf{0.662}   &  {0.310} & \textbf{0.471}  &  \textbf{0.471} &  \textbf{0.454}   & \textbf{0.644} & \textbf{0.452} &  \underline{0.435} & \textbf{0.366}    \\
			\midrule
	\multirow{7}{*}{Hit@1 (\%)}
	& RuleN   &  63.5   &   61.1  &  34.7  &  59.2  &  \textbf{30.9}  & 34.7 & 34.5 & 33.8  & \textbf{54.5}&30.4& 30.3  & \underline{24.8}\\
	&  Neural LP &	59.2	&	57.5	&	30.4	&	58.3	&	24.3	&	28.6	&	30.9	&	28.9	&	50.0	&		24.9   &	26.7 	& 13.7	\\
	& DRUM      &  61.3  &   59.5   &    33.0  &  58.6    &  24.7     &   28.4    &   30.8    &   30.9    & 50.0  &  27.1  &  26.2  &   16.3  \\  \cline{2-14}
	& GraIL     &   55.4    &    54.2  &   27.8   &    44.3  &   20.5    &   20.2    &   16.5    &    14.3   &  42.5  & 19.9  &   22.4  &  15.3  \\ 
	& CoMPILE  &    47.3   &  48.5   &   25.8  & 47.3     &    20.8   &  17.8     &   16.6    &    13.4   &   10.5 &   15.6  &  22.6   & 15.9   \\ 
	& {RED-GNN}      &  \underline{65.3}     &   \underline{63.3}   &  \underline{36.8}    &   \underline{60.6}   &  \underline{30.2}     &   \underline{38.1}    &   \underline{35.1}    &  \underline{34.0}   & \textbf{52.5}   &  \underline{31.9}  &  \textbf{34.5}  &  \textbf{25.9}  \\    
	& NBFNet     &    59.2   &   57.5   &   30.4   &  57.4    &  19.0     &   22.9    &   20.6  &    18.5  &  50.0 & 27.1 &  26.2  & 23.3  \\
	\cline{2-14}
	&    \textbf{AdaProp} &   \textbf{66.8}   &  \textbf{64.2}  & \textbf{39.6}   &    \textbf{61.1}	 &  {19.1} & \textbf{37.2}  &  \textbf{37.7} &   \textbf{35.3}	&  52.2 &  \textbf{34.4}  & \underline{33.7}  &  24.7   \\
	\bottomrule
\end{tabular}
%\vspace{-6px}
\end{table*}

\subsection{Influence of temperature $\tau$}
%\footnote{$\surd$+yq+: reorganize this part}
%In Section~\ref{exp:understanding},
Besides,
we evaluate the influence of different propagation depth $L$
and number of entities $K$ in Section~\ref{sec:exp}.
Here, 
%we study the other three factors related to the sampling distribution
%in Figure\ref{fig:ablation_L_and_K} and Table\ref{tab:ablation_perf_tau}.
we show the performances of different temperatures $\tau$
in Table~\ref{tab:ablation_perf_tau}.
As can be seen,
the influence of different choices of $\tau$
is not high on the different datasets. 
This means that the sampling signal is \textit{stable}
and be adjusted by the learnable sampling weights.

\begin{table}[H]
%	\vspace{-4px}
	\vspace{-6px}
	\centering
	\caption{Model performance (MRR) of different $\tau$ values. 
	}
	\vspace{-10pt}
	\label{tab:ablation_perf_tau}
%	\small
	\setlength\tabcolsep{20pt}
	\begin{tabular}{c|c|c}
		\toprule
		$\tau$ & WN18RR & FB15k237-v1    \\ \hline
		0.5  &  0.562    &    0.551 \\
		1.0  &   0.559   &  0.546 \\
		2.0  &  0.560   &  0.549 \\
		\bottomrule
	\end{tabular}
%\vspace{-6px}
\end{table}

\subsection{Parameter size}

The parameter size is shown in Table~\ref{tab:params}.
As can be seen,
AdaProp is \textit{parameter-efficient},
although it has larger propagation steps than other GNN-based models
and has additional sampler parameters.
Since AdaProp works without entity embeddings,
the number of parameters is much smaller than the others.
%In addition,
%as AdaProp does not need to 
%learn entity-dependent parameters,
%and just hold relatively low hidden sizes of 
%$\bm h^\ell\!$  and $\bm m^\ell\!$,
%it keeps the parameter size in a small scale.

\begin{table}[ht]
%	\vspace{-10px}
	\centering
	\caption{Number of parameters of different models.}
	\label{tab:params}
	\vspace{-10px}
	\setlength\tabcolsep{17pt}
	\begin{tabular}{c|cc}
		\toprule
		methods & WN18RR         & FB15k237     \\ \midrule
		RotatE          & 20,477k        &   29,319k   \\
		CompGCN         & 12,069k        &    9,447k \\
		RED-GNN       & 58k            &       118k   \\
		NBFNet         &      89k         &     3,103k \\
		AdaProp        &      77k        &    195k   \\		
		\bottomrule
	\end{tabular}
%	\vspace{-5px}
\end{table}

\subsection{Importance of deeper propagation step}
\label{sssec:long-range}
\noindent \\
In Section~\ref{ssec: transductive},
%In Section~\ref{ssec:revisit},
we attribute the better performance 
of a deeper propagation path
to the long-range information along with a deeper GNN model.
Here,
we compare the influence of the two factors via explicit decoupling.
In Table~\ref{tab:decouple},
we show the performance of progressive propagation method RED-GNN
by alternately increasing the number of GNN-layers 
or the depth of  propagation path per row.
The propagation path $L$ with deeper GNN-layer $L+1$ is
achieved by repeating the last propagation entities,
i.e., $\mathcal V^{L+1}:=\mathcal V^L$.
%by sampling the subgraph with fixed propagation depth
%and then propagating on this enclosed subgraph.
For example, a propagation path of 2-depth
means that the message propagation is constrained in at most 2-hop neighborhoods,
and the messages cannot be propagated further even with a 3-layer GNN.
As indicated by the bold face numbers of row-wise performance improvement,
the influence of larger propagation depth
is more significant than deeper GNN-layers.
Based on the above analysis,
we conclude that \textit{the influence of long-range information
is more important than a deeper model.}

\begin{table}[ht]
	\vspace{-6px}
	\centering
%	\vspace{-6px}
	\caption{Per-hop performance of RED-GNN 
		with different GNN layers and propagation depth.
		``$\uparrow$ per row'' is computed by minimizing the MRR or Hit@10 value in the current row
		with that in the previous row.}
	\label{tab:decouple}
	\vspace{-10px}
	\setlength\tabcolsep{1.8pt}
	\begin{tabular}{cc|cc|cc|cc}
		\toprule
		\multicolumn{8}{c}{WN18RR (evaluated by MRR)} \\
		depth & layers & 1-hop & $\uparrow$ per row & 2-hop & $\uparrow$ per row & 3-hop & $\uparrow$ per row 
		\\  \midrule
		2 	&   2		&  0.940		& -- 	& 0.338		& --   & 0.106		&  -- \\   \hline
		2  &   3 		& 0.955		& 0.015	& 0.347		& 0.009  & 0.111		& 0.005 \\
		3 	&	3 		& 0.981		& \textbf{0.026	}& 0.463		& \textbf{0.116}  & 0.419		& \textbf{0.313} \\	\hline
		3   &	4 		& 0.983	& 0.002	& 0.475		& 0.012  & 0.422		& 0.316 \\
		4  	&	4		&	0.990	& \textbf{0.007}	& 0.576	& \textbf{0.101}  & 0.477		& \textbf{0.371} \\
		\midrule
	\end{tabular}
	\begin{tabular}{cc|cc|cc|cc}
		\midrule
		\multicolumn{8}{c}{FB15k237-v1 (evaluated by Hit@10)} \\
		depth & layers & 2-hop & $\uparrow$ per row & 3-hop & $\uparrow$ per row & 4-hop & $\uparrow$ per row 
		\\  \midrule
		2 	&   2    	& 50.0		& --         & 0.4		&  --         & 1.8  &  -- \\	\hline
		2   &   3       & 52.4		& 2.4  & 0.9		& 0.5  & 1.8  &  0.0 \\
		3 	&	3 		& 56.0		& \textbf{6.0}  & 23.3		& \textbf{22.9}   & 1.8  &  0.0 \\	\hline
		3   &	4 		& 56.4		& 6.4  & 23.6		& 23.2  & 1.8  &  0.0 \\
		4  	&	4		& 58.5		& \textbf{8.5}  & 37.6		& \textbf{37.2}  & 7.1 &  \textbf{5.3} \\
		\bottomrule
	\end{tabular}
%	\vspace{-10px}
\end{table}

\subsection{Visualization of propagation paths}
\begin{figure*}[ht]
	\centering
	\subfigure[AdaProp, $q_1$]{\includegraphics[height=3.4cm]{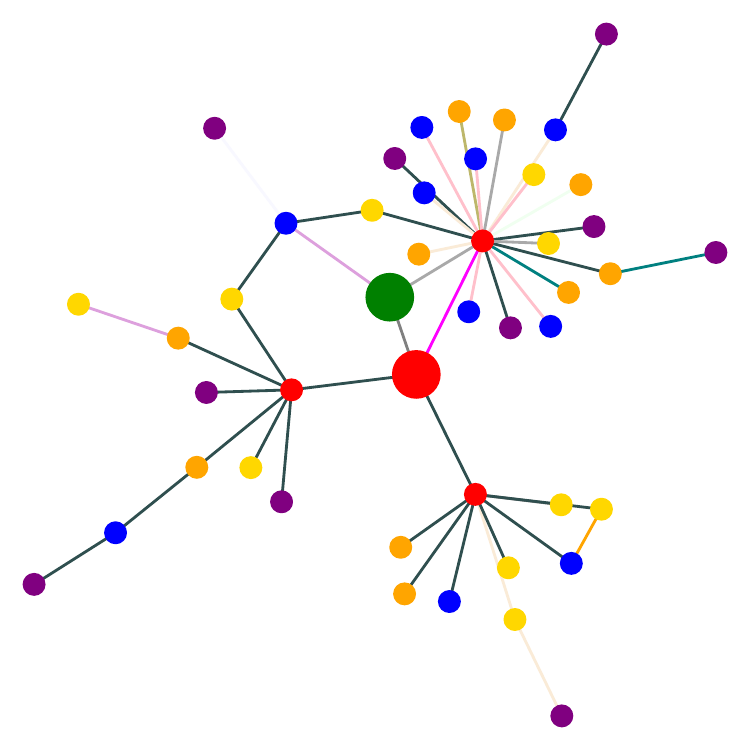}}\quad
	\subfigure[AdaProp, $q_2$]{\includegraphics[height=3.4cm]{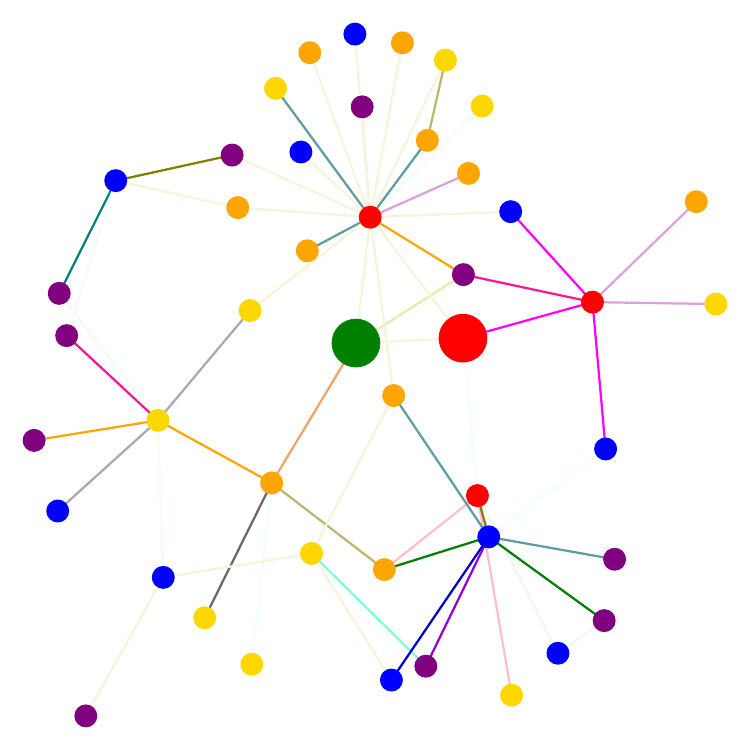}}\quad
	\subfigure[Progressive, $q_1$]{\includegraphics[height=3.4cm]{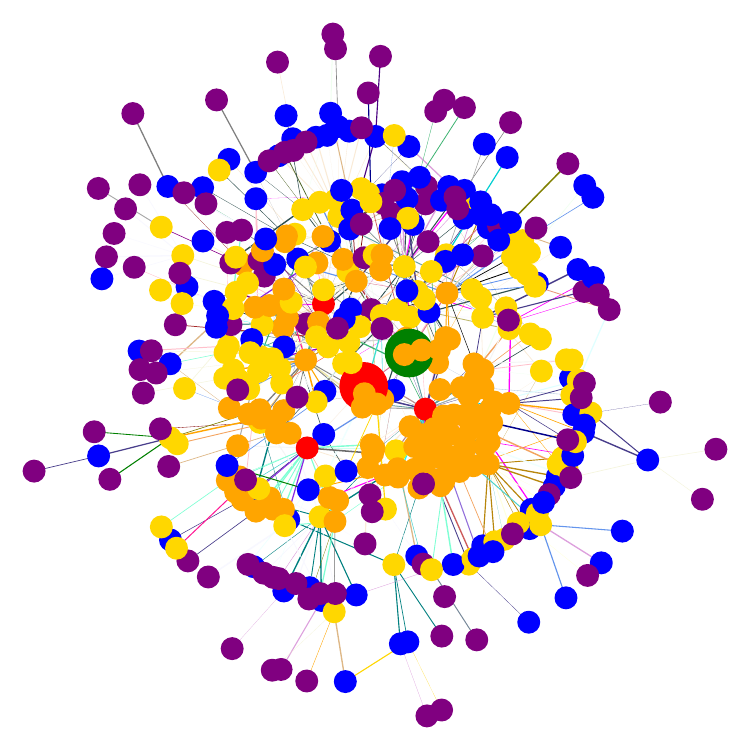}}\quad
	\subfigure[Progressive, $q_2$]{\includegraphics[height=3.4cm]{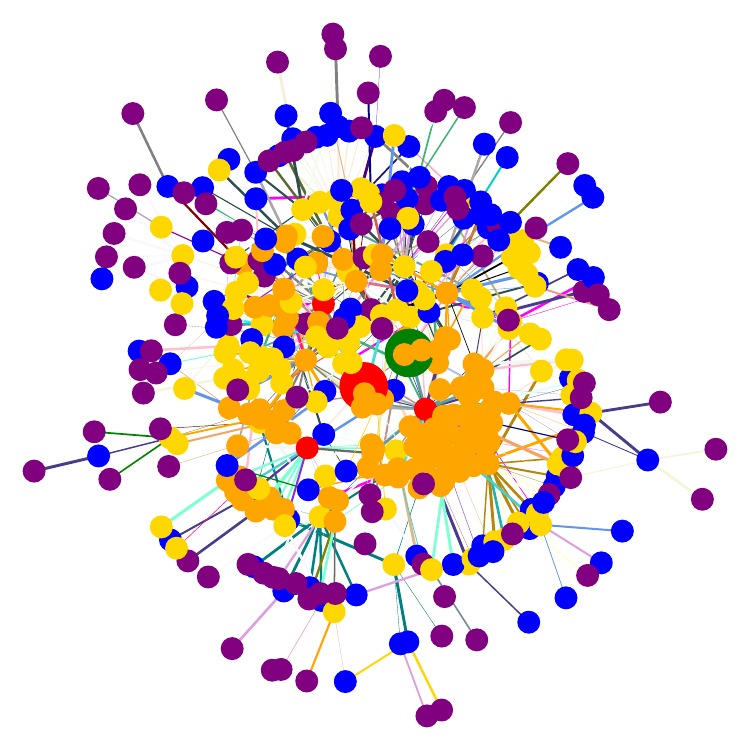}}
	\vspace{-5px}
	
	\subfigure[AdaProp, $q_1$]{\includegraphics[height=3.4cm]{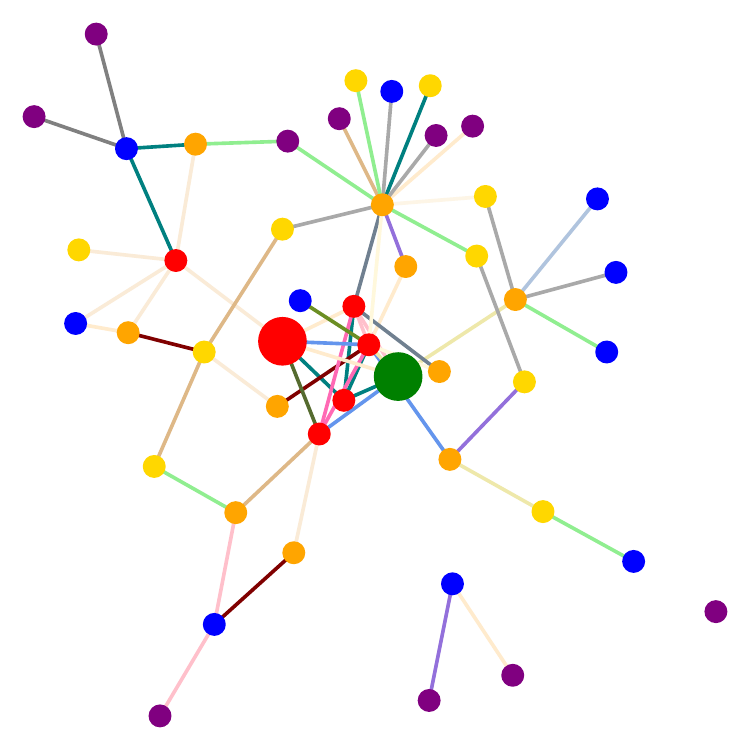}}\quad
	\subfigure[AdaProp, $q_2$]{\includegraphics[height=3.4cm]{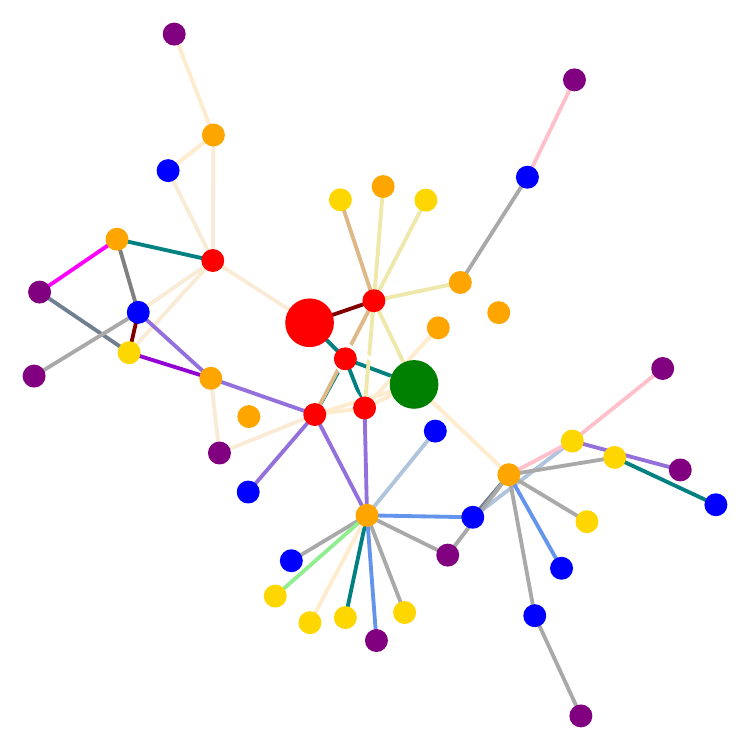}}\quad
	\subfigure[Progressive, $q_1$]{\includegraphics[height=3.4cm]{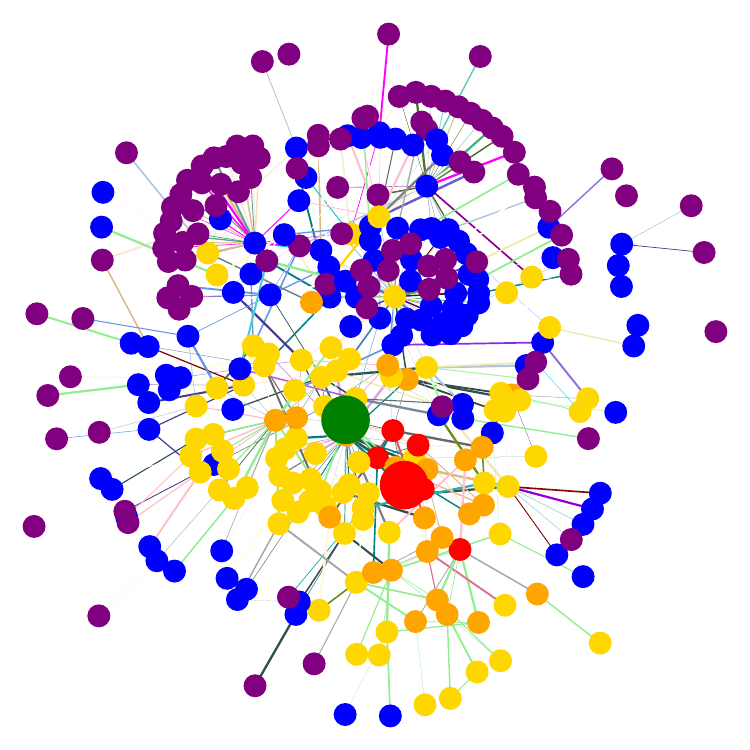}}\quad
	\subfigure[Progressive, $q_2$]{\includegraphics[height=3.4cm]{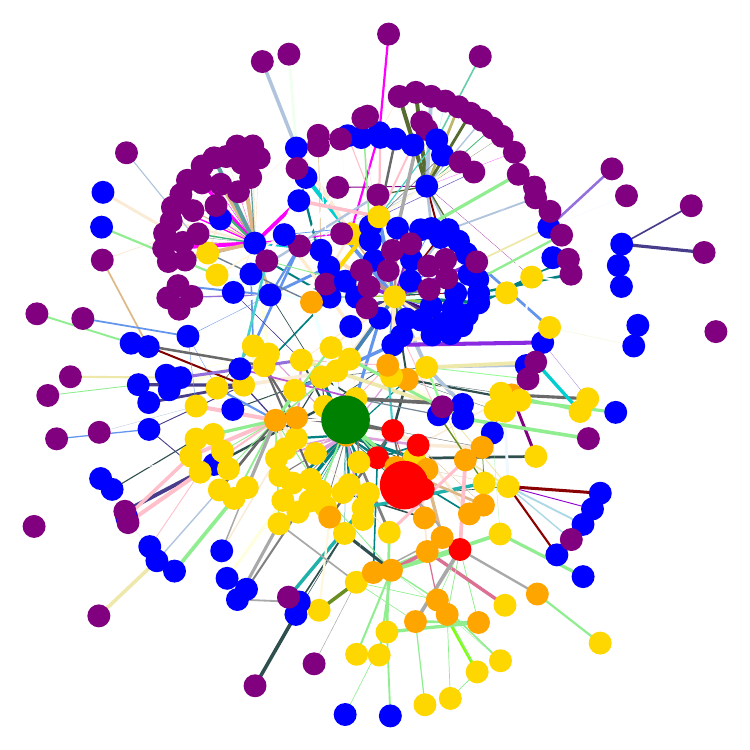}}
	\vspace{-5px}
	
	\subfigure[AdaProp, $q_1$]{\includegraphics[height=3.4cm]{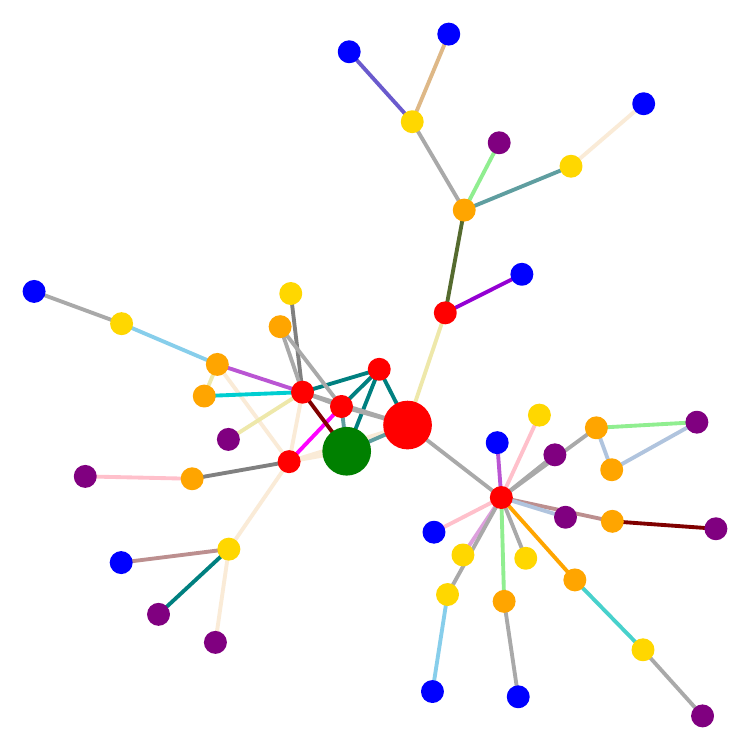}}\quad
	\subfigure[AdaProp, $q_2$]{\includegraphics[height=3.4cm]{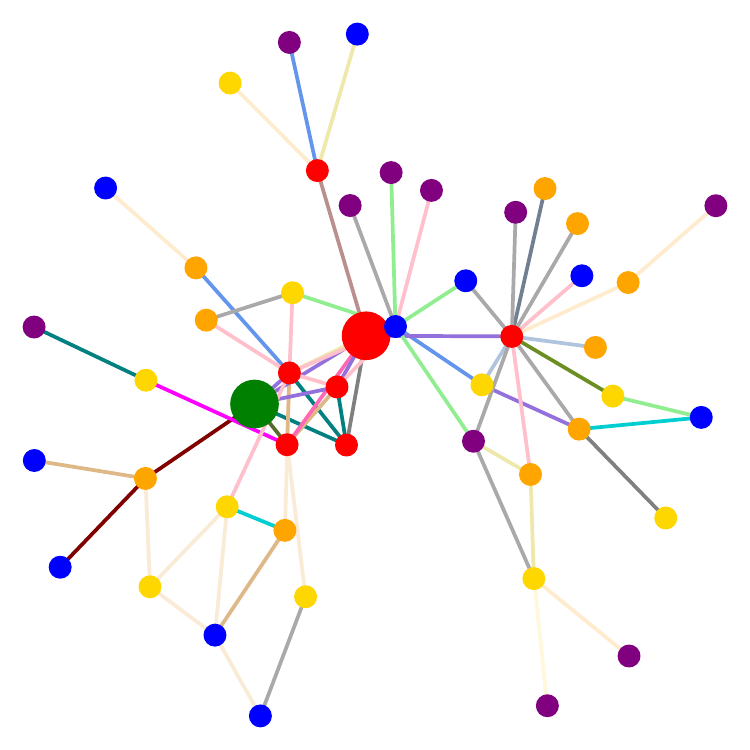}}\quad
	\subfigure[Progressive, $q_1$]{\includegraphics[height=3.4cm]{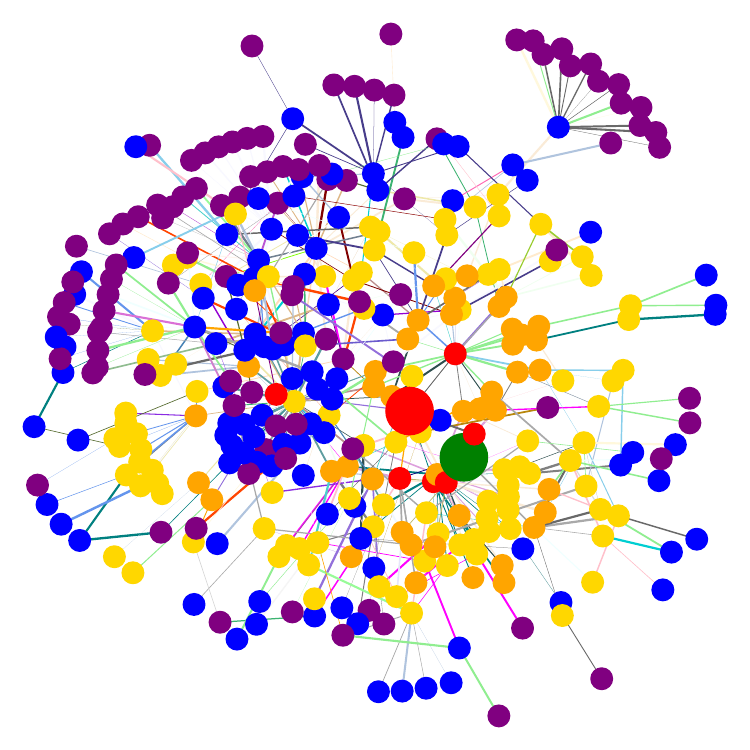}}\quad
	\subfigure[Progressive, $q_2$]{\includegraphics[height=3.4cm]{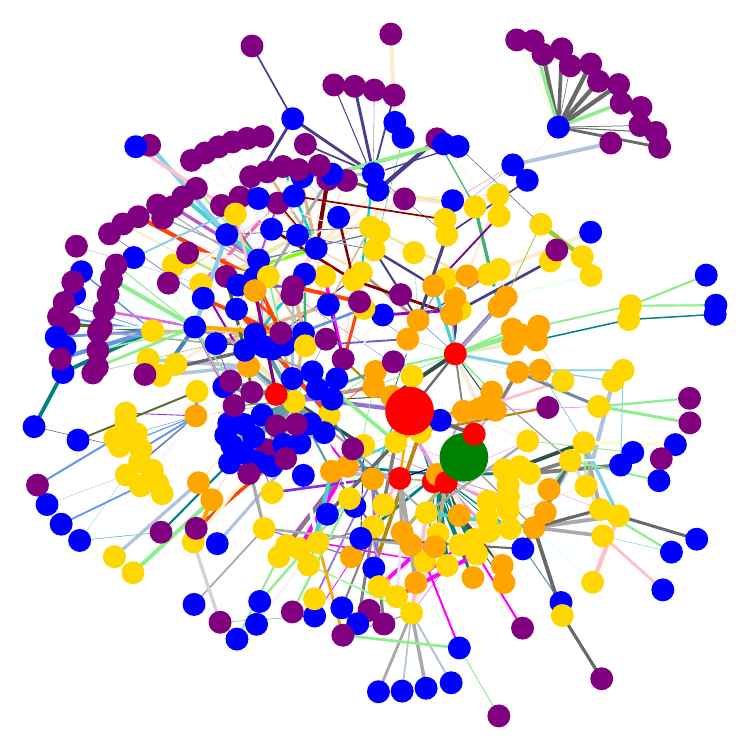}}
	\vspace{-5px}
	
	\subfigure[AdaProp, $q_1$]{\includegraphics[height=3.4cm]{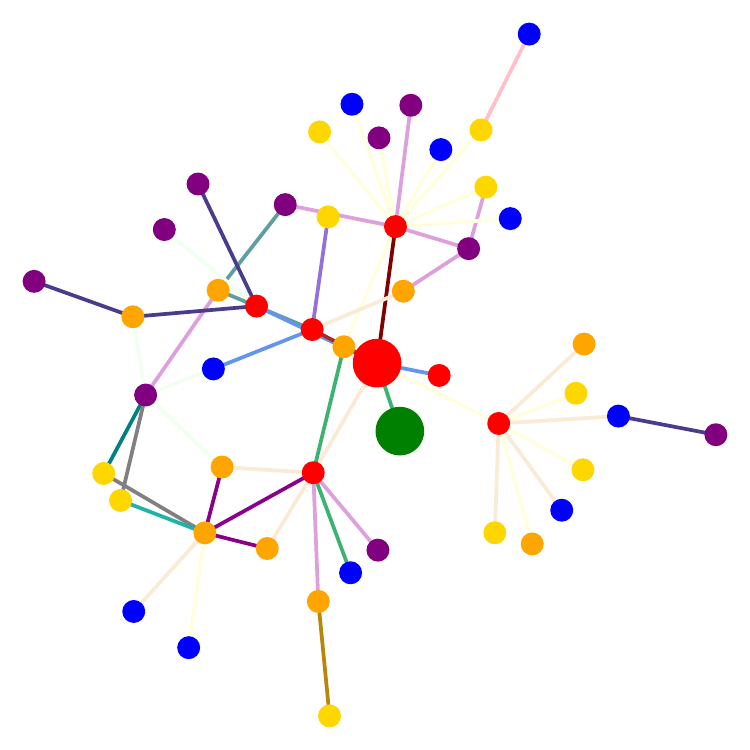}}\quad
	\subfigure[AdaProp, $q_2$]{\includegraphics[height=3.4cm]{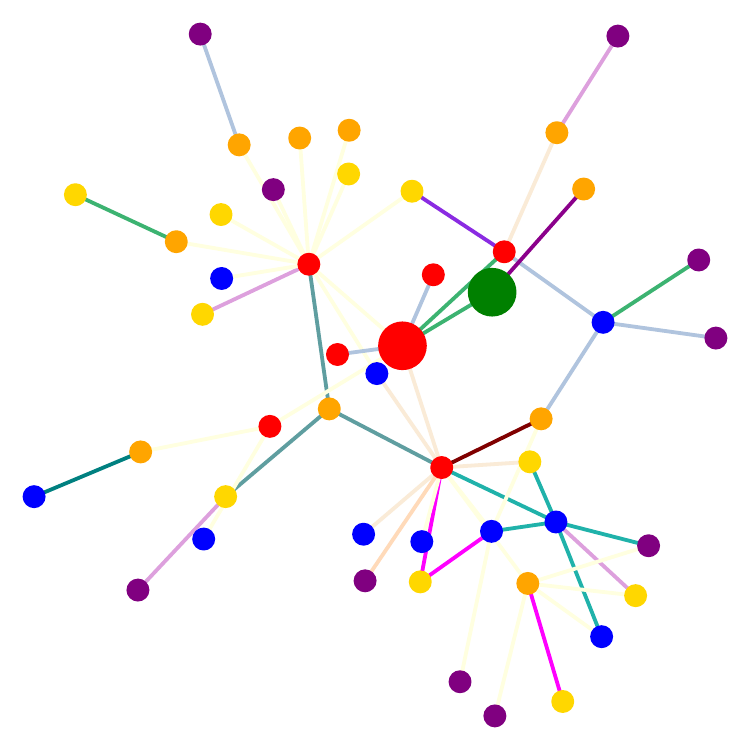}}\quad
	\subfigure[Progressive, $q_1$]{\includegraphics[height=3.4cm]{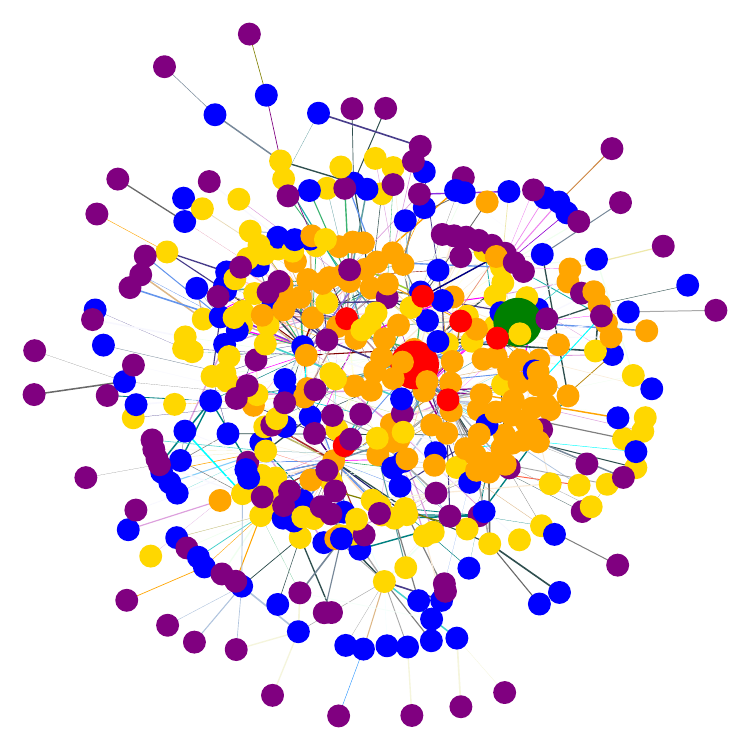}}\quad
	\subfigure[Progressive, $q_2$]{\includegraphics[height=3.4cm]{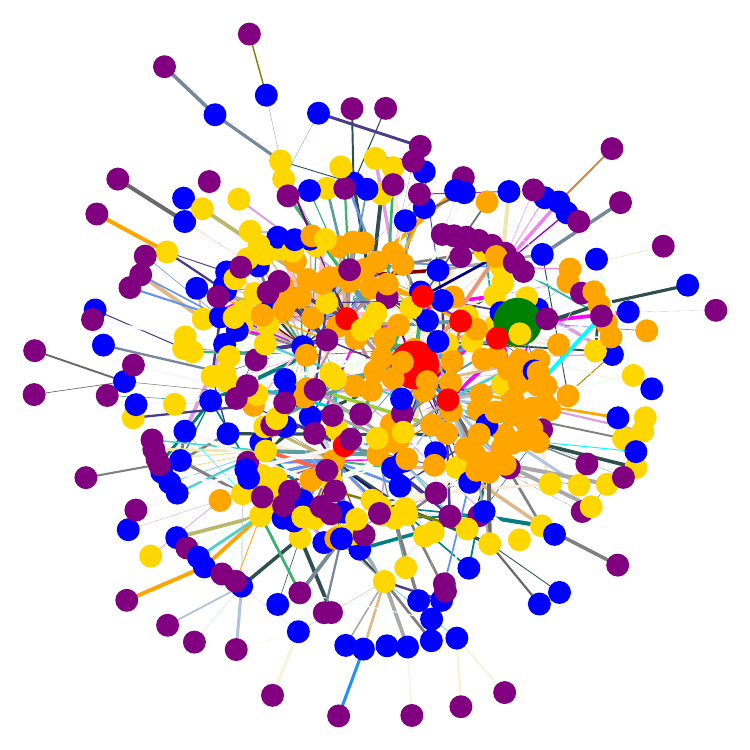}}
	\vspace{-5px}
	
	\subfigure[AdaProp, $q_1$]{\includegraphics[height=3.4cm]{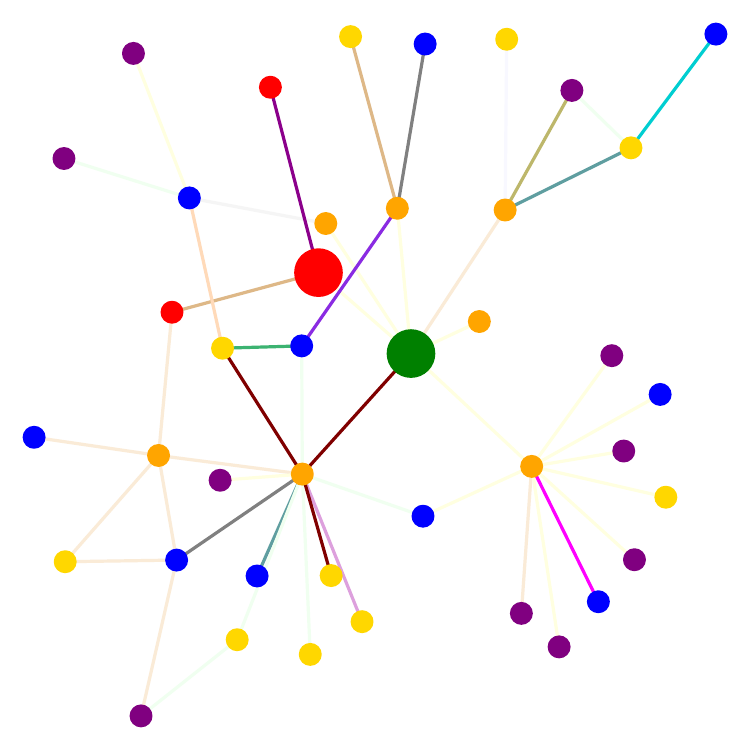}}\quad
	\subfigure[AdaProp, $q_2$]{\includegraphics[height=3.4cm]{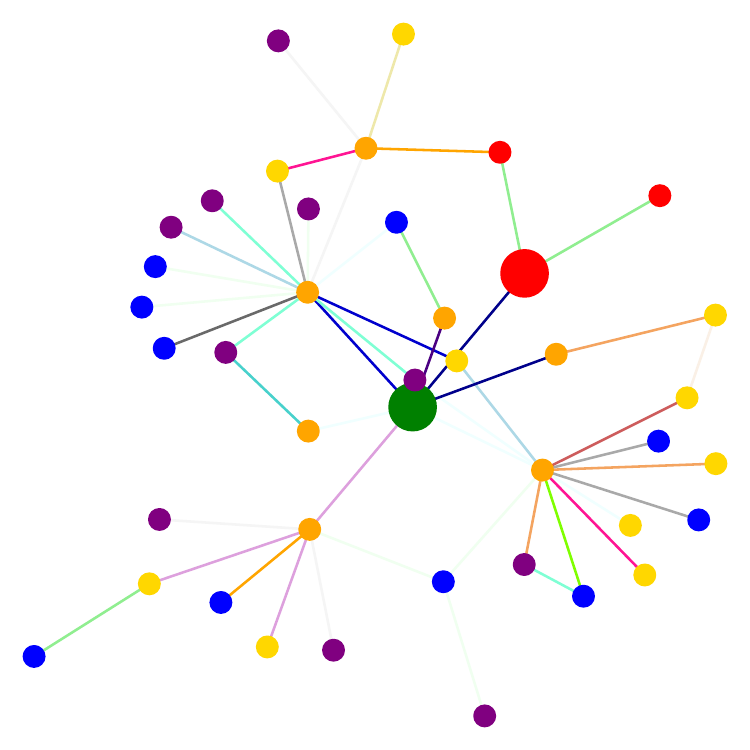}}\quad
	\subfigure[Progressive, $q_1$]{\includegraphics[height=3.4cm]{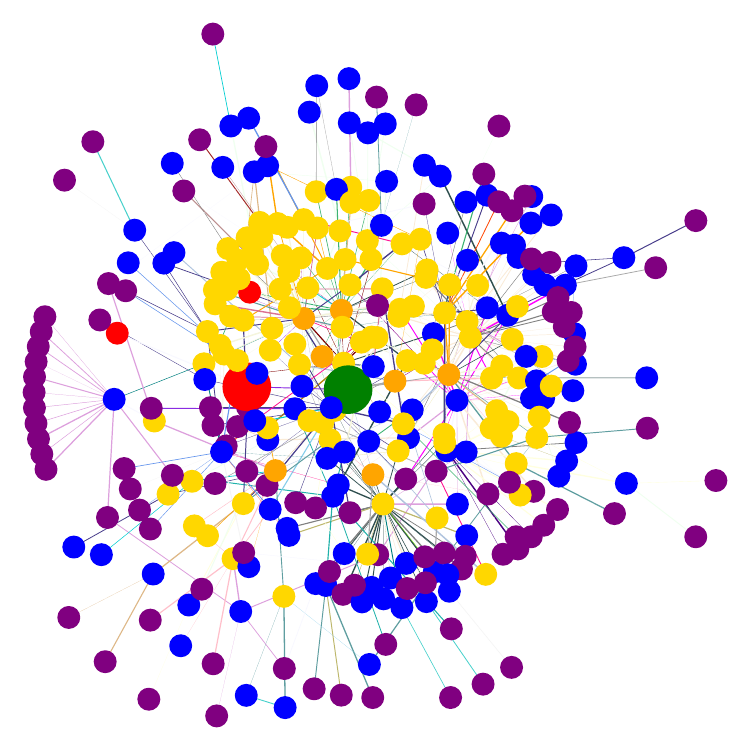}}\quad
	\subfigure[Progressive, $q_2$]{\includegraphics[height=3.4cm]{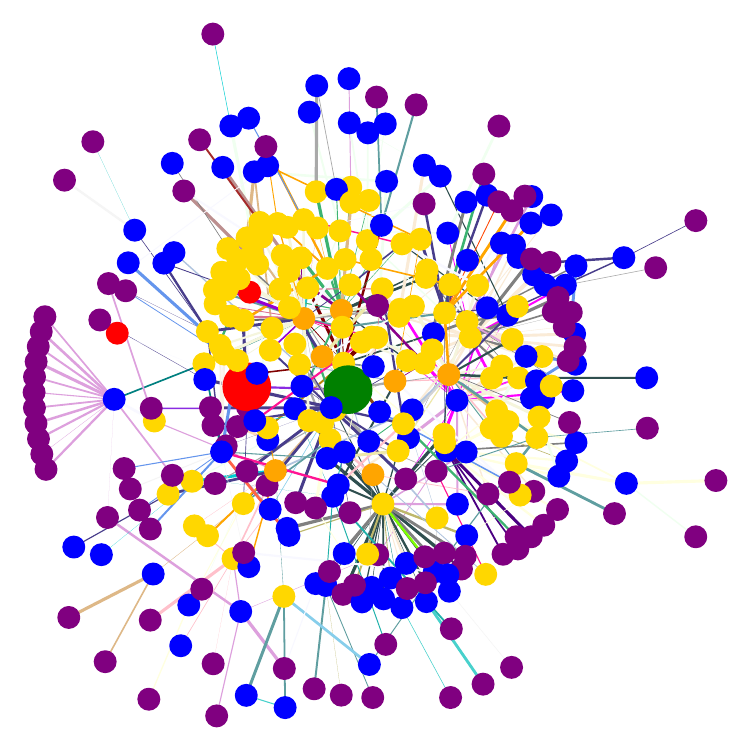}}
	%	\caption{Examples of the propagation paths.}
	\vspace{-10pt}
	\caption{Exemplar propagation paths on {\sf FB15k237-v1} dataset. 
		Specifically,
		the big red and green nodes indicate the query entity and answer entity, respectively. 
		The small nodes in red, yellow, orange, blue, purple are entities involved in $1 \! \sim \! 5$ propagation steps.
		Besides,
		the colors and thickness of edges indicate the relation types and the learned edge weights, respectively.
	}
	\label{fig:more_visualized_examples_2}
\end{figure*}

\noindent \\
In Section~\ref{sssec:sample:learn},
we claim that
the sampling distribution is related to the query-dependent representations 
{\small $\bm h_{e_o}^\ell$}.
We plot more examples of the propagation path of AdaProp (with {\small $L\!=\!5, K\!=\!10$})
and RED-GNN (with {\small $L=5$})
for two different queries $q_1=(e_q, r_1,?)$ and $q_2=(e_q,r_2,?)$ 
in Figure~\ref{fig:more_visualized_examples_2},
which have the same query entity $e_q$ but different query relations 
(i.e., $r_1\neq r_2$)
on {\sf FB15k237-v1} dataset.
%We provide more visualization results 
%in Fig~\ref{fig:more_visualized_examples_2}.
%In each line,
%we show the queries $q_1$ and $q_2$ with the same query entity but different query relations.
%
As shown,
there is a clear difference between
the two propagation paths of $q_1$ and $q_2$ learned by AdaProp.
But they are the same for the progressive propagation methods,
where only the edge weights are different.
The above visualization cases further demonstrate that
\textit{AdaProp indeed can learn the query dependent propagation paths.}

\begin{figure}[H]
	\vspace{-5px}
	\centering
%	\vspace{-8px}
	\subfigure[{\small $\widehat{\mathcal G}^L(\bm \theta)$} of query (2720, son, ?).]
%	{\includegraphics[height=3.7cm]{figures/family1}}
	{\includegraphics[height=3.8cm]{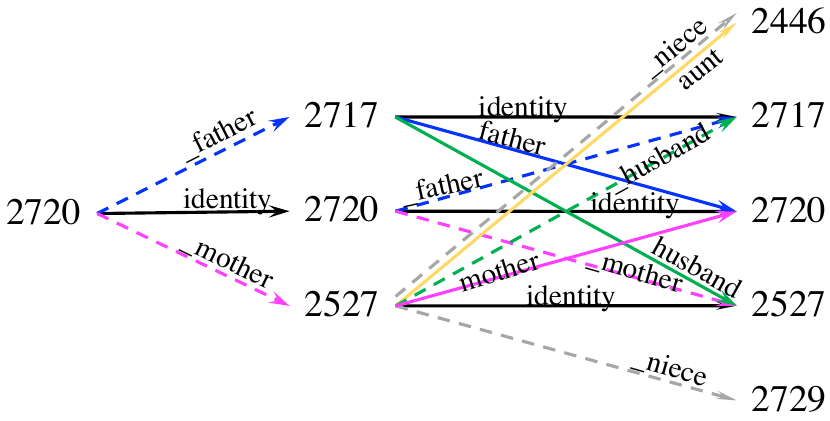}}
	%	\qquad
	\subfigure[{\small $\widehat{\mathcal G}^L(\bm \theta)$} of query (2720, nephew, ?).]
%	{\includegraphics[height=3.6cm]{figures/family2}}
	{\includegraphics[height=3.7cm]{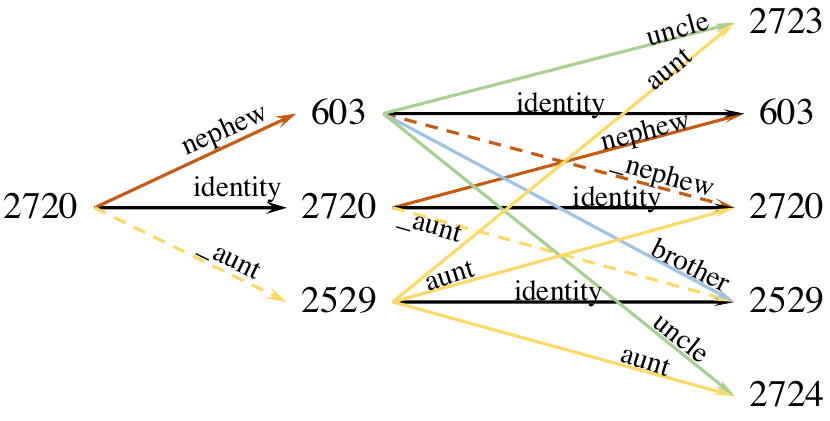}}
	\vspace{-12px}
	\caption{Visualization of the learned propagation path 
		%		(with $L \! = \! 2,K \! = \! 2$) 
		on {\sf Family} dataset. 
		Dashed lines mean reverse relations.
		%		with "\_" in front.
	}
	\label{fig:family_visualization}
	%	\vspace{-10px}
\end{figure}

We also plot the learned propagation paths
{$\widehat{\mathcal G}^L(\bm \theta)$} with {$L\!=\!2, K\!=\!2$}
in Figure~\ref{fig:family_visualization}
with the {\sf Family} dataset.
For the query $(2720, son, ?)$,
the propagation path goes to find who is the father or mother of $2720$.
As for $(2720, nephew, ?)$ with the same query entity,
the propagation path tends to select the aunts and uncles.
This observation shows that \textit{the learned propagation path is semantically-aware.}
In addition,
since AdaProp uses incremental sampling, 
all the entities in previous steps are preserved during propagation steps.
Such a design can provide a further \textit{explanation} to GNNs
due to its \textit{tractable} as well as \textit{interpretable} reasoning procedures.

%It can be observed that
%the performances change more substantially
%when varying the $\tau$ value.
%We speculate that
%the influence of different $\tau$
%is relevant to the KG properties: 
%since the propagation path 
%without sampling 
%will swiftly cover all the entities
%as it goes further steps.
%AdaProp only samples
%a small fraction 
%of the entire propagation path, 
%and it is likely that 
%a higher $\tau$ encourages 
%to sample more entities 
%with low sampling probability 
%in their categorical distribution.
%Hence, the Gumbel top-k scheme
%can help to train the reasoning model
%by injecting a certain amount of 
%stochastic signals .
%in the training phase.
%FB15k237-v1 is much denser than WN18RR.
%Overall,
%the $\tau$ values 
%do not present regular pattern
%w.r.t. the model performances,
%and thus
%searching for an appropriate $\tau$
%would adapt better for a specific KG.

\subsection{Distributions of degree and running time}

%**Q5.** How do you handle the case when there is a hub node in the graph?
%We did not explicitly handle the hub nodes 
%since the complexity is controlled under incremental sampling. 
Here, we show the degree distribution and running time distribution 
in the following Table~\ref{tab:running time} and Table~\ref{tab:degree}. 
Comparing std with mean value, we observe that the running time of different queries does not change much. Based on Proposition~\ref{pr:complexity}, the number of sampled entities is linearly bounded. Thus, the running time is still under control even if hub nodes appear.

\begin{table}[ht]
	\centering
	\caption{Distribution of running time (in seconds, BS=1).}
	\label{tab:running time}
	\vspace{-10px}
	\setlength\tabcolsep{10pt}
	\begin{tabular}{c|cccc}
		\toprule
		dataset & max & min & average & std \\ 
		\midrule
		WN18RR     & 0.13  & 0.02 & 0.03   & 0.01   \\
		FB15k237  & 0.22  & 0.03 & 0.13   & 0.0  \\
		NELL-995   & 0.16  & 0.02 & 0.07   & 0.0  \\
		YAGO3-10   & 0.44  & 0.11 & 0.32   & 0.0  \\	
		\bottomrule
	\end{tabular}
%	\vspace{-10px}
\end{table}

\begin{table}[ht]
	\centering
	\caption{Distribution of degree.}
	\label{tab:degree}
	\vspace{-10px}
	\setlength\tabcolsep{10pt}
	\begin{tabular}{c|cccc}
		\toprule
		dataset & max & min & average & std \\ 
		\midrule
		WN18RR   & 494 &  0  &  2.3   &  3.7 \\ 
		FB15k237 & 1518 &  0  & 21.3   & 33.5 \\ 
		NELL-995  & 1011 &  0  &  2.1   & 10.9 \\ 
		YAGO3-10  & 264 &  0  &  8.8   &  8.8 \\ 
		\bottomrule
	\end{tabular}
%	\vspace{-5px}
\end{table}

\subsection{Detailed values of $\texttt{TC}(L)$ and $\texttt{EI}(L)$}

%**Q3.3** Although Figure 3 shows the ratio of target over entities per hop, the specific TC values (the ratio of target entity coverage) are not clear. 
%**Reply:** Thank you for the suggestion. 

Here,
we provide the values of $\texttt{TC}(L)$ and $\texttt{EI}(L)$ in Eq.\eqref{eq:factor}.
As can be seen from the following
Table~\ref{tab:TC} and Table~\ref{tab:EI}.
AdaProp has smaller values of TC compared with the baselines since it conducts sampling. However, the gap is not large ($<10\%$) in all the cases.

\begin{table}[ht]
	\centering
	\caption{The values of $\texttt{TC}(L)$.
	Here,
	\textit{Full} indicates the full propagation methods,
	while \textit{Progressive} means the methods with progressive propagation.
	}
	\label{tab:TC}
	\vspace{-10px}
	\setlength\tabcolsep{5.3pt}
	\begin{tabular}{c|ccccccc}
		\toprule
		WN18RR           & 2    & 3    & 4    & 5    & 6    & 7    & 8    \\
		\midrule
		Full                    & 1.00 & 1.00 & 1.00 & 1.00 & N.A. & N.A. & N.A. \\
		Progressive      & 0.44 & 0.66 & 0.73 & 0.82 & 0.86 & 0.86 & N.A. \\
		AdaProp          & 0.44 & 0.64 & 0.70 & 0.80 & 0.83 & 0.84 & 0.85 \\
		\midrule
		NELL-995         & 2    & 3    & 4    & 5    & 6    & 7    & 8    \\
		\midrule
		Full          & 1.00 & 1.00 & 1.00 & N.A. & N.A. & N.A. & N.A.  \\
		Progressive   & 0.58 & 0.94  & 0.97 & 0.98 & 0.99 & N.A. & N.A. \\
		AdaProp          & 0.58 & 0.85 & 0.88 & 0.90 & 0.94 & 0.97 & 0.98 \\
		\bottomrule
	\end{tabular}
%	\vspace{-5px}
\end{table}

\begin{table}[ht]
	\centering
	\caption{The values of $\texttt{EI}(L)$.}
	\label{tab:EI}
	\vspace{-10px}
	\setlength\tabcolsep{3pt}
	\begin{tabular}{c|ccccccc}
		\toprule
		WN18RR           & 2    & 3    & 4    & 5    & 6    & 7    & 8    \\
		\midrule
		Full          & 43410 & 43410 & 43410 & 43410 & N.A. & N.A. & N.A. \\
		Progressive   & 59 & 281 & 1159 & 3773 & 9703 & 18935 & N.A. \\
		AdaProp          & 59 & 268 & 848 & 1705 & 2641 & 3592 & 4544 \\
		\midrule
		NELL-995         & 2    & 3    & 4    & 5    & 6    & 7    & 8    \\
		\midrule
		Full          & 74536 & 74536 & 74536 & N.A. & N.A. & N.A. & N.A. \\
		Progressive   & 691 & 5171 & 19674 & 35298 & 43820 & N.A. & N.A. \\
		AdaProp          & 596 & 1966 & 3744 & 5569 & 7406 & 9242 & 11081 \\
		\bottomrule
	\end{tabular}
%	\vspace{-5px}
\end{table}

\subsection{Discussion on samples that get improved with larger steps}

%**Q3.4** And I suggest authors show and discuss the queries whose metrics are improved in the larger layer propagation.
%**Reply:**  Thank you! This is a good suggestion to get deep into our method. 

We add analysis of samples that get improved with larger steps. 
Specifically, we select the samples $(e_q,r_q,e_a)$ that are not correctly predicted at step 4 but are correctly predicted at all the steps $L>4$ (denoted as Group 4). 
The criterion to judge whether a sample is correctly predicted is based on whether the ranking of $e_a$ is $\leq10$. We do the same steps for steps 5 and 6 (denoted as Group 5 and 6).

\begin{table}[ht]
	\centering
	\caption{Distribution of samples in each group.}
	\label{tab:distribution-group}
	\vspace{-10px}
	\setlength\tabcolsep{4pt}
	\begin{tabular}{cc|ccccc}
		\toprule
		& N-sample & dist=1 & dist=2 & dist=3 & dist=4 & dist$\geq$5 \\
		\midrule
		Group 4     &  743    &  1.6\%    & 16.3\%     & 42.7\%     &  22.7\%   &   16.7\%   \\
		Group 5     &  366    &  1.9\%    & 11.5\%     &  34.4\%    &  31.1\%    &  21.1\%    \\
		Group 6     &  118    &  1.7\%    & 9.3\%     &  24.6\%    & 35.6\%    &  28.8\%    \\
		\bottomrule
	\end{tabular}
%	\vspace{-5px}
\end{table}

In the above Table~\ref{tab:distribution-group}, we show the number of samples in each group, and the distribution of the shortest path distance on WN18RR. We observe that
(1) the number of samples that will be improved at larger steps gets smaller with increased values of $L$ (indicated by N-sample values);
(2) comparing the shortest part distances, Group 6 will have farther away distance samples than Group 5, and the same for Group 5 v.s. Group 4.
These results indicate that the samples 
that are improved in the larger steps 
are mainly those that have larger shortest path distances.

\end{document}